\documentclass{article}

\usepackage{times}
\usepackage{latexsym}
\usepackage{ifthen}
\usepackage{algorithm}
\usepackage{graphicx}
\usepackage{url}
\usepackage{subfigure}
\usepackage{framed}
\usepackage{amsmath}
\usepackage{color}
\usepackage{nicefrac}
\usepackage{textcomp}
\usepackage{enumerate}
\usepackage{pgfplots,pgfplotstable}
\usepackage{filecontents}
\usepackage{tabularx}
\usepackage{float}
\usepackage{amssymb}
\usepackage{bbm}
\usepackage{multirow}

\usepackage[T1]{fontenc}
\usepackage[utf8]{inputenc} 
\usepackage[english]{babel} 
\usepackage[stable]{footmisc} 
\usepackage{booktabs} 
\usepackage{longtable} 
\usepackage{listings} 
\usepackage{colortbl}
\usepackage[noend]{algpseudocode}
\usepackage{ellipsis} 
\usepackage{enumitem} 
\usepackage[shortcuts]{extdash} 
\usepackage{setspace} 
\usepackage{hyperref} 
\usepackage{subfigure}
\usepackage{relsize}

\usepackage[draft]{todonotes}   

\usepackage{pgfplots}

\pdfoutput=1

\newboolean{comments}
\setboolean{comments}{true}

\newcommand{\mnist}{\ensuremath{\textit{MNIST}}}
\newcommand{\mnitransfer}{\ensuremath{\mnist_{\textit{transfer}}}}

\newcommand{\mniappr}{\ensuremath{\mnist_{\textit{appear}}}}

\newcommand{\mniflip}{\ensuremath{\mnist_{\textit{flip}}}}
\newcommand{\mniremap}{\ensuremath{\mnist_{\textit{remap}}}}
\newcommand{\mnirotate}{\ensuremath{\mnist_{\textit{rotate}}}}

\newcommand{\nist}{\ensuremath{\textit{NIST}}}
\newcommand{\nitransfer}{\ensuremath{\nist_{\textit{transfer}}}}

\newcommand{\niappr}{\ensuremath{\nist_{\textit{appear}}}}

\newcommand{\niflip}{\ensuremath{\nist_{\textit{flip}}}}
\newcommand{\niremap}{\ensuremath{\nist_{\textit{remap}}}}
\newcommand{\nirotate}{\ensuremath{\nist_{\textit{rotate}}}}

\newcommand{\patching}{\textit{Patching}}
\newcommand{\nnpatching}{\textit{NN-Patching}}

\definecolor{green}{RGB}{0,180,0}
\definecolor{gray}{RGB}{128,128,128}

\graphicspath{{./img/}}

\title{Towards Neural Network Patching: Evaluating Engagement-Layers and Patch-Architectures}

\author{Sebastian Kauschke, David H. Lehmann\\ \\Knowledge Engineering Group\\TU Darmstadt, Germany\\ kauschke@ke.tu-darmstadt.de}

\begin{document}

\maketitle

\begin{abstract}
In this report we investigate fundamental requirements for the application of classifier patching \cite{Kauschke2018} on neural networks. 
  Neural network patching is an approach for adapting 
  neural network models to handle concept drift in nonstationary environments. Instead of creating or updating the existing network to accommodate concept drift,
  neural network patching leverages the inner layers of the network as well as its output to learn a patch that enhances the classification and corrects errors caused by the drift.
  It learns (i) a predictor that estimates whether the original network will misclassify an instance, 
  and (ii) a patching network that fixes the misclassification.
  Neural network patching is based on the idea that the original network can still classify a majority of instances well, and that the inner feature representations encoded in the deep network aid the classifier to cope with unseen or changed inputs.
 In order to apply this kind of patching, we evaluate different engagement layers and patch architectures in this report, and find a set of generally applicable heuristics, which aid in parametrizing the patching procedure.
\end{abstract}


\newpage
\tableofcontents
\newpage

\section{Introduction}
\label{sec:intro}

Nowadays, machine learning research is dominated by neural networks. Although they have been around since the 1940s, it took a long time to leverage their potential, mostly because of the computational complexity involved. This changed in the mid 2000s, when new methods and hardware emerged that allowed bigger networks to be trained faster, opening up new possibilities of application. 
The main advantage of deep networks is their layered architecture, which turns out to be easier to train compared to networks with a single hidden layer, given enough training data is present.

The possibility of training bigger and deeper networks has enabled neural networks to deal with more complex problems.
The current understanding of this is, that
each layer of a network
represents a different stage of abstraction from the input data, similar to how we believe the human brain processes information. 
Besides of the abstraction, specific functional units such as
convolutional layers or long-short-term-memory units provide functionality that is beneficial to certain problems, for example when dealing with image data or sequential prediction tasks.
A typical network for image classification consists of multiple layers of convolutional units \cite{He2016}, each representing feature detectors with different grades of abstraction. Early layers detect simple structures like edges or corners. Later layers comprise more complex features related to the given task, for example eyes or ears, when recognizing faces.

Due to the large amounts of data available today, building highly capable deep neural networks for certain tasks has become feasible. However, most domains are subject to changing conditions in the long run. That means, either the data, the data distribution, or the target classification function changes. This is usually caused by concept drift or other kinds of non-stationarity. The result is that once perfectly capable systems degrade in their performance or even become unusable over time.

An example could be an image classification task, where previously unknown classes need to be detected. This usually requires a retraining of at least a part of the network, in order to accommodate the new classes.
Another example could be a piece of complex machinery, as used in productive environments such as factories. This machine might be fitted with hard- and software to finely detect its current state, and a predictive model for failures on top of it. When the next hardware revision of that machine is sold by the manufacturer, new data from the machine has to be collected and the failure predicting model has to be retrained, which can be very expensive.
A final example to motivate the necessity of adaptation is the personalization setting. A product is sold with a general prediction model that covers a wide variety of users. However, personalization would help to make it even more suitable for a specific user. This is a type of adaptation that is difficult to manage with neural networks as underlying models.

In order to solve these problems, we build upon \emph{patching}, a framework that has recently been proposed to cope with such problems \cite{Kauschke2018}.
Contrary to many conventional techniques, this framework does 
not assume that it is feasible to re-train the model from scratch with newly recorded data. 
Instead, it tries to recognize regions where the model errs, and tries to learn local models---so-called \emph{patches}---that repair the original model in these error regions.

In this paper, we present \emph{neural network patching} (\nnpatching), a variant of patching that is specifically tailored to neural network classifiers.
We recognize the fact, that building a well-working neural network for a certain task can be cumbersome and require many iterations w.r.t. the choice of architecture and the hyper-parameters. Once such a network is established and properly trained, a prolonged use of it is usually appreciated. 
However, it is not guaranteed that the underlying problem domain remains stationary, and it is desirable that the network can adapt to such changes.
\nnpatching{} allows existing neural networks to be adapted to new scenarios by adding a network layer on top of the existing network. This layer is not only fed by the output of the base network, but also leverages inner layers of the network that enhance its capabilities. Furthermore, the patching network is only activated, when the underlying base network gives erroneous results.

This report is structured as follows. 
In Section~\ref{sec:adaptingdnn} we elaborate on the concept of \nnpatching{} and define its requirements. We derive a set of experiments in Section~\ref{sec:setup} and test various assumptions and methods based on this setup in Sections~4 and 5. 
We compare our methodology against known transfer learning mechanisms in Section~\ref{Final Results}, and conclude our findings in Section~\ref{sec:conclusion}.

\newpage
\section{Deep Neural Network Adaptation}
\label{sec:adaptingdnn}
Since neural networks are usually trained by backpropagation, adapting a neural network towards a changed scenario can be achieved via training on the latest examples, hence refining the weights in the network towards the current concept.
However, this may lead to catastrophic forgetting \cite{French1999} and---depending on the size of the networks---may be costly. To mitigate this issue, a common approach is to train only part of the network and not adapt the more general layers \cite{Yosinski2014}, but only the specific layers relevant to the target function. For example, \cite{Ciresan2012}  leverages this behavior to achieve transfer to problems with higher complexity than the original problem the network was intended for. In summary, we make three observations:
\begin{itemize}
\item[(i)] neural networks are useful towards adaptation tasks, caused by their hierarchical structure, 
\item[(ii)] neural networks can be trained such that they adapt to changed environments via new examples, and 
\item[(iii)] this adaptation may lead to catastrophic forgetting. 
\end{itemize}
In our proposed method, we want to leverage the advantages of (i) and (ii), but avoid the disadvantages of (iii). In the next Section we will explain the patching procedure for neural networks.

\subsection{Patching for Neural Networks}
\label{sec:nnpatching}

\begin{figure}{t}
\centering
\includegraphics[width=0.5\textwidth]{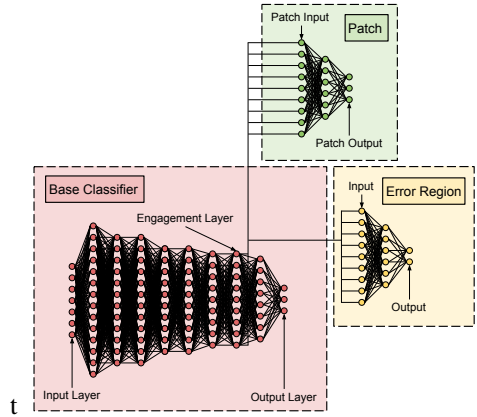}
\caption{The \nnpatching{} algorithm engages into one of the inner layers of the base classifier}
\label{fig:nnpatching_engagement}
\end{figure}

\begin{figure}[t]
	\centering
		\subfigure[\label{fig:patch1} Base classifier. Dashed black line symbolizes the decision boundary of the base classifier.]{\includegraphics[scale=0.25]{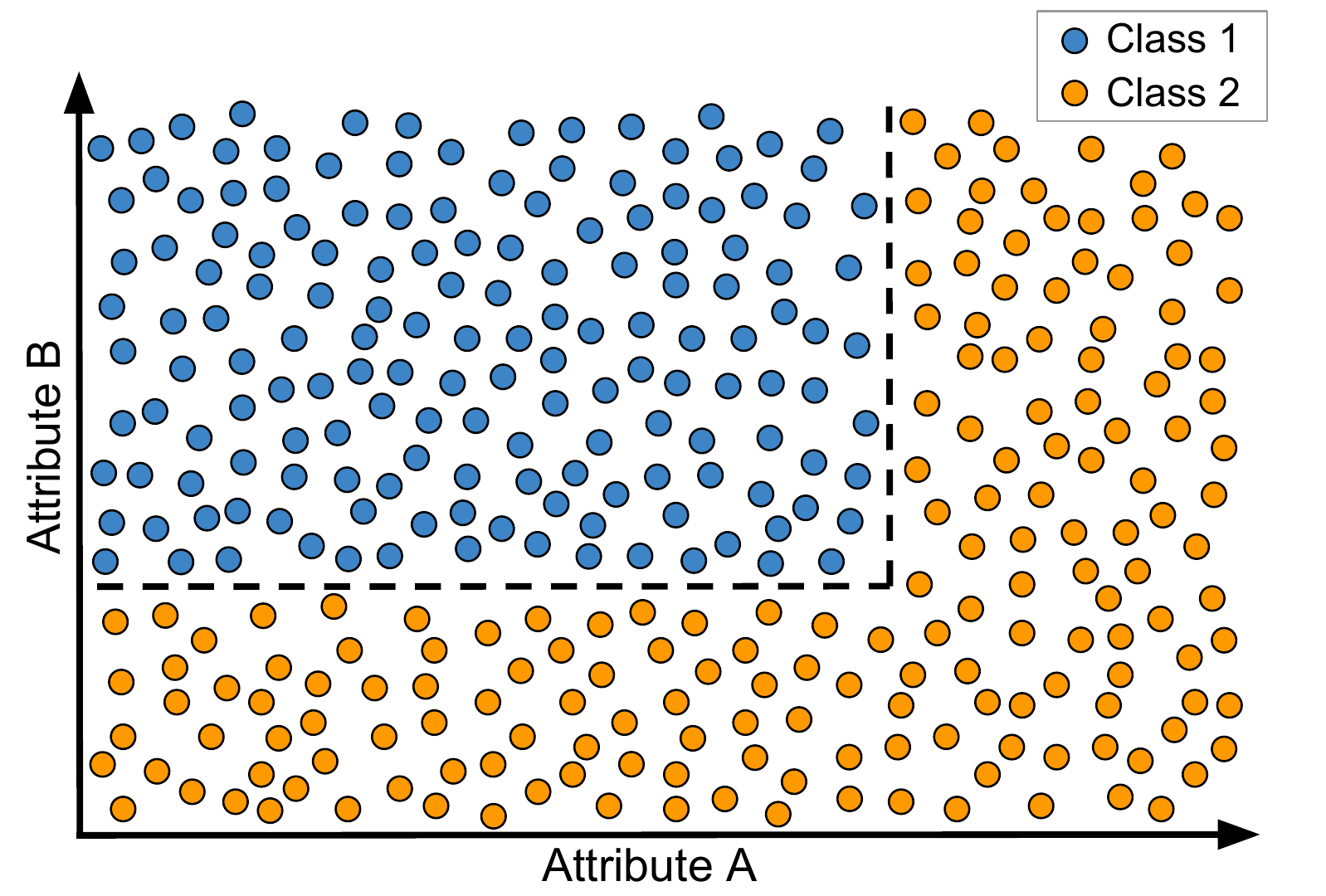}}
		\hspace{0.5cm}\subfigure[\label{fig:patch2} Concept drift. Some areas in the instance space are now error-prone.]{\includegraphics[scale=0.25]{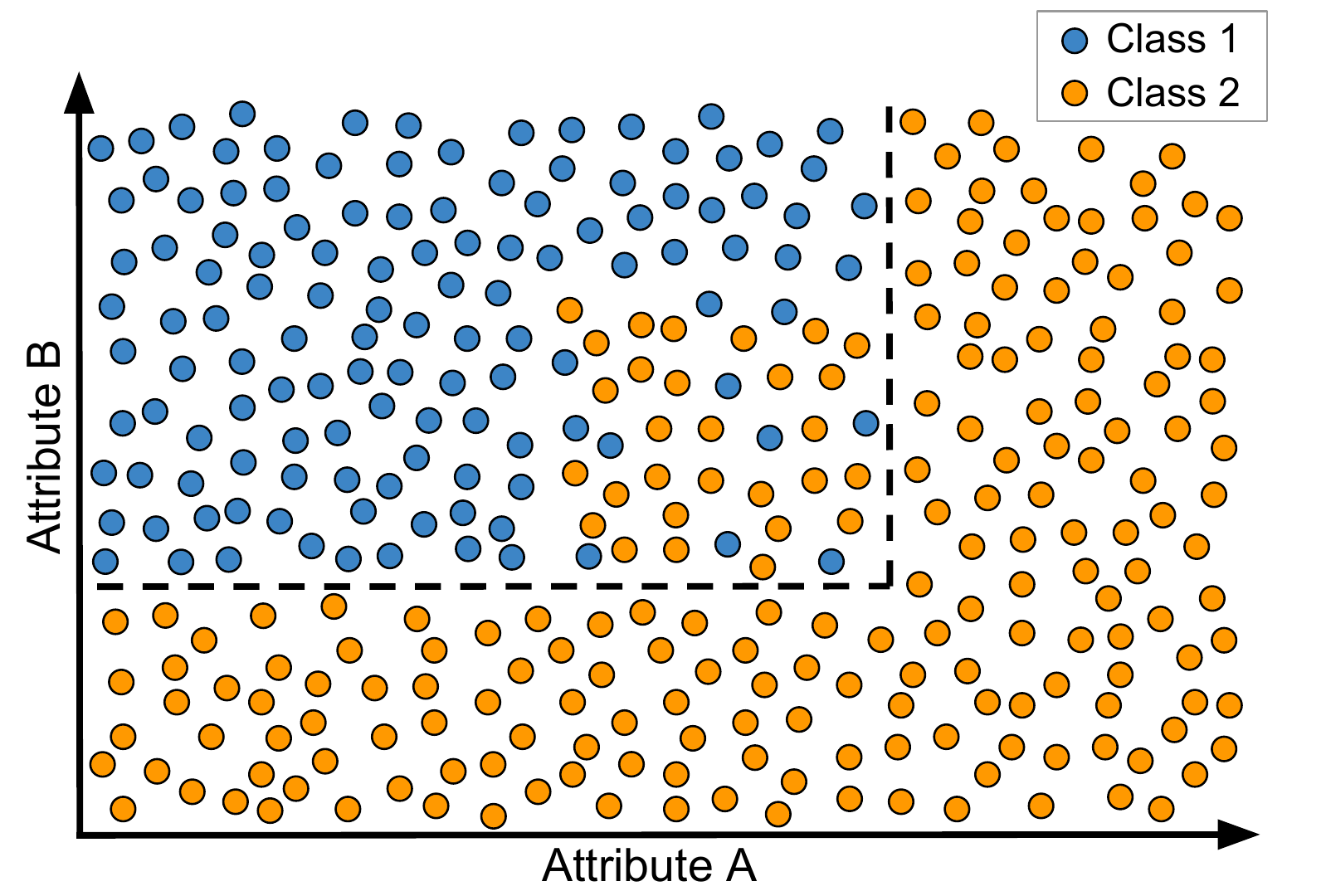}}
		\subfigure[\label{fig:patch3} Patching error regions. The green area symbolizes the detected error region.]{\includegraphics[scale=0.25]{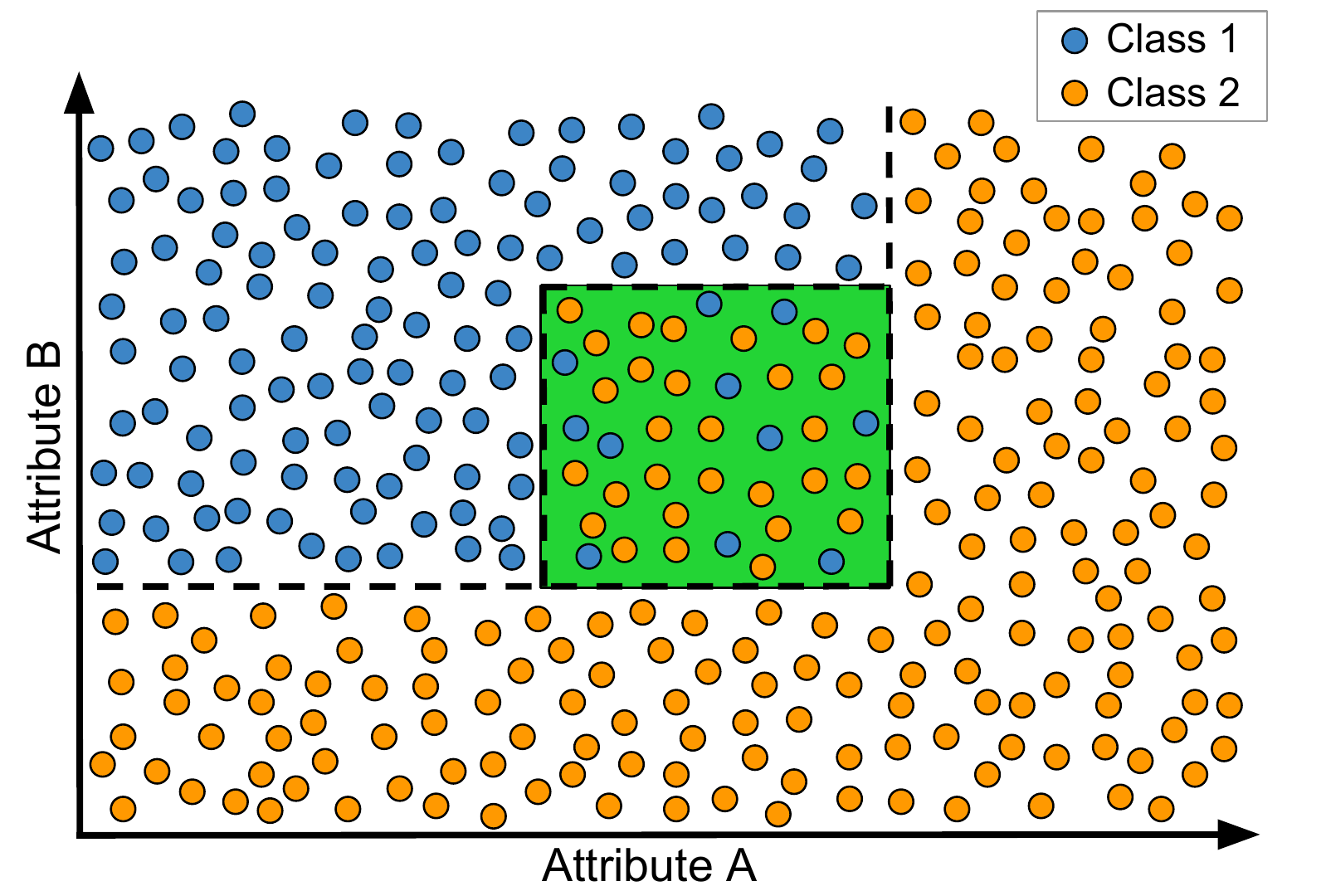}}
	\caption{\textbf{Patching Algorithm.} (a) shows the instance space of the classes 1 and 2. The dashed black line marks the decision boundary of the base classifier. The instances are classified satisfactory. In (b), concept drift occurs. We have a error-prone region in the instance space. In (c) the error region is detected and a patch classifier is learned. The classification of an instance from the error region is diverted to the patch.}
	\label{fig:patching}
\end{figure}

We tailor the \patching{}-procedure \cite{Kauschke2018} to the specific case of neural network classifiers. The idea is depicted in Figure~\ref{fig:patching}. 
\patching{} relies on the existence of a given classifier $M$, which is able to classify an existing scenario well. This is also a requirement for \nnpatching{}, but here we rely on $M$ being a (deep) neural network. The \nnpatching{} procedure consists of three steps: 
\begin{enumerate}
	\item \textbf{Learn a classifier $\mathbf{E}$ that determines where $\mathbf{M}$ errs.}
		In this step, when receiving a new batch of labeled data, the data is used to learn a classifier that estimates where $\mathbf{M}$ will misclassify instances.
	\item \textbf{Learn a patch network $\mathbf{P}$.}
		The patch network engages to one inner layer of $\mathbf{M}$, and takes the activations of these layers as its own input (Fig.~\ref{fig:nnpatching_engagement}).
	\item \textbf{Divert classification from $\mathbf{M}$ to $\mathbf{P}$, if $\mathbf{E}$ is confident.} When an instance is to be classified, the error detector $E$ is executed. If the result is positive, classification is diverted to $P$, otherwise to $M$.
\end{enumerate}

In contrast to the original procedure (cf. \cite{Kauschke2018}), neural networks enable us to iteratively update both $E$ and $P$ over time. We will hence not create separate versions for each new batch, but rely on the existing one and update it via backpropagation with the instances from the latest batch.

In order to learn the patch network, we must engage in one of the inner layers of $M$. The selection of this layer is non-trivial. Furthermore, we need to determine an appropriate architecture for the patch $P$ itself. The experiments described in the next sections will aid in determining some generalized heuristics to approach this parametrization problem.

\section{Experimental Setup}
\label{sec:setup}

\begin{figure}[t]
\centering
\includegraphics[scale=0.4]{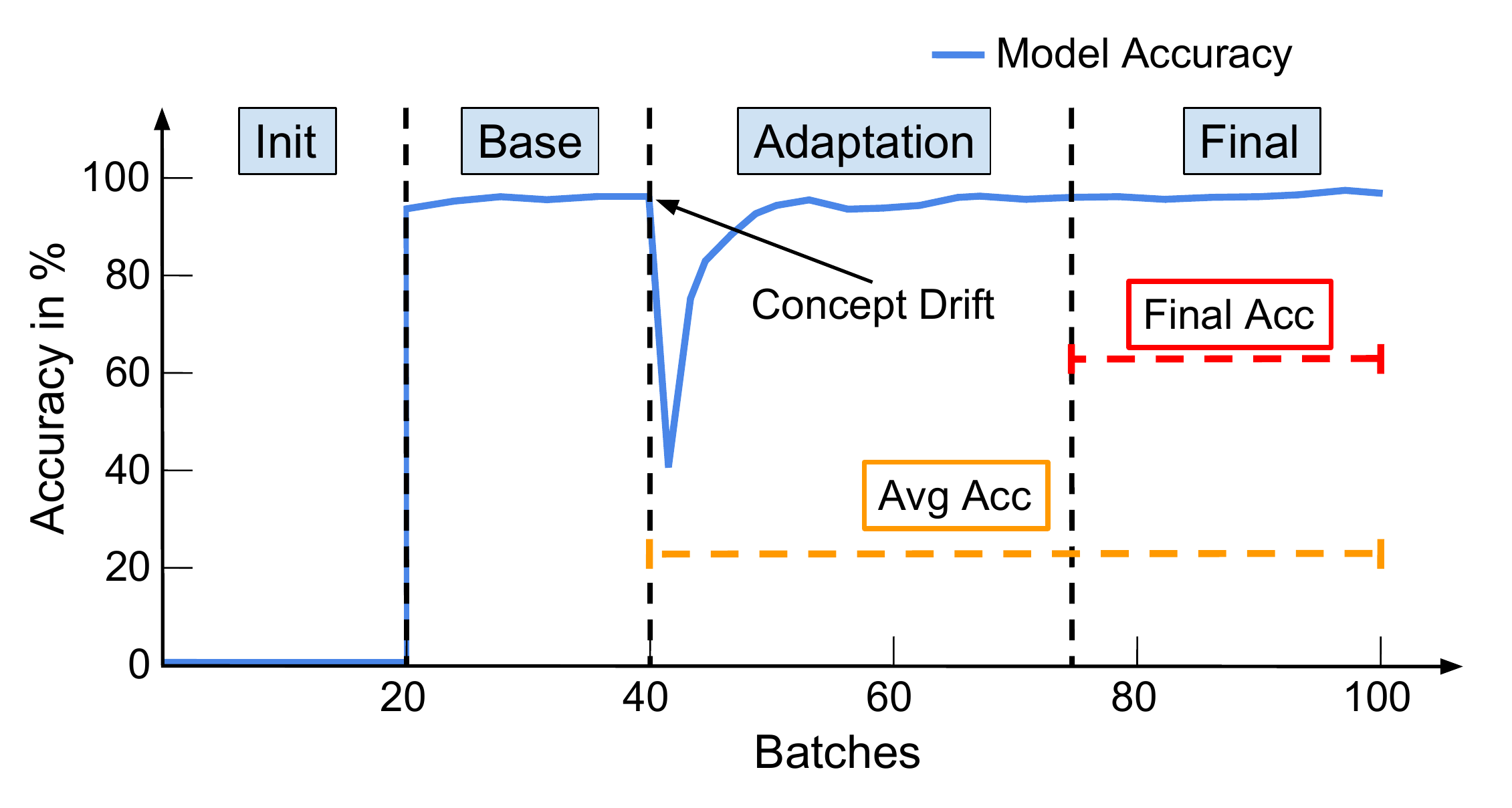}
\caption{Different phases of the experimental scenario and evaluation measures.}
\label{fig:scenario}
\end{figure}

In this Section, we will elaborate on the datasets we use to determine optimal engagement layers and patch architecture. Our datasets are derived from well known datasets and are engineered to give a stream of instances, where each stream contains one or multiple drifts of the underlying concept. We evaluate these streams as sequence of instances, where the true labels are retrieved in regular intervals. These are so called batches of instances. On the end of each batch it allows us to retrospectively evaluate the performance of the classifier, and make adaptations for the next batch, a so called test-then-train evaluation strategy. Bifet et al. \cite{BifetMOA} describe this process more thoroughly w.r.t. their Massive Online Analysis (MOA) framework. We applied the same principles, although we implemented our solution in Python.

\subsection{Evaluation Datasets}
\label{sec:datasets}

\begin{table}[H]
	\centering
	\footnotesize
	\caption{Summary of the datasets used in the experiments}
	\label{tab:datasets}
	\begin{tabular}{|l|c c c c |}
		\hline
		{Dataset} &  Init & CPs & Total & Chunks \\
		\hline\hline
		\multicolumn{5}{|c|}{\emph{MNIST Dataset}}\\
		\hline 
		\mniflip   & 40k & \#70k & 140k & 100  \\
		\mnirotate & 20k & \#35k & 70k & 100 \\
		\mniappr   & 15k & \#20.4k & 57.2k & 100 \\
		\mniremap & 20k & \#35.7k & 70k & 100 \\
		\mnitransfer & 20k & \#35.7k  & 70k & 100 \\
		\hline \hline
		
		\multicolumn{5}{|c|}{\emph{NIST Dataset}}\\
		\hline 
		\niflip   & 30k & \#40k & 100k & 100  \\
		\nirotate & 30k & \#40k &  100k & 100 \\
		\niappr   & 20k & \#28.6k & 88.6k & 100 \\
		\niremap  & 20k & \#28k & 55.8k & 100 \\
		\nitransfer & 20k & \#30k  & 80k & 100  \\
		\hline 
	\end{tabular}
\end{table}

We will evaluate our findings in 10 scenarios which are based on two datasets. Each scenario represents a different type of concept drift with varying severity up until a complete transfer of knowledge to an unknown problem. The scenarios are summarized in Table~\ref{tab:datasets}.
\subsubsection{The MNIST Dataset.} 
The first dataset is the \mnist\footnote{\url{http://yann.lecun.com/exdb/mnist/}} dataset of handwritten digits. It contains the pixel data of 70,000 digits (28x28 pixel resolution), which we treat as a stream of data and introduce changes to. We created the following scenarios.

\begin{itemize}
	\item \mniflip: The second half of the dataset consists of vertically and horizontally flipped digits.
	\item \mnirotate: The digits in the dataset are rotated at a random angle from instance \#35k onwards with increasing degree of rotation up to 180 degrees (at \#65k).
	\item \mniappr:  The digits change during the stream, such that classes 5--9 do not exist in the beginning, but only start to appear at the change point (in addition to 0--4).
	\item \mniremap: In the first half, only the digits 0--4 exist. The input images of 0--4 are then replaced by the images of 5--9 for the second half (labels remain 0--4). Here we only have 5 classes.
	\item \mnitransfer: The first half of the stream only consists of digits 0-4, while the second half only consists of the before unseen digits 5-9.
\end{itemize}
An overview of the used \textit{MNIST} datasets is given in Table \ref{tab:mnist}. The second column \textit{Init} refers to as the amount of instances used to train the base classifier. The third column \textit{Change Point} (\textit{CP}) refers to as the instance where the concept drift occurs. The column \textit{Total} states the size of the dataset in instances and \textit{Chunks} state the number of batches the dataset is divided into. 

\begin{table}[h]
\begin{centering}
  \begin{tabular}{ | l | l  l  l  l |}
    \hline
		\multicolumn{5}{|c|}{\textit{MNIST} Datasets} \\ \hline
		Dataset & Init & CP & Total & Chunks \\ \hline
		$\mathbf{\textit{MNIST}_{\textit{appear}}}$ & 15k & \#20.4k & 50.4k & 100 \\
    $\mathbf{\textit{MNIST}_{\textit{flip}}}$ & 40k & \#70k & 140k & 100 \\ 
		$\mathbf{\textit{MNIST}_{\textit{remap}}}$ & 20k & \#35.7k & 70k & 100 \\
		$\mathbf{\textit{MNIST}_{\textit{rotate}}}$ & 20k & \#35k & 70k & 100 \\ 
		$\mathbf{\textit{MNIST}_{\textit{transfer}}}$ & 20k & \#35.7k & 70k & 100 \\
    \hline  
  \end{tabular}
\caption{Overview of \textit{MNIST} datasets.}
\label{tab:mnist}
\end{centering}
\end{table}
	
\subsubsection{The NIST Dataset.}
The second dataset is the \nist\footnote{\url{https://www.nist.gov/srd/nist-special-database-19}} dataset of handprinted forms and characters. It contains 810,000 digits and characters, to which we will apply similar transformations as to the \mnist{} data. Contrary to \mnist, \nist{} items are not pre-aligned, and the image size is 128x128 pixels. We will use all digits 0-9 and uppercase characters A-Z for a total of 36 classes as data stream and draw a random sample for each scenario. We sample a dataset with 100,000 instances.
\begin{itemize}
	\item \niflip: After instance \#40k, the instances are mirrored vertically and horizontally. We train the base classifier on the first 30k instances.
	\item \nirotate: The images in the dataset start to rotate randomly at instance \#35k with increasing rotation up to $\pm 180$ degrees for the last 10k instances. We train the base classifier on the first 30k instances.
	\item \niappr: The distribution of the images changes during the stream so that instances of classes 0--9 do not exist in the beginning, but only start to appear at the change point (mixed in between the characters A--Z). This variant consists of a total of 88,600 instances.
	We train the base classifier on the first 20k instances.
	\item \niremap: In the first half, only the digits 0--9 exist. The input images are then replaced by the images of the letters A--J for the second half, but the labels remain 0--9. Here we only have 10 classes. We train the base classifier on the first 20k instances, in total there are 55,800 instances.
	\item \nitransfer: The first 30k instances of the stream only consists of digits 0--9, while the following 60k are solely characters A--Z with their respective, correct labels. We train the base classifier on the first 20k instances, in total there are 80k instances.
\end{itemize}

An overview of the used \textit{NIST} datasets is given in Table~\ref{tab:nist}.
\begin{table}[h]
\begin{centering}
  \begin{tabular}{ | l | l  l  l  l |}
    \hline
		\multicolumn{5}{|c|}{\textit{NIST} Datasets} \\ \hline
		Dataset & Init & CP & Total & Chunks \\ \hline
		$\mathbf{\textit{NIST}_{\textit{appear}}}$ & 20k & \#28.6k & 88.6k & 100 \\
    $\mathbf{\textit{NIST}_{\textit{flip}}}$ & 30k & \#40k & 100k & 100 \\
		$\mathbf{\textit{NIST}_{\textit{remap}}}$ & 20k & \#28k & 55.8k & 100 \\
		$\mathbf{\textit{NIST}_{\textit{rotate}}}$ & 30k & \#40k & 100k & 100 \\ 
		$\mathbf{\textit{NIST}_{\textit{transfer}}}$ & 20k & \#30k & 80k & 100 \\
    \hline  
  \end{tabular}
\caption{Overview of \textit{NIST} datasets.}
\label{tab:nist}
\end{centering}
\end{table}

\subsection{The Base Classifiers}
\label{Base Classifier Architectures}
In the original \patching{}-procedure, it is assumed that a base classifier exists, which we can learn errors and build patches upon. Since these are not given in our case, we use part of the dataset to create them based on popular neural network architectures.

We exploited three architectures that are generally suited to solve the scenarios we described: (i) a fully-connected deep neural network (FC-NN), (ii) a convolutional neural network (CNN), and (iii) a residual network (ResNet) architecture.
Each classifier architecture is tuned to achieve high accuracy on the unaltered datasets (Table~\ref{tab:base performance}). We assume ReLU activation for all fully-connected and convolutional layers, except in the ResNet and the residual blocks. In this case the application of ReLU activation is stated explicitly whenever used. The CNN and the FC-NN are trained for 10 epochs on the initialization fraction (Fig.~\ref{fig:scenario}) of the dataset. The ResNet architectures are trained for 20 epochs instead, since the deep structure requires more epochs to lead to convergence. For the training in the initialization phase we use a batch size of 64.
\begin{table}[H]
\begin{centering}
\caption{Base classifier accuracy on unaltered datasets. \nist{} only consists of uppercase letters and numbers. The test set consisted of 10,000 instances.}
  \begin{tabular}{ | l | c  c  c |}
    \hline
		Dataset & FC-NN & CNN & ResNet\\ \hline
    	\mnist & 98.87\% & 99.28\% & 99.35\%  \\
		\nist & 94.07\% & 97.77\% & 98.03\% \\
    \hline  
  \end{tabular}

\label{tab:base performance}
\end{centering}
\end{table}

In the following Sections we show the architectural details w.r.t. layer configuration and activations of the chosen networks.

\subsubsection{Fully-Connected Architectures}
The fully-connected architectures for \nist{} and \mnist{} are stated in table~\ref{tab:Fully-Connected Architectures}. The networks both tend to overfit, hence two dropout layers are utilized to counteract this problem. We use fully-connected layers with decreasing number of nodes to build the architectures.  
\begin{table}[H]
\scriptsize
\begin{centering}
\caption{Fully-Connected Architectures.}
\label{tab:Fully-Connected Architectures}
  \begin{tabular}{ | l | l |}
    \hline
		\multicolumn{1}{|c|}{\mnist} & \multicolumn{1}{c|}{\nist} \\ \hline
		\multicolumn{1}{|c|}{InputLayer(28x28) - Flatten() - Dropout(0.2) -} & \multicolumn{1}{c|}{InputLayer(128x128) - Flatten() - Dropout(0.2) -}\\
		\multicolumn{1}{|c|}{FC(2048) - FC(1024) - FC(1024) - FC(512) -} & \multicolumn{1}{c|}{FC(1024) - FC(1024) - FC(768) - FC(512) -}\\
		\multicolumn{1}{|c|}{FC(128) - Dropout(0.5) - Softmax(\#classes)} & \multicolumn{1}{c|}{FC(512) - FC(256) - FC(256) -}\\
		 & \multicolumn{1}{c|}{Dropout(0.5) - Softmax(\#classes)}\\
		\hline
		\multicolumn{2}{|l|}{InputLayer(i): Input layer, i = shape of the input}\\
		\multicolumn{2}{|l|}{Flatten(): Flatten input to one dimension}\\
		\multicolumn{2}{|l|}{FC(n): Fully Connected, n = number of units}\\
		\multicolumn{2}{|l|}{Dropout(d): Dropout, d = dropout rate}\\
		\multicolumn{2}{|l|}{Softmax(n): FC layer with softmax activation, n = number of units}\\
    \hline  
  \end{tabular}

\end{centering}
\end{table}

\subsubsection{Convolutional Architectures}
In the CNN architectures we additionally use convolutional and pooling layers. In the architecture for \mnist{} only two convolutional layers and one pooling layer are required to achieve an accuracy greater than 99.25\,\%. The \nist{} dataset has a total of 128x128 = 16,384 attributes. We use one convolutional layer with stride=2 and two pooling layers to reduce the dimensionality of the data, while propagating through the network. In both cases we counteract overfitting with the help of two dropout layers.
\begin{table}[H]
\scriptsize
\caption{Convolutional Architectures.}
\label{tab:Convolutional Architectures}
\begin{centering}
  \begin{tabular}{ | l | l |}
    \hline
		\multicolumn{1}{|c|}{\mnist} & \multicolumn{1}{c|}{\nist} \\ \hline
		\multicolumn{1}{|c|}{InputLayer(28x28) - Conv2D(32,(3,3),1) - } & \multicolumn{1}{c|}{InputLayer(128x128) - Conv2D(32,(7,7),2) - }\\
		\multicolumn{1}{|c|}{Conv2D(64,(3,3),1) - MaxPooling((2,2),2) - } & \multicolumn{1}{c|}{MaxPooling((2,2),2) - Conv2D(64,(5,5),1) - }\\
		\multicolumn{1}{|c|}{Dropout(0.25) - Flatten() - FC(128) - } & \multicolumn{1}{c|}{Conv2D(64,(5,5),1) - Conv2D(64,(3,3),1) - }\\
		\multicolumn{1}{|c|}{Dropout(0.5) - Softmax(\#classes)} & \multicolumn{1}{c|}{Conv2D(64,(3,3),1) - MaxPooling((2,2),2) - }\\
		                                         & \multicolumn{1}{c|}{Dropout(0.25) -  Flatten() - FC(256) - }\\
																						 & \multicolumn{1}{c|}{Dropout(0.5) - Softmax(\#classes)}\\
		\hline
		\multicolumn{2}{|l|}{InputLayer(i): Input layer, i = shape of the input}\\
		\multicolumn{2}{|l|}{Flatten(): Flatten input to one dimension}\\
		\multicolumn{2}{|l|}{FC(n): Fully Connected, n = number of units}\\
		\multicolumn{2}{|l|}{Conv2D(f,k,s): 2D Convolution, f = number of filters, k = kernel size, s =stride}\\
		\multicolumn{2}{|l|}{MaxPooling(k,s): Max Pooling, k = kernel size, s =stride}\\
		\multicolumn{2}{|l|}{Dropout(d): Dropout, d = dropout rate}\\
		\multicolumn{2}{|l|}{Softmax(n): FC layer with softmax activation, n = number of units}\\
    \hline  
  \end{tabular}

\end{centering}
\end{table}

\subsubsection{Residual Architectures}
\begin{figure}[t]
	\centering
		\hspace{1cm}\subfigure[\label{fig:Identity block} Identity block]{\includegraphics[scale=0.25]{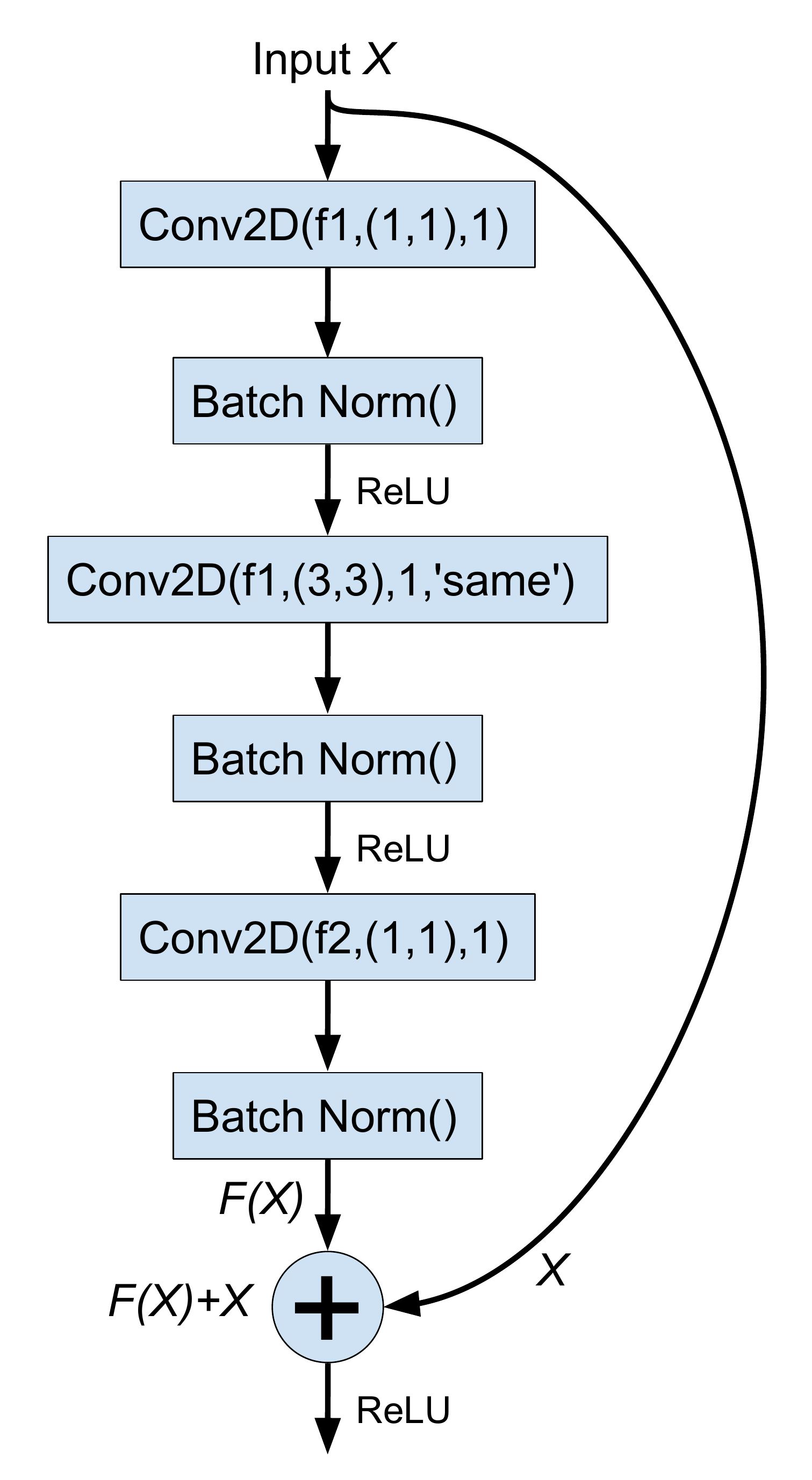}}\hspace{1cm}
		\subfigure[\label{fig:Convolutional block} Convolutional block]{\includegraphics[scale=0.25]{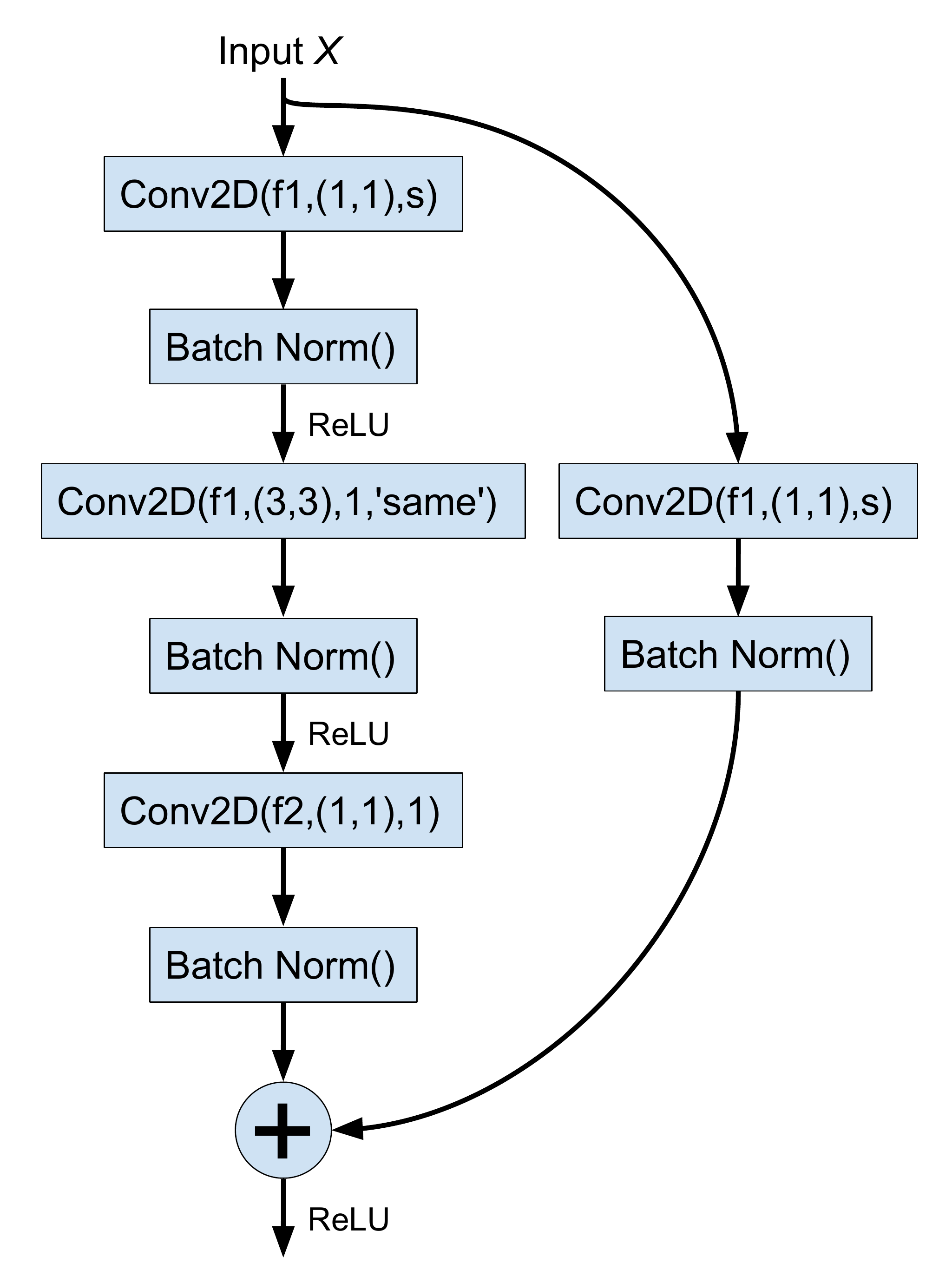}}
	\caption{Residual block types.}
	\label{fig:Residual block types}
\end{figure}

Our ResNet architecture is based on the contest winning model by He et al. \cite{He2016}. It consists of two different residual block types (Figure~\ref{fig:Residual block types}). An important tool in both block types is the 1x1 convolutional layer. 1x1 convolutions can be used to change the dimensionality in the filter space. Both residual block types follow the same pattern. At first a 1x1 convolution is used to reduce the dimensionality, then a 3x3 convolution is applied on the data with reduced dimensionality. Finally, another 1x1 convolution is utilized to restore the original filter space. The reduction of dimensionality results in a reduced computational cost for applying the 3x3 convolutions. The optional layer parameter 'same' refers to zero padding in Keras \cite{chollet2015keras}. Zeros are added around the image in such a way that for stride=1 the width and height for the input and output of the layer would be the same.  
The identity block preserves the input size, whereas the convolutional block can be used to change the width and height of each feature map. Hence, the convolutional block has an additional convolutional layer in the residual connection. With a stride greater than one, the width and height of the block output can be manipulated. The ResNet architectures for \nist{} and \mnist{} are stated in table~\ref{tab:Residual Architectures}.
 
\begin{table}[H]
\scriptsize
\caption{Residual Architectures.}
\label{tab:Residual Architectures}
\begin{centering}
  \begin{tabular}{ | l | l |}
    \hline
		\multicolumn{1}{|c|}{\mnist} & \multicolumn{1}{c|}{\nist} \\ \hline
		\multicolumn{1}{|c|}{InputLayer(28x28) - Dropout(0.2) -} & \multicolumn{1}{c|}{InputLayer(128x128) - Dropout(0.2) -}\\
		\multicolumn{1}{|c|}{Conv2D(64,(5,5),2,'same') - BatchNorm() - } & \multicolumn{1}{c|}{Conv2D(64,(7,7),2) - BatchNorm() - }\\
		\multicolumn{1}{|c|}{ReLU() - ConvBlock((64,256),1) - } & \multicolumn{1}{c|}{ReLU() - MaxPooling((3,3),3) - }\\
		\multicolumn{1}{|c|}{IdBlock(64,256) - IdBlock(64,256) - } & \multicolumn{1}{c|}{ConvBlock((64,256),1) - IdBlock(64,256) - }\\
		\multicolumn{1}{|c|}{ConvBlock((128,512),2) - IdBlock(128,512) - } & \multicolumn{1}{c|}{IdBlock(64,256) - ConvBlock((128,512),2) - }\\
		\multicolumn{1}{|c|}{IdBlock(128,512) - IdBlock(128,512) - } & \multicolumn{1}{c|}{IdBlock(128,512) - IdBlock(128,512) - }\\
		\multicolumn{1}{|c|}{ConvBlock((256,1024),2) - IdBlock(256,1024)  - } & \multicolumn{1}{c|}{IdBlock(128,512) - ConvBlock((256,1024),2) - }\\
		\multicolumn{1}{|c|}{IdBlock(256,1024) - IdBlock(256,1024)  - } & \multicolumn{1}{c|}{IdBlock(256,1024) - IdBlock(256,1024) - }\\
		\multicolumn{1}{|c|}{IdBlock(256,1024) - AveragePooling2D((2,2),2) - } & \multicolumn{1}{c|}{IdBlock(256,1024) - }\\
		\multicolumn{1}{|c|}{Flatten() - Dropout(0.5) - Softmax(\#classes)} & \multicolumn{1}{c|}{AveragePooling2D((2,2),2,'same') - Flatten() - }\\
			\multicolumn{1}{|c|}{} & \multicolumn{1}{c|}{Dropout(0.5) - Softmax(\#classes)}\\
		\hline 
		\multicolumn{2}{|l|}{InputLayer(i): Input layer, i = shape of the input}\\
		\multicolumn{2}{|l|}{Flatten(): Flatten input to one dimension}\\
		\multicolumn{2}{|l|}{Conv2D(f,k,s): 2D Convolution, f = number of filters, k = kernel size, s = stride}\\
		\multicolumn{2}{|l|}{MaxPooling(k,s): Max Pooling, k = kernel size, s =stride}\\
		\multicolumn{2}{|l|}{AvgPooling(k,s): Average Pooling, k = kernel size, s =stride}\\
		\multicolumn{2}{|l|}{IdBlock(f1,f2): Identity Block, f1 =  \#reduced filters, f2 = \#output filters}\\
		\multicolumn{2}{|l|}{ConvBlock((f1,f2),s): Convolutional Block, f1 =  \#reduced filters, f2 = \#output filters, s = stride}\\
		\multicolumn{2}{|l|}{BatchNorm(): Batch Normalization}\\
		\multicolumn{2}{|l|}{ReLU(): ReLU Activation}\\		
		\multicolumn{2}{|l|}{Dropout(d): Dropout, d = dropout rate}\\
		\multicolumn{2}{|l|}{Softmax(n): FC layer with softmax activation, n = number of units}\\
    \hline  
  \end{tabular}

\end{centering}
\end{table}

Although batch normalization is frequently used in ResNet, the CNN seems to be more robust in terms of initialization than the ResNet structure. Ioffe and Szegedy \cite{DBLP:journals/corr/IoffeS15} reported that batch normalization increases the robustness to initialization of a network. On rare occasions the ResNet gets stuck in a local minimum during training. We never observed this for the CNN or FC-NN. In this case the ResNet converges in the first epoch of the training process. During experiment runtime we detect this stagnation in the training process and discard the current model. We reinitialize the network with a new random seed and restart the training process.

\subsection{Evaluation Measures}
\label{sec:evaluationmeasures}
For the comparison of the different architectures we use the following metrics:
\begin{itemize}
	\item{\it Final Accuracy (F.Acc):} Classification accuracy, measured in the \emph{Finish} phase, which consists of the last five batches of the stream.
	\item{\it Average Accuracy (Avg.Acc): Average accuracy in the \textit{Adaptation} and \textit{Finish} phases (after first change point).}
	\item{\it Recovery Speed (R.Spd):} Number of instances that a classifier requires during the \textit{Adaptation} phase to achieve 90\% of its final accuracy.
	\item{\it Adaptation Rank (Ad.Rk):} Average rank of the classifier during the \textit{Adaptation} phase.
	\item{\it Final Rank (F.Rk):} Average rank of the classifier during the \textit{Finish} phase.	
\end{itemize}

\subsection{Adaptation Methods from Transfer Learning}
\begin{figure}[H]
	\centering
	\includegraphics[scale=0.25]{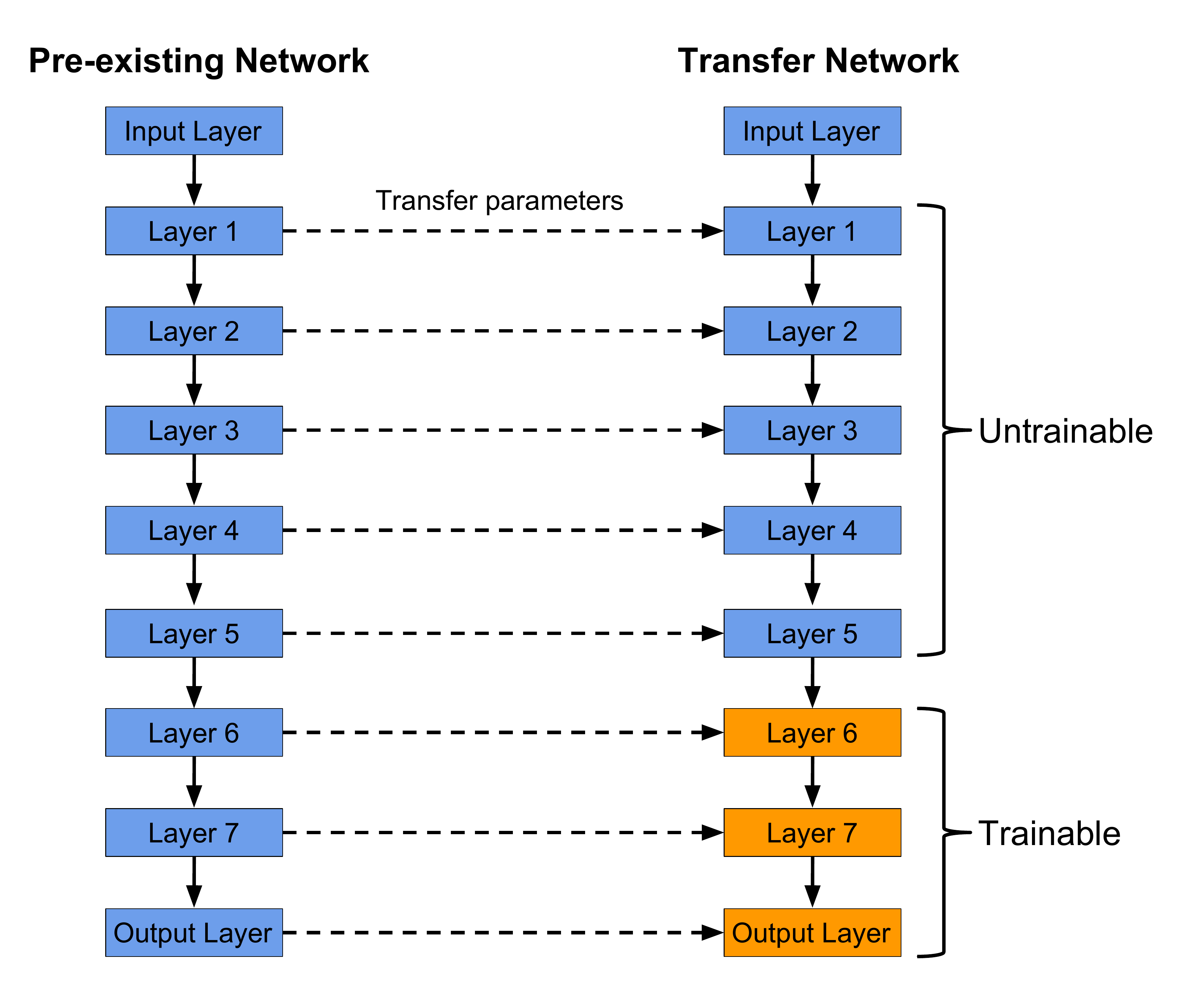}
	\caption{\textbf{Transfer learning with neural networks.} The parameters from the pre-existing network are copied to the transfer network. The transfer network has the same architecture as the pre-existing network. The first layers of the transfer network are non-trainable. The last layers of the transfer network are trainable.}
	\label{fig:transfer_learning}
\end{figure}

Transfer learning techniques are commonly used in machine learning to adapt an existing model to a new environment. We will compare our efforts against two well-known approaches from related work:

\textit{Freezing} follows the approach from Oquab et al.~\cite{Oquab2014} by merely retraining the last layers of the network. We call this method \textit{Freezing}, since the weights of some network layers are frozen (non-trainable). This is also sometimes known as "pre-training". 
\begin{description}[font=\normalfont]
\item[\textbf{\textit{Freezing}}] The base classifier is split into trainable and non-trainable parts as shown in Figure~\ref{fig:transfer_learning}. To compare this method to NN-Patching, all layers including the engagement layer are non-trainable. In contrast to NN-Patching, the initialization of the trainable layers is not at random, but adopting the weights from the base classifier.
\item[\textbf{\textit{Base}$\mathbf{_{\textit{update}}}$}] The whole base classifier is trained. All weights and parameters are trainable. This approach has the highest number of trainable parameters, hence the model capacity to represent concepts is also high. This approach can also be regarded as a special case of transfer learning, where all layer weights are trainable.
\end{description}

\pagebreak
\section{Optimizing Neural Network Patching}
\label{sec:networkarchitectureandengagement}
In this section, we investigate various approaches to leverage the performance of neural network patching. Both, engagement layer selection and patch architecture selection are important decisions for neural network patching, since they highly influence the model performance. 

In Section~\ref{Engagement Layer Selection} we discuss the engagement layer selection. The optimal engagement layer depends on specifics of the dataset and the architecture of the base classifier. After we obtained heuristics in order to select adequate engagement layers, we tackle the problem of optimizing the patch architecture in Section~\ref{Patch Architecture Selection}. Patches with more hidden layers have potentially higher capabilities to learn more complex representations and tasks. However, multi-layered patch networks will adapt slower to a new concept.

Moreover, we discuss \textit{inclusive} and \textit{exclusive} patch training. Inclusive training means that the patch is trained on all instances after the drift, whereas exclusive training implies that the patch is only trained on instances from the error region of the base classifier.
We obtain theoretical performance boundaries for them in Section~\ref{Theoretical Advantage of exclusive over inclusive Patch Network Training}.  The findings motivate a third patch training scheme called \textit{semi-exclusive} training. Finally, we discuss the differences and advantages of each training scheme in Section~\ref{sec:semiexclusivetraining}.

\subsection{Engagement Layer Selection}
\label{Engagement Layer Selection}
The optimal engagement layer depends on the base network architecture and the nature of the concept drift. Simple concept drifts, where the drift affects the resulting labels and not the input data, can be solved by using a layer close to the network output as the engagement layer for patching. Layers close to the network output tend to perform classification tasks, whereas early layers usually perform feature extraction. In contrast, complex drifts may require earlier engagement layers. 

In this section, we specifically discuss the influence of different network architectures on the engagement layer selection. Later, we finalize our findings by formulating heuristics, which act as a guideline on selecting a suitable engagement layer for each network archetype. 

The output of the engagement layer is the input for the patch network. Thus, the selection of the engagement layer impacts the performance of neural network patching. Choosing an engagement layer without useful features results in a low classification performance. The model used in the engagement layer selection experiments is 
a patching network that is trained on all instances, without the usual error estimator network.
After the concept drift all instances are diverted to the patch for classification. By this, we obtain an estimate of the maximum performance a patch can achieve, when attached to a specific layer.

\subsubsection{Network Architecture Dependence of Engagement Layers}
We want to select a suitable engagement layer for our patch, therefore we have to consider the architecture of the base classifier. For some architectures, well performing engagement layers are found in the higher layers of the network, close to the output layer, whereas for other network architectures it is preferable to choose engagement layers close to the network input. If we categorize the networks into three different archetypes, we can recognize essential similarities and differences. The distinguished archetypes are: Fully-Connected Network (FC-NN), Convolutional Network (CNN), and Residual Network (ResNet). 

\begin{figure}[H]
	\centering
		\subfigure[\label{fig:Fully-Connected Network engagement layer} Fully-Connected Network]{\includegraphics[scale=0.35]{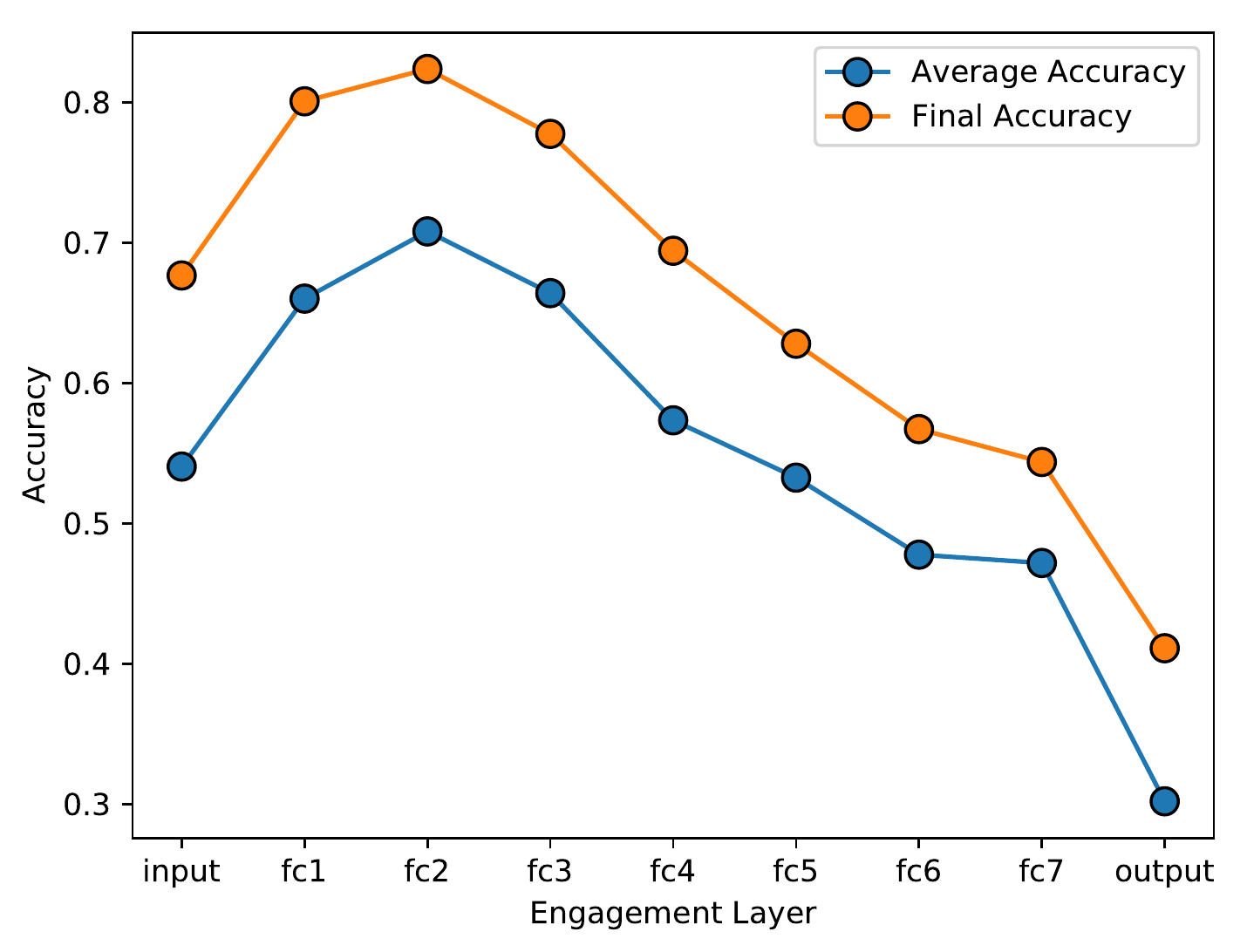}}
		\subfigure[\label{fig:Convolutional Network engagement layer} Convolutional Network]{\includegraphics[scale=0.35]{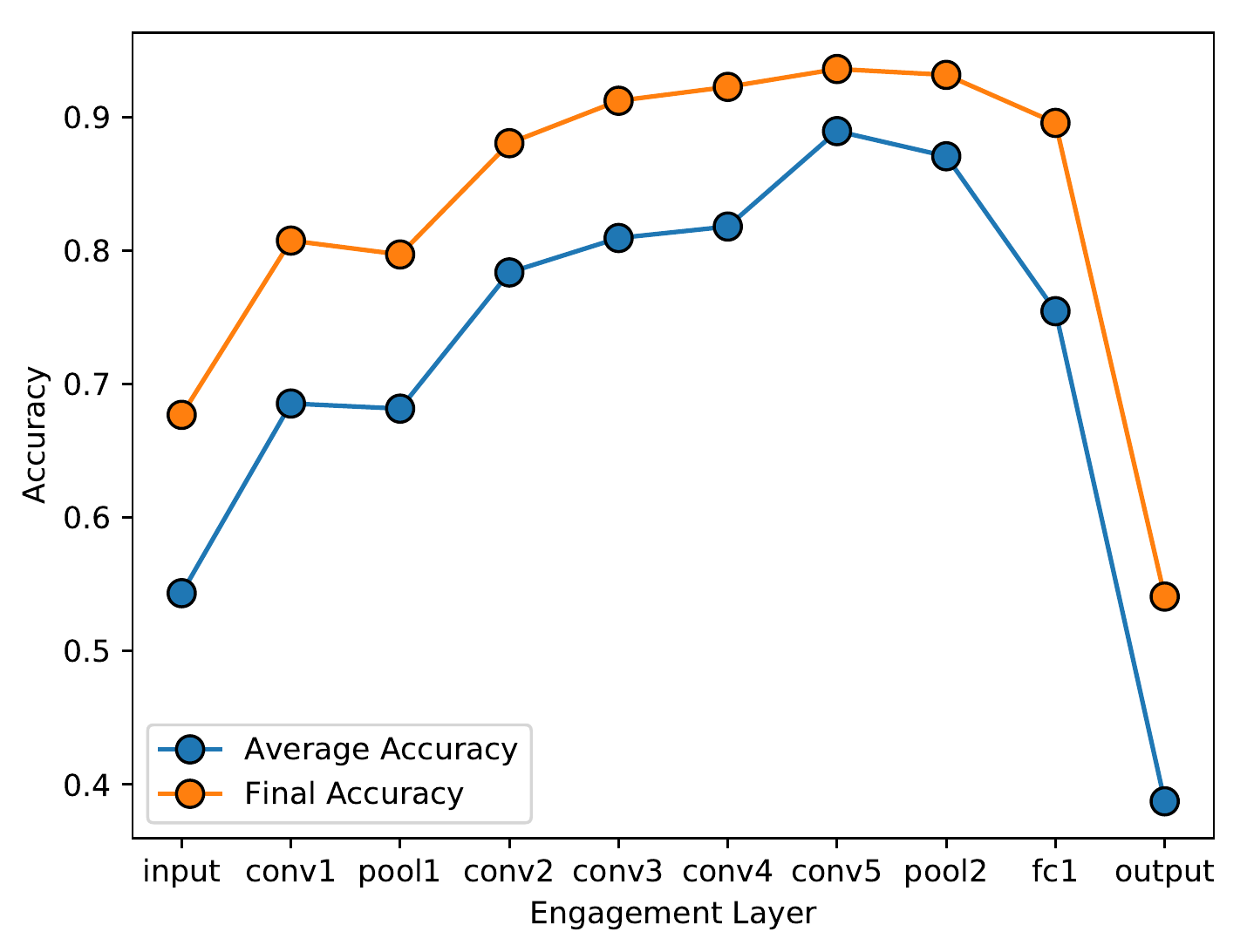}}\\
		\subfigure[\label{fig:Residual Network engagement layer} Residual Network]{\includegraphics[scale=0.35]{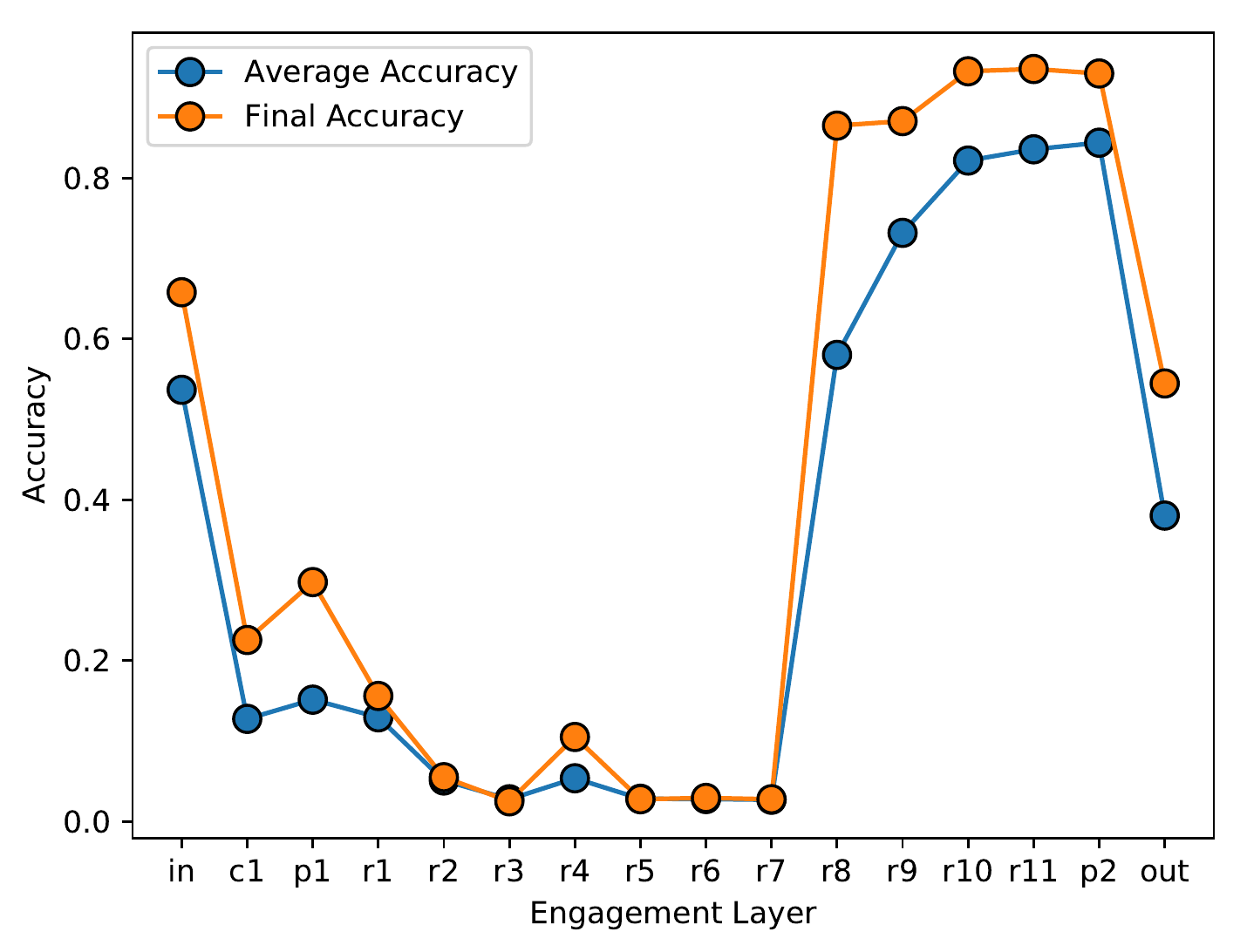}}
	\caption{\textbf{Patching performance by engagement layer for three different base classifier archetypes.} The used dataset is \textit{NIST}$_{flip}$ and the patch architecture consists of a single hidden layer with 128 units. The plots show the evaluation measures average accuracy and final accuracy. The evaluation measures are obtained by using the output of the engagement layer as the input for the patch network. The x-axis states the layer type of the engagement layer. The abbreviation 'fc' stands for fully-connected layer, 'conv' or 'c' for convolutional layer, 'pool' or 'p' for pooling layer, and 'r' for the output of a residual block. The left side of the x-axis starts with the input layer, which is the raw data. The last network layer is the output layer. The output layer is a softmax layer with the number of nodes equal to the amount of different classes in the dataset.}
	\label{fig:engagement_layer}
\end{figure}

In Figure~\ref{fig:engagement_layer} we show an engagement layer accuracy progression, which is representative for each network archetype. The results are the accuracy that can be achieved when attaching the patch to a certain layer. The layers are shown on the x-axis.
Flatten and Dropout layers are excluded from the evaluation for all archetypes, since they are equivalent to the layer before or do not have an effect on the forward pass at all (Dropout). 
For residual networks we only consider the output of each residual block in the performance evaluation. Hence, for residual networks we additionally exclude the parallel layers inside the residual blocks. The exact base classifier architectures are described in Section~\ref{Base Classifier Architectures}.   
\paragraph{Engagement Layers in Fully-Connected Networks.}
Figure~\ref{fig:Fully-Connected Network engagement layer} shows the layer-wise patching performance for a fully-connected network architecture. The optimal engagement layer in the presented configuration is the second fully-connected layer of the network. We observe that the average and final accuracy increase up to the second fully-connected layer. After this point, the accuracy decreases gradually. One reason of this effect could be, that the following fully-connected layers tend to perform classification tasks, instead of extracting transferable features. Yosinski et al.~(2014) describe this behaviour as \textit{general} versus \textit{specific}. The features in early layers tend to be general, whereas later layers consist of more specific features with respect to the classification task.

Furthermore, average and final accuracy show a strong correlation. The final accuracy is higher than the average accuracy, since the patch network has had more time to adapt to the new concept when the final phase begins (Fig.~\ref{fig:scenario}). But besides the accuracy offset, average and final accuracy are highly correlated. The optimal engagement layer for maximizing the average accuracy is usually also ideal with respect to the final accuracy. This property holds for all examined network archetypes.

\paragraph{Engagement Layers in Convolutional Networks.}
Figure~\ref{fig:Convolutional Network engagement layer} shows the patching performance based on engagement layer for a convolutional network. The graph shows a gradual increase in accuracy for layers further away from the input layer. The accuracy maximum is reached at the fifth convolutional layer as engagement layer for the patch. The following pooling layer shows a marginal loss in average and final accuracy. Moreover, we know that stacking convolutional layers generates a strong feature hierarchy~\cite{DBLP:journals/corr/ZeilerF13}. The graph indicates that in contrast to fully-connected layers, convolutional layers tend to extract transferable features. The last two network layers in the CNN are fully-connected layers. The graph shows a significant performance decrease for using these layers as engagement layer. Although it is intuitive that layers close to the network output perform classification, 
it is obvious in this case.
Therefore, fully-connected layers in CNNs seem not suitable as engagement layers for patching.

\paragraph{Engagement Layers in Residual Networks.}
Our residual network architecture consists of convolution, batch normalization, add, dropout and pooling layers. The only fully-connected layer in the network is the output layer. In contrast to the gradual feature extraction by the CNN, the patching performance for the last residual layers (Fig.~\ref{fig:Residual Network engagement layer}) suddenly increases in the ResNet. The residual blocks 'r1' to 'r7' are not suited as an engagement layer for patching. The patching accuracy of these layers is comparable to using noise, without any relation to the classification task, in order to train the patch. Although the output from the early residual blocks seem to not contain any useful information for classifying instances following the new concept, the last residual blocks of the network recover useful and transferable features. 
Since the transferable features are recovered from the poor performing residual block output, these blocks either contain useful information, or the information is recovered from the residual connections. 
This could be caused by the ResNet being over-specified for the given task, such that the earlier layers do not learn any useful latent features.
We observe this behaviour for our residual networks on all datasets. 


\subsubsection{On the Importance of the Activation Function}
\label{On the Importance of the Activation Function}
From Figure~\ref{fig:engagement_layer} we notice, that the first convolutional layer of the ResNet shows poor performance. This contradicts the assumption, that convolutional layers are good feature extractors. 

So far, we use the output of the engagement layer after applying the activation function. For consistency, we also use the output of the convolutional layer in the ResNet after the activation layer. Since the ResNet architecture from \cite{He2016} applies batch normalization before the ReLU activation, the patching accuracy is obtained after applying batch normalization and ReLU activation. In Table~\ref{tab:resnet_first_layers} the patching accuracies for these layers are shown in detail. 

\begin{table}[h]
\begin{centering}
\caption{\textbf{Comparison of the average patching accuracies for the first layers of the ResNet.} The table shows the patching accuracy decrease after applying batch normalization and ReLU activation. The used dataset is $\mathbf{\textit{NIST}_{\textit{flip}}}$ and the patch architecture is a single hidden layer with 128 units (the setting is similar to the experiment in Figure~\ref{fig:engagement_layer}).}
\label{tab:resnet_first_layers}

  \begin{tabular}{ | l | l | l |}
    \hline
		\#& Layer Type& Avg. Acc. \\  \hline
		 2 & Convolutional & 0.49  \\ 
		 3 & Batch Normalization &  0.22  \\ 
		 4 & ReLU Activation  & 0.09  \\ 
		 5 & Max Pooling & 0.15  \\ 
    \hline  
  \end{tabular}

\end{centering}
\end{table}    
When we use the convolutional layer before applying activation or batch normalization as engagement layer, we achieve an average accuracy of 49\%. After applying batch normalization, the accuracy for patching decreases to 22\%. If, additionally, ReLU activation is applied, the accuracy drops to 9\%. 

\begin{figure}[H]
	\centering
		\subfigure[\label{fig:Fully-Connected Network engagement layer relu_vs_none} Fully-Connected Network]{\includegraphics[scale=0.35]{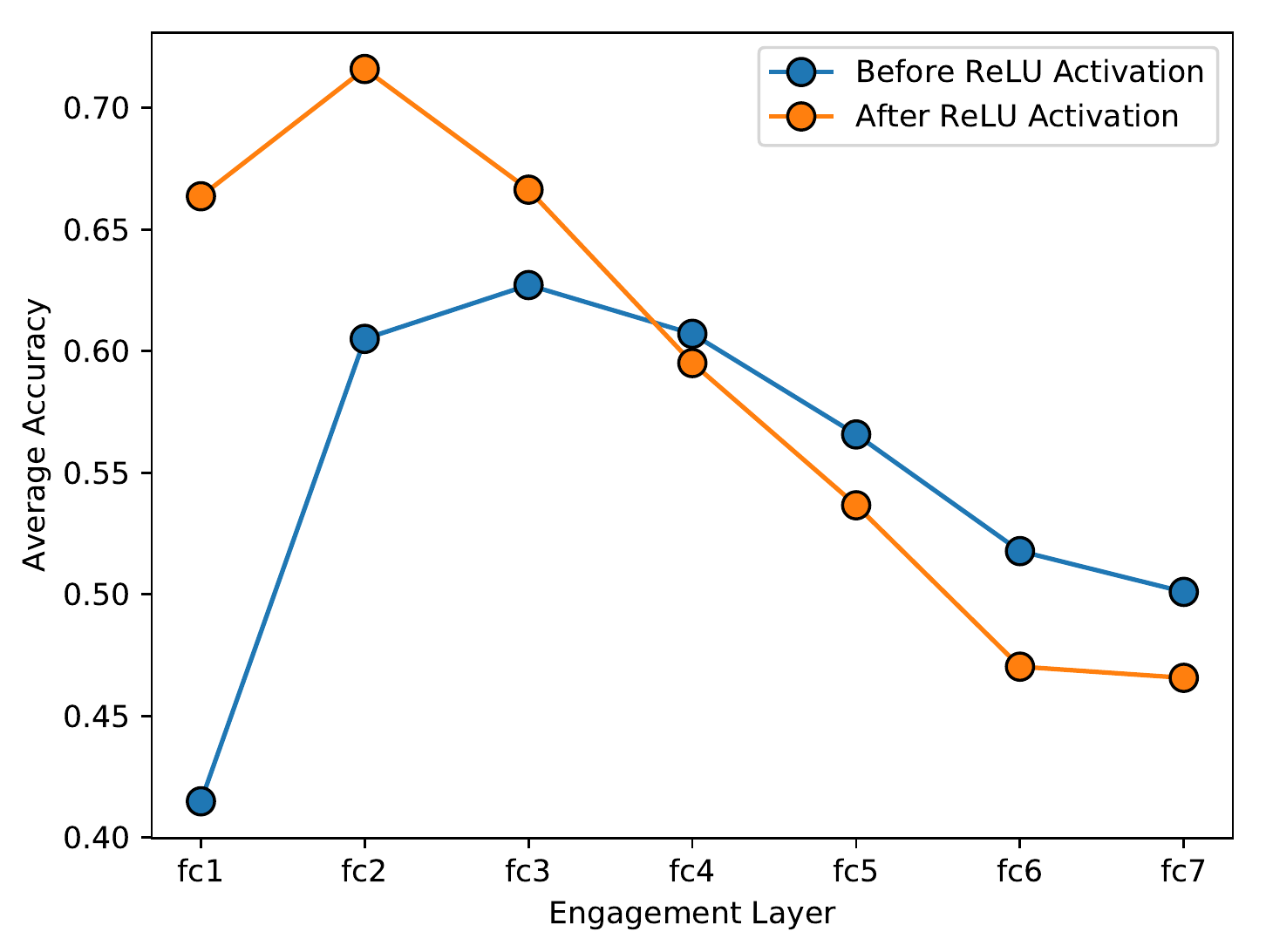}}
		\subfigure[\label{fig:Convolutional Network engagement layer relu_vs_none} Convolutional Network]{\includegraphics[scale=0.35]{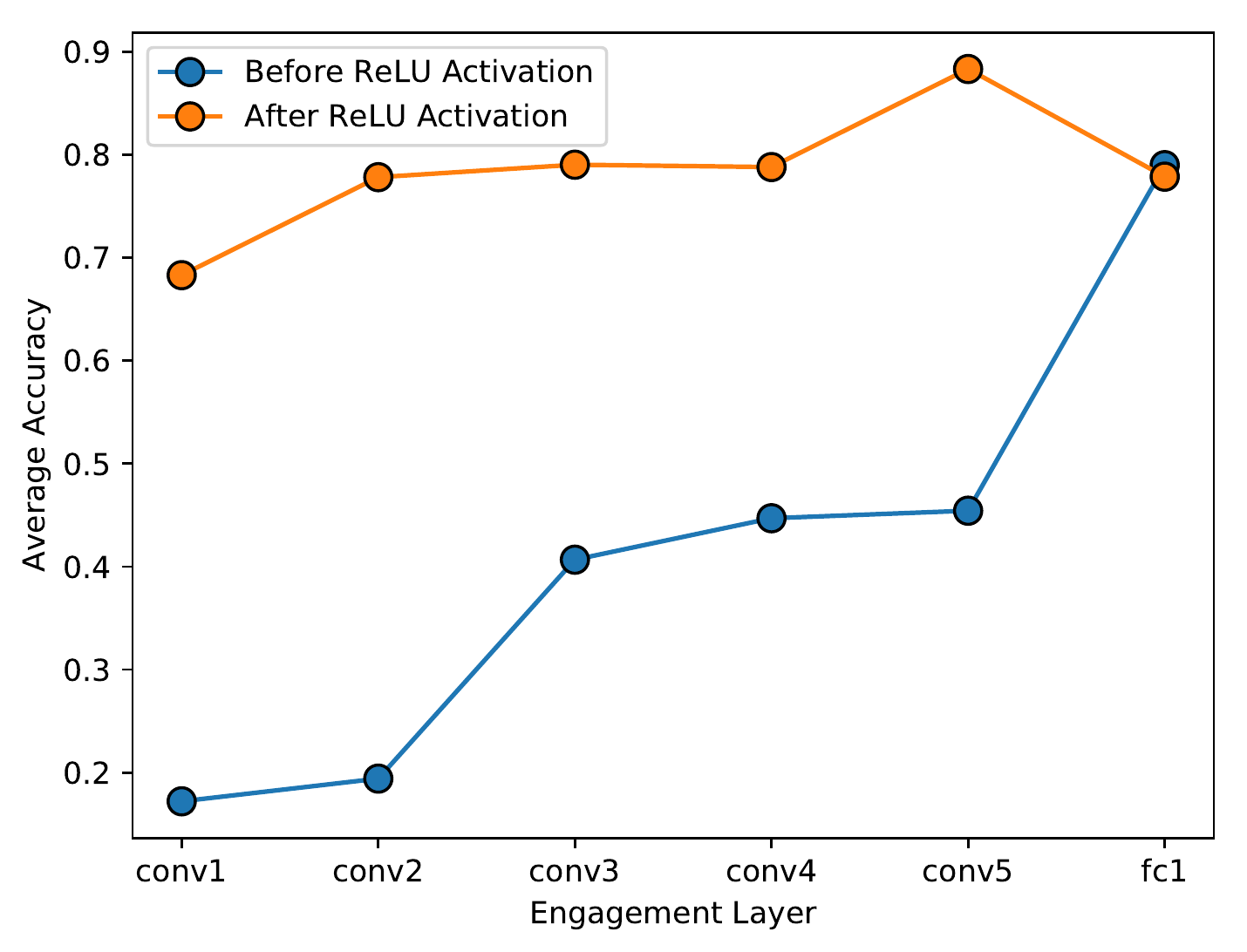}}\\
		\subfigure[\label{fig:Residual Network engagement layer relu_vs_none} Residual Network]{\includegraphics[scale=0.35]{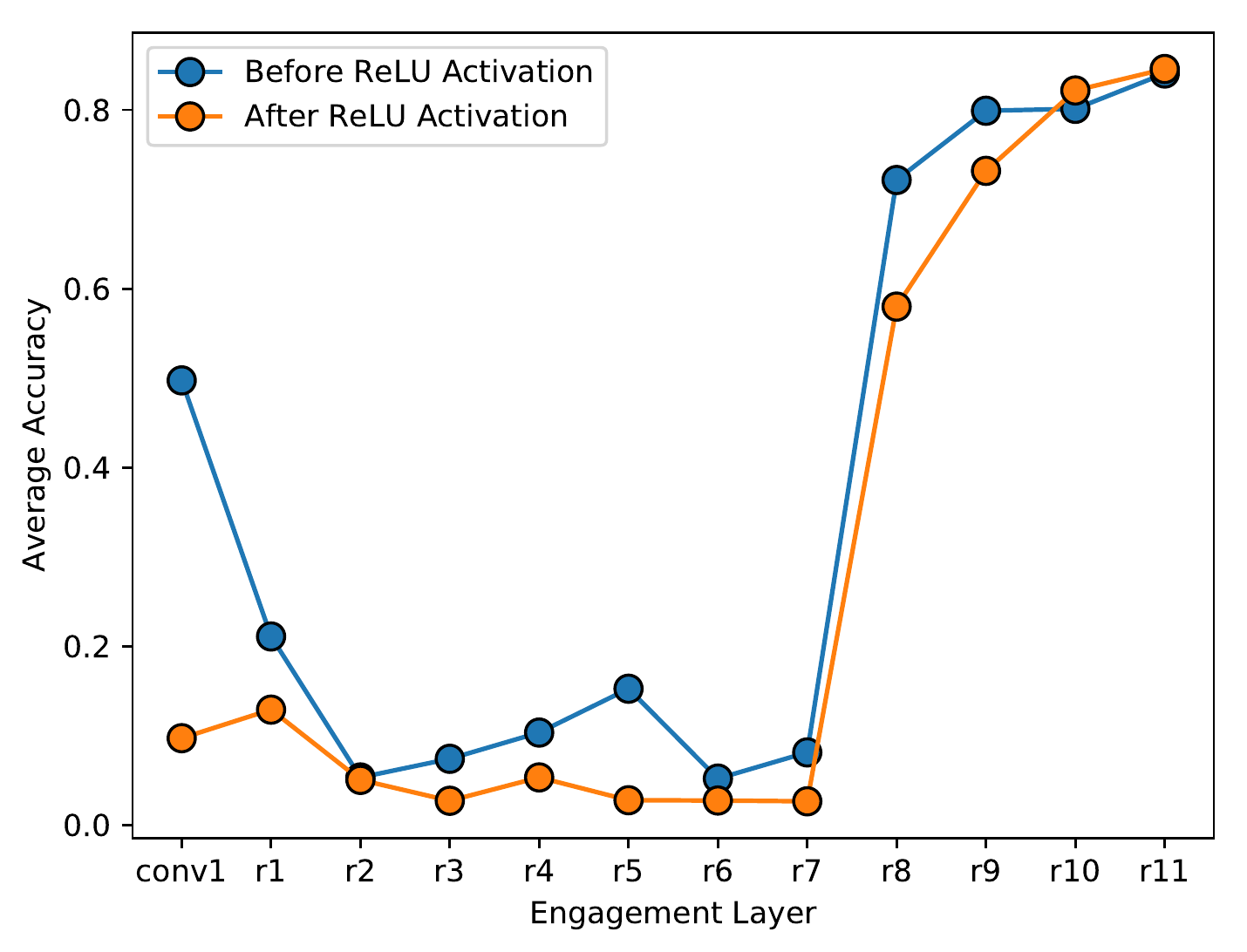}}
\caption{\textbf{Comparison of the average patching accuracies for each network archetype before and after application of ReLU activation.} The used dataset is $\mathbf{\textit{NIST}_{\textit{flip}}}$ and the patch architecture is 128xSoftmax (the setting is similar to the experiment leading to Figure~\ref{fig:engagement_layer}).}
\label{fig:relu_vs_none}
\end{figure}

In our experiments, the application of ReLU activation sometimes decreases the patching performance in comparison to the pure layer output (without applied activation). ReLU activation returns the identity for each value greater than zero and zero for every negative input value. If we consider these characteristics, it becomes clear that ReLU activation obliterates information contained in negative values. The patch can still use this information to achieve a better performance. This should be intuitive, since information, which is unusable for the original task, may be useful after the concept has changed. In Figure~\ref{fig:relu_vs_none} we present a comparison between using a layer as the engagement layer before and after applying the ReLU activation. 
 
In case of the FC-NN (Fig.~\ref{fig:Fully-Connected Network engagement layer relu_vs_none}) the patching accuracy obtained with the engagement layer after applying the activation is higher for the first network layers. After the fourth fully-connected layer, the accuracy from the raw layer output without activation surpasses it. 
In the CNN architecture (Fig.~\ref{fig:Convolutional Network engagement layer relu_vs_none}), it is always beneficial to apply the activation before using the engagement layer output for patching. Only for the fully-connected layer, which is the second to last layer in the CNN, both variants show comparable accuracy. 

With the ResNet (Fig.~\ref{fig:Residual Network engagement layer relu_vs_none})  it is beneficial to retrieve the engagement layer output before applying the activation for most layers, except the last two residual block outputs.

This comparison shows two effects of the ReLU activation on the information in a network layer. Sometimes the patching performance decreases after applying ReLU to an engagement layer output. In contrast, we also observe performance increase through applying the ReLU activation. Since the ReLU function maps every negative value to zero and returns the identity for non-negative values, the ReLU activation discards the information implied by negative values. For positive values exact activations remain, but for negative values only the information about the negative sign is preserved.

In order to explain the behavior shown in Figure~\ref{fig:relu_vs_none}, we recognize that applying the ReLU activation is only beneficial for layers with general features. For engagement layers which are already showing decreased patching performance due to the specificity of features, applying no activation function is beneficial. We propose, that solving discrimination tasks with neural networks consist of two phases, to which the network layers can be allocated: (i) the phase where feature extraction is conducted and (ii) the phase where classification tasks are performed. This definition is related to the characterization into general and specific layers by \cite{Yosinski2014}. Features may be more general in the early layers, but our experiments indicate that they are not necessarily features that are beneficial for the target task. 

The optimal engagement layer is the layer with features general enough to solve the drift task and specific enough to contain suitable high-level features related to the drift task. All layers before the optimal engagement layer are too general and all layers after are too specific with respect to the given drift task.

\paragraph{On Engagement Layers with general Features.}
For engagement layers with general features it is beneficial to apply the ReLU activation in order to increase patching accuracy. The ReLU activation discards information about unsuitable features, which are not beneficial for the original classification task. Since the dropped features are general, these features tend to be unsuited for the drift task as well.

\paragraph{On Engagement Layers with specific Features}
Contrarily, in engagement layers with specific features the discarded features may be relevant for the target task due to their specificity. High-level features, which are dispensable for the original concept, may still be useful for the drift task. Hence, applying ReLU activation yields a performance decrease for specific layers.

This explanation holds for the FC-NN and the CNN. To explain the performance decrease through ReLU activation in the ResNet, we propose that the early layers serve an information preserving purpose and neither work as feature extraction nor classification layers.


\paragraph{On the Effects of distractive Information.}
Another way to interpret the application of the ReLU function to the engagement
layer output is the effect of distractive information on neural networks. The engagement layers in the CNN (Fig.~\ref{fig:Convolutional Network engagement layer relu_vs_none}) show a huge performance increase by applying
the ReLU activation. We interpret it in such a way, that the ReLU function maps distractive information to zero. Since an activation of zero is similar to the absence of the
neuron, we conclude that removal of irrelevant information can achieve huge performance increases for neural networks.

\paragraph{Conclusion on the Activation Function.}
Concluding the discussion on using the engagement layer output before or after the ReLU activation, we recognize that, in the experiments we conducted, \textbf{the highest performing engagement layer always has the ReLU activation applied}. Hence, we use the engagement layer output after activation as the input for the patch network in further experiments.

 
\subsubsection{Dataset Dependence of Engagement Layers}
\label{Dataset Dependence}
In the previous section, we concluded that the optimal engagement layer is highly dependent on the base classifier archetype. In this section, we investigate the engagement
layer dependence on the dataset. Hence, we conducted test series for every dataset and base classifier archetype. The results are presented in Table~\ref{tab:Best engagement layer by dataset and classifier}.

The best engagement layer for the FC-NN is always the first or second fully-connected
layer of the base classifier. For the CNN, the best engagement layer is either the last
pooling layer or the last convolutional layer. The best engagement layer for the ResNet
follows a similar pattern: Either the output of the last residual block or the average pooling layer show highest patching accuracy. The ResNet architecture on $\mathbf{\textit{MNIST}_{\textit{appear}}}$ shows the best final accuracy for the second last residual block. We consider this a coincidence, since the patching performance of the last and second to last residual block
hardly differs in this specific case.

We consider $\mathbf{\textit{NIST}_{\textit{remap}}}$ and $\mathbf{\textit{NIST}_{\textit{transfer}}}$ the datasets with the most difficult concept drifts, since the base classifier is trained on numbers and has to adapt to letters. 

\begin{table}[h]
\begin{centering}

\caption{\textbf{Best engagement layer by dataset and classifier.} The optimal engagement layer is given for the evaluation measures average accuracy and final accuracy. We state the name of the optimal engagement layer and position in the base classifier network. The position in the base classifier network is shown in parentheses (e.g. (3.last) refers to the third to last layer of the base network). The dataset we used is $\mathbf{\textit{NIST}_{\textit{flip}}}$ and the patch had a single hidden layer with 128 units.}
\label{tab:Best engagement layer by dataset and classifier}

\resizebox{\columnwidth}{!}{%
  \begin{tabular}{ | l | l l | l l | l l |}
    \hline
		 Base Archetype:  &  \multicolumn{2}{ c|}{Fully-Connected} &  \multicolumn{2}{ c|}{Convolutional} & \multicolumn{2}{ c|}{ Residual } \\ \hline
		Best Layer by:   & Avg. Acc.& F. Acc.& Avg. Acc.& F. Acc.& Avg. Acc.& F. Acc. \\ \hline
		$\mathbf{\textit{MNIST}_{\textit{appear}}}$ & FC1(1.)      &  FC1(1.)     & Pool1(3.last)    & Pool1(3.last)     & Pool1(2.last)     & R11(4.last)           \\ 
		$\mathbf{\textit{MNIST}_{\textit{flip}}}$ & FC1(1.)       &  FC1(1.)       & Pool1(3.last)    & Pool1(3.last)     & R12(3.last)      & R12(3.last)           \\ 
		$\mathbf{\textit{MNIST}_{\textit{remap}}}$  & FC1(1.)       &  FC1(1.)       & Pool1(3.last)    & Conv2(4.last)     & R12(3.last)      & R12(3.last)     \\
		$\mathbf{\textit{MNIST}_{\textit{rotate}}}$ & FC1(1.)       &  FC1(1.)       & Conv2(4.last)    & Pool1(3.last)     & R12(3.last)      & Pool1(2.last)    \\ 
	 $\mathbf{\textit{MNIST}_{\textit{transfer}}}$& FC1(1.)       &  FC1(1.)       & Conv2(4.last)    & Conv2(4.last)     & Pool1(2.last)    & R12(3.last)     \\ \hline
		$\mathbf{\textit{NIST}_{\textit{appear}}}$  & FC2(2.)       &  FC2(2.)      & Conv5(4.last)    & Conv5(4.last)     & Pool2(2.last)    & R11(3.last)   \\
    $\mathbf{\textit{NIST}_{\textit{flip}}}$  & FC2(2.)     &  FC2(2.)      & Conv5(4.last)    & Conv5(4.last)     & Pool2(2.last)    & R11(3.last)   \\ 
		$\mathbf{\textit{NIST}_{\textit{remap}}}$   & FC1(1.)       &  FC1(1.)       & Conv5(4.last)    & Conv5(4.last)     & R11(3.last)      & R11(3.last)   \\
		$\mathbf{\textit{NIST}_{\textit{rotate}}}$  & FC2(2.)      &  FC2(2.)      & Conv5(4.last)    & Pool2(3.last)     & R11(3.last)      & R11(3.last)  \\ 
		$\mathbf{\textit{NIST}_{\textit{transfer}}}$& FC1(1.)       &  FC1(1.)       & Conv5(4.last)    & Conv5(4.last)     & R11(3.last)      & R11(3.last)   \\
    \hline  
  \end{tabular}}

\end{centering}
\end{table}


This indicates that more complex concept drifts tend to be solved with the information of earlier engagement layers, whereas for moderate drifts the ideal engagement
layer tends to be later in the network. The second fully connected layer appears to be too
specific for the target tasks in $\mathbf{\textit{NIST}_{\textit{remap}}}$ and $\mathbf{\textit{NIST}_{\textit{transfer}}}$. This is in line with our previous observations regarding the generality vs. specificity dilemma of fully-connected
networks.

However, this behaviour was not observed with the other two base classifier architectures. We suggest that this phenomenon still occurs. But since convolutional layers generate fairly general features it does not show in this case.
The observed difference in generality and specificity is huge when we compare the
last convolutional layer (the residual block also consists of convolutions) and the first
fully-connected layer of both the CNN and the ResNet. Fully-connected layers in the
CNN and ResNet are apparently too specific to deal with all the different types of drift.

Between the last convolution and the fully-connected layer is a pooling layer in both
the ResNet and the CNN architecture. The pooling layer is apparently more specific
than the convolutional layer. For some datasets the pooling layer is the best engagement
layer, but never for $\mathbf{\textit{NIST}_{\textit{remap}}}$ and $\mathbf{\textit{NIST}_{\textit{transfer}}}$.

The observed property of fully-connected layers to perform specific classification
tasks and the generality of convolutional layers lead to a strong division between general
and specific sections in a neural networks. We can use this property to make a robust
selection regarding suitable engagement layers for patching. For every different base
classifier architecture, only two layers qualify for the highest performing engagement
layer across all datasets.

\subsubsection{Heuristics for Engagement Layer Selection}
\label{Rules of Thumb on Engagement Layer Selection}
After we investigated the dependencies of engagement layer selection, we
want to formulate a heuristic rule for each network archetype, indicating suitable engagement layers for patching. In order to do this, we
consider the main findings of the previous sections.

%
%
%

After considering the findings of the previous sections, we state the following heuristic rules for engagement layer selection:
\begin{description}[font=\normalfont]
\item[\textbf{Fully-Connected Neural Network:}] The best engagement layer is either the first or second fully-connected layer in the network.
\item[\textbf{Convolutional Neural Network:}] The best engagement layer is either the last convolutional layer or the last pooling layer of the network.
\item[\textbf{Residual Neural Network:}] The best engagement layer is either the output of the last residual block or the last pooling layer of the network.
\end{description}
Selecting the best engagement layer is important, since we observe a significant performance difference depending on the engagement layer.
These heuristics
narrow down the search space for the optimal engagement layer to two layers. Best practice is to try both candidate layers.

\newpage
\subsection{Patch Architecture Selection}
\label{Patch Architecture Selection}

In this section, we investigate the influence of different patch architectures on the patching performance. We evaluate 25 different patch architectures on all datasets for FC-NN, CNN and ResNet. Each experiment was conducted five times with varying random
seeds. All presented values are averaged over these five runs. We exclude all engagement layers except the two most promising layers from the previous sections. The two
candidate engagement layers are selected by applying our selection heuristics for engagement layers (Sec.~\ref{Rules of Thumb on Engagement Layer Selection}). The 25 patch architectures have between one and three
hidden layers. Only fully-connected layers are used building the patch. If the engagement layer output is multidimensional, the first layer of the patch network is a Flatten
layer. 

The patch architectures are presented without explicitly indicating the softmax classification layer as the last layer of every patch, since the presence of an output layer is mandatory. Therefore, '256x128' refers to a patch architecture with the following consecutive layers: 
\begin{verbatim}
Input() - FC(256) - FC(128) - Softmax(num_classes). 
\end{verbatim}

The architecture '128' refers to: 
\begin{verbatim}
Input() - FC(128) - Softmax(num_classes).
\end{verbatim}

The model used in the patch architecture experiments is 
a patching model that does not learn a error estimator, learns from all arriving instances after the concept drift, and diverts all instances to the patch for classification.

%

\subsubsection{Ideal Engagement Layer and Patch Architecture Combination}

In Table~\ref{tab:Best engagement layer and patch architecture by average accuracy for each dataset and base archetype} we show the best engagement layer/patch architecture combination with respect to maximizing average accuracy. All patch architectures, which are maximizing the average accuracy, consist of one fully-connected layer and a softmax classification layer. The nature of the dataset (i.e. the inherent concept drift) has an influence on the engagement layer. As predicted by our heuristic rules for engagement layer selection, it is not possible to select a single perfect engagement layer across all datasets without considering the nature of the concept drift.
\begin{table}[h]
\begin{centering}
\caption{\textbf{Ideal engagement layer and patch architecture combination by average accuracy.} The table shows the best engagement layer/patch architecture combination for each dataset and classifier with respect to the average accuracy.}
\label{tab:Best engagement layer and patch architecture by average accuracy for each dataset and base archetype}
  \begin{tabular}{ | l |  l  l | l  l | l  l | }
    \hline
	Archetype:	&\multicolumn{2}{ c|}{FC-NN} &  \multicolumn{2}{ c|}{CNN} & \multicolumn{2}{ c|}{ResNet}\\ \hline
		   Dataset&Layer& Patch Arch.&Layer& Patch Arch.&Layer& Patch Arch.\\ \hline
$\mathbf{\textit{MNIST}_{\textit{appear}}}$ & fc1 & 2048 & conv2 & 256 & p1 & 128\\
$\mathbf{\textit{MNIST}_{\textit{flip}}}$ & fc1 & 2048 & pool1 & 256 & r12 & 128\\
$\mathbf{\textit{MNIST}_{\textit{remap}}}$ & fc1 & 1536 & pool1 & 256 & r12 & 128\\
$\mathbf{\textit{MNIST}_{\textit{rotate}}}$ & fc1 & 2048 & pool1 & 1536 & r12 & 256\\
$\mathbf{\textit{MNIST}_{\textit{transfer}}}$ & fc1 & 2048 & conv2 & 256 & p1 & 128\\ \hline
$\mathbf{\textit{NIST}_{\textit{appear}}}$ & fc2 & 2048 & conv5 & 2048 & p2 & 2048\\
$\mathbf{\textit{NIST}_{\textit{flip}}}$ & fc2 & 2048 & conv5 & 1536 & p2 & 1536\\
$\mathbf{\textit{NIST}_{\textit{remap}}}$ & fc1 & 1024 & conv5 & 1024 & r11 & 128\\
$\mathbf{\textit{NIST}_{\textit{rotate}}}$ & fc2 & 2048 & conv5 & 1536 & p2 & 1024\\
$\mathbf{\textit{NIST}_{\textit{transfer}}}$ & fc1 & 1536 & conv5 & 512 & p2 & 1024\\
    \hline  
  \end{tabular}

\end{centering}
\end{table}

\begin{table}[h]
\begin{centering}
\caption{\textbf{Ideal engagement layer and patch architecture combination by final accuracy.} The table shows the best engagement layer/patch architecture combination for each dataset and classifier with respect to the final accuracy.}
\label{tab:Best engagement layer and patch architecture by final accuracy for each dataset and base archetype}

  \begin{tabular}{ | l |  l  l | l  l | l  l | }
    \hline
	Archetype:	&\multicolumn{2}{ c|}{FC-NN} &  \multicolumn{2}{ c|}{CNN} & \multicolumn{2}{ c|}{ResNet}\\ \hline
		   Dataset&Layer& Patch Arch.&Layer& Patch Arch.&Layer& Patch Arch.\\ \hline
$\mathbf{\textit{MNIST}_{\textit{appear}}}$ & fc1 & 1024x512 & pool1 & 256 & r12 & 256\\
$\mathbf{\textit{MNIST}_{\textit{flip}}}$ & fc1 & 2048x512x256 & pool1 & 1024 & r12 & 128\\
$\mathbf{\textit{MNIST}_{\textit{remap}}}$ & fc1 & 1536x256 & conv2 & 128 & r12 & 128\\
$\mathbf{\textit{MNIST}_{\textit{rotate}}}$ & fc1 & 128 & pool1 & 2048 & r12 & 512x256\\
$\mathbf{\textit{MNIST}_{\textit{transfer}}}$ & fc1 & 1024 & conv2 & 256x128 & r12 & 256x128\\ \hline
$\mathbf{\textit{NIST}_{\textit{appear}}}$ & fc2 & 2048 & conv5 & 512 & p2 & 128\\
$\mathbf{\textit{NIST}_{\textit{flip}}}$ & fc1 & 2048 & conv5 & 512 & r11 & 1024\\
$\mathbf{\textit{NIST}_{\textit{remap}}}$ & fc1 & 1536 & conv5 & 256 & r11 & 256x128\\
$\mathbf{\textit{NIST}_{\textit{rotate}}}$ & fc2 & 2048 & conv5 & 1536 & r11 & 1536\\
$\mathbf{\textit{NIST}_{\textit{transfer}}}$ & fc1 & 1536x512 & conv5 & 1024 & r11 & 256\\
    \hline  
  \end{tabular}

\end{centering}
\end{table}

If we consider the best patch architecture with respect to maximizing the final accuracy instead of average accuracy (Tab.~\ref{tab:Best engagement layer and patch architecture by final accuracy for each dataset and base archetype}), occasionally deeper patch architectures
with two hidden layers achieve highest performance. Deeper architectures require more
training to converge opposed to shallow architectures.
The average accuracy is obtained as the average accuracy of all batches after the concept drift. In comparison, the final accuracy is obtained by averaging the accuracy of
the patch network on the last five batches of the data stream. The amount of 
training data available is larger for final accuracy. It is expected that deeper patch architectures perform better for final accuracy than for average accuracy, since the patch
network receives more training.  

Moreover, we observe that, in comparison to average accuracy, final accuracy is
more often higher with the earlier, more general layer of the two candidate layers (Tab.~\ref{tab:Number of shallow and deep layers acting as the best engagement layer for different evaluation measures}). 

\begin{table}[h]
\begin{centering}

\caption{\textbf{Ideal engagement layer and patch architecture combination by recovery speed.} The table shows the best engagement layer/patch architecture combination for each dataset and classifier with respect to the recovery speed.}
\label{tab:Best engagement layer and patch architecture by recovery speed for each dataset and base archetype}

  \begin{tabular}{ | l |  l  l | l  l | l  l | }
    \hline
	Archetype:	&\multicolumn{2}{ c|}{FC-NN} &  \multicolumn{2}{ c|}{CNN} & \multicolumn{2}{ c|}{ResNet}\\ \hline
		   Dataset&Layer& Patch Arch.&Layer& Patch Arch.&Layer& Patch Arch.\\ \hline
$\mathbf{\textit{MNIST}_{\textit{appear}}}$ & fc1 & 2048 & pool1 & 2048 & p1 & 128\\
$\mathbf{\textit{MNIST}_{\textit{flip}}}$ & fc1 & 1024 & conv2 & 256 & p1 & 1024\\
$\mathbf{\textit{MNIST}_{\textit{remap}}}$ & fc1 & 2048 & pool1 & 1024 & p1 & 1024\\
$\mathbf{\textit{MNIST}_{\textit{rotate}}}$ & fc2 & 1024 & pool1 & 2048 & r12 & 128\\
$\mathbf{\textit{MNIST}_{\textit{transfer}}}$ & fc1 & 1536 & pool1 & 1536 & p1 & 512\\ \hline
$\mathbf{\textit{NIST}_{\textit{appear}}}$ & fc2 & 1024 & conv5 & 1024 & p2 & 2048\\
$\mathbf{\textit{NIST}_{\textit{flip}}}$ & fc2 & 1024 & conv5 & 2048 & p2 & 1024\\
$\mathbf{\textit{NIST}_{\textit{remap}}}$ & fc1 & 1024 & conv5 & 1536 & p2 & 256\\
$\mathbf{\textit{NIST}_{\textit{rotate}}}$ & fc2 & 2048 & pool2 & 1536 & p2 & 2048\\
$\mathbf{\textit{NIST}_{\textit{transfer}}}$ & fc1 & 1024 & conv5 & 1024 & r11 & 1024\\
    \hline  
  \end{tabular}

\end{centering}
\end{table}

The third evaluation measure we examine is recovery speed. Recovery speed states
the amount of batches, and therefore the amount of training required to recover to 90\%
of the base classifier accuracy before the concept drift. Recovery speed is optimized, if
the model performs well on the first batches of the data stream right after the drift. Thus, fast adaptation to
the new concept is demanded. Final accuracy is optimized if the model performs well
on the last batches of the data stream. Hence, the ability of the model to represent the
new concept arbitrarily well is required to optimize final accuracy. Sometimes this is
achieved by more complex (deeper) models, but we only occasionally observe this in
our experiments. Average accuracy can be interpreted as a trade-off between recovery
speed and final accuracy, since the model performance on all batches after the concept
drift are considered obtaining this evaluation measure.

The best engagement layer and patch architecture with respect to recovery speed
for each dataset and base archetype is shown in Table~\ref{tab:Best engagement layer and patch architecture by recovery speed for each dataset and base archetype}. Similar to average accuracy,
all patch architectures optimizing the recovery speed are architectures with one hidden
layer. Shallow network architectures converge faster than deeper networks, therefore
they are well suited for quick adaptation to new concepts. In comparison to final accuracy, more often the latter of the two engagement layer candidates optimizes the
recovery speed (Tab.~\ref{tab:Number of shallow and deep layers acting as the best engagement layer for different evaluation measures}).

\begin{table}[h]
\begin{centering}
\caption{\textbf{Number of early and late layers acting as the best engagement layer for different
evaluation measures.} The table shows the number of times the earlier and later engagement
layer, of the two evaluated candidate layers, are the best engagement layer. We have three base classifier archetypes and ten different datasets, hence each column adds up to $3\cdot{10} =30$. This table can be reproduced with the contents of Table~\ref{tab:Best engagement layer and patch architecture by average accuracy for each dataset and base archetype}-~\ref{tab:Best engagement layer and patch architecture by recovery speed for each dataset and base archetype}.}
\label{tab:Number of shallow and deep layers acting as the best engagement layer for different evaluation measures}

  \begin{tabular}{ | l | l | l | l | }
    \hline
	Engagement Layer&\multicolumn{1}{ c|}{Rec. Speed} &  \multicolumn{1}{ c|}{Avg. Acc} & \multicolumn{1}{ c|}{Final Acc.}\\ \hline
		   Earlier Layer & 13 & 18 & 24\\
			 Later Layer    & 17 & 12 &  6\\
    \hline  
  \end{tabular}

\end{centering}
\end{table} 

The difference between the two candidate layers in the CNN and ResNet is the
pooling operation (e.g. max-pooling for CNNs, average-pooling for ResNets). The aggregated information in the pooling layers tends to be better suited for fast adaptation,
whereas the more comprehensive features from the previous layer results in a better
performance in final accuracy.

\subsubsection{Performance Differences between Patch Architectures}

In Table~\ref{tab:Various patch architectures sorted by average accuracy, final accuracy and recovery speed} we show the 25 patch architectures sorted by average accuracy, final
accuracy and recovery speed. We notice that the shallow architectures with one hidden
layer perform best on average for all evaluation measures. In average accuracy and recovery speed we notice a strong performance separation by patch depth. The maximum
difference in average accuracy between highest and lowest performing of the six architectures with one hidden layer is 0.89\,\%,, whereas the performance decrease between
the lowest performing architecture with a single hidden layer and the best performing
layer with two hidden layers is 1.43\,\%,. 

\begin{table}[h]
\begin{centering}
\caption{\textbf{Various patch architectures sorted by average accuracy, final accuracy and recovery speed.} For each evaluation measure we state the patch architecture and according accuracy (or batches to recover). The patch architectures are sorted by the evaluation measure. Architecture with the best performance comes first. The accuracy is shown in percent. The values are averaged over five runs with different random seed. The dataset we used is $\mathbf{\textit{NIST}_{\textit{flip}}}$,the base classifier is a CNN and the engagement layer is Conv5.}
\label{tab:Various patch architectures sorted by average accuracy, final accuracy and recovery speed}

\smaller

  \begin{tabular}{ |  l  l | l  l | l  l | }
    \hline
		\multicolumn{6}{ |c|}{Dataset: $\mathbf{\textit{NIST}_{\textit{flip}}}$\hspace{1cm} Classifier: CNN \hspace{1cm} Engagement Layer: Conv5}\\ \hline
		\multicolumn{2}{ |c|}{Average Accuracy} &  \multicolumn{2}{ c|}{Final Accuracy} & \multicolumn{2}{ c|}{ Recovery Speed }\\ \hline
		   Patch Architecture& Accuracy& Patch Architecture& Accuracy&Patch Architecture& Batches\\ \hline
1536 & 89.35 & 512 & 93.88 & 2048 & 8.4\\
1024 & 89.34 & 1024 & 93.88 & 1536 & 9.4\\
2048 & 89.33 & 2048 & 93.82 & 256 & 9.4\\
512 & 89.29 & 1536 & 93.76 & 512 & 9.4\\
256 & 89.03 & 256 & 93.65 & 1024 & 9.8\\
128 & 88.46 & 128 & 93.55 & 128 & 10.2\\
1024x512 & 87.03 & 2048x256 & 93.49 & 1536x256 & 11.2\\
1536x512 & 87.02 & 1536x512 & 93.4 & 2048x256 & 11.6\\
2048x512 & 86.99 & 1024x512 & 93.35 & 512x128 & 12.0\\
2048x256 & 86.93 & 512x128 & 93.35 & 1536x512 & 12.2\\
1024x256 & 86.91 & 256x128 & 93.32 & 2048x512 & 12.2\\
1536x256 & 86.91 & 2048x512 & 93.27 & 1024x256 & 12.4\\
512x256 & 86.62 & 1536x256 & 93.24 & 256x128 & 12.8\\
512x128 & 86.41 & 1024x256 & 93.13 & 1024x512 & 13.4\\
256x128 & 85.88 & 512x256 & 93.11 & 512x256 & 13.8\\
2048x512x256 & 83.57 & 1024x512x256 & 92.97 & 1536x256x128 & 16.0\\
1024x512x256 & 83.41 & 512x256x128 & 92.78 & 1536x512x128 & 16.0\\
1536x512x256 & 83.38 & 1536x256x128 & 92.53 & 1024x256x128 & 16.2\\
1024x256x128 & 82.88 & 2048x512x256 & 92.51 & 2048x256x128 & 16.6\\
1024x512x128 & 82.79 & 1536x512x128 & 92.45 & 1024x512x256 & 17.0\\
1536x512x128 & 82.66 & 2048x512x128 & 92.43 & 1536x512x256 & 17.0\\
2048x256x128 & 82.64 & 1024x256x128 & 92.38 & 2048x512x128 & 17.4\\
512x256x128 & 82.6 & 1024x512x128 & 92.24 & 2048x512x256 & 17.4\\
1536x256x128 & 82.53 & 1536x512x256 & 92.0 & 512x256x128 & 17.8\\
2048x512x128 & 82.51 & 2048x256x128 & 91.95 & 1024x512x128 & 18.2\\
    \hline  
  \end{tabular}

\end{centering}
\end{table}

In contrast, these large performance steps are not observed for final accuracy. For
final accuracy, the performance discrepancy between architectures of different depth are
less significant. The architectures are still listed by depth, but the transition is fluent.

Moreover, the performance difference between one-hidden-layer architectures '512', '1024', '1536', and '2048' in average and final accuracy is negligible. 

\begin{table}[h]
\smaller
\begin{centering}
\caption{\textbf{Patch architectures ranked by evaluation measures.} The table shows the average rank of each patch architecture for average accuracy, final accuracy and recovery speed. The rank is calculated as the average rank over all datasets, classifiers and both candidate layers. Each configuration is tested for five times with varying random seeds. The ranks are averaged over these five runs.}
\label{tab:Patch architectures ranked by evaluation measures}

  \begin{tabular}{ |  l  l | l  l | l  l | }
    \hline
		\multicolumn{2}{ |c|}{Average Accuracy} &  \multicolumn{2}{ c|}{Final Accuracy} & \multicolumn{2}{ c|}{ Recovery Speed }\\ \hline
		   Patch Architecture& Avg. Rank& Patch Architecture&  Avg. Rank&Patch Architecture&  Avg. Rank\\ \hline
1024 & 4.07 & 2048 & 5.57 & 1024 & 4.43\\
512 & 4.1 & 1024 & 5.92 & 1536 & 6.37\\
2048 & 4.25 & 512 & 6.28 & 2048 & 6.6\\
1536 & 4.28 & 1536 & 6.42 & 512 & 7.83\\
256 & 5.4 & 256 & 7.78 & 256 & 7.92\\
128 & 7.83 & 1024x512 & 9.87 & 128 & 8.6\\
512x256 & 9.6 & 128 & 10.08 & 1024x512 & 9.68\\
1024x512 & 10.33 & 1024x256 & 10.17 & 1024x256 & 9.75\\
1024x256 & 10.4 & 2048x256 & 10.33 & 1536x256 & 10.58\\
512x128 & 10.6 & 1536x256 & 10.55 & 1536x512 & 10.92\\
1536x256 & 10.95 & 1536x512 & 10.65 & 512x256 & 12.17\\
1536x512 & 11.0 & 512x256 & 10.65 & 2048x256 & 12.18\\
256x128 & 11.28 & 512x128 & 10.67 & 2048x512 & 12.47\\
2048x512 & 11.75 & 2048x512 & 11.2 & 512x128 & 12.9\\
2048x256 & 11.8 & 256x128 & 11.92 & 256x128 & 13.7\\
1024x512x256 & 18.2 & 1024x256x128 & 17.65 & 1024x256x128 & 15.77\\
512x256x128 & 18.48 & 1536x512x128 & 18.12 & 1024x512x256 & 16.03\\
1024x256x128 & 18.77 & 1024x512x256 & 18.42 & 1536x512x128 & 16.77\\
1536x512x256 & 18.9 & 1536x512x256 & 18.43 & 1024x512x128 & 17.18\\
1536x256x128 & 19.82 & 2048x512x128 & 18.47 & 1536x256x128 & 17.68\\
1536x512x128 & 19.82 & 512x256x128 & 18.5 & 1536x512x256 & 17.83\\
1024x512x128 & 19.92 & 1024x512x128 & 18.78 & 2048x512x256 & 18.65\\
2048x512x256 & 20.5 & 1536x256x128 & 19.2 & 2048x512x128 & 19.5\\
2048x256x128 & 21.25 & 2048x256x128 & 19.47 & 2048x256x128 & 19.55\\
2048x512x128 & 21.7 & 2048x512x256 & 19.92 & 512x256x128 & 19.93\\
    \hline  
  \end{tabular}

\end{centering}
\end{table}

In order to get a more comprehensive idea of the performance of different patch architectures, we ranked the 25 patch architectures by our three evaluation criteria (Tab.~\ref{tab:Patch architectures ranked by evaluation measures}).
The rank is calculated as the average rank over all datasets, classifiers and both candidate layers. For each configuration the rank is obtained by sorting the patch architectures by the respective evaluation measure and assigning ranks. The architectures ’512’,
’1024’, ’1536’, and ’2048’ are the top 4 architectures for all evaluation measures.

The results indicate, that it is not beneficial to increase the number of nodes in the
hidden layer to an arbitrarily high amount. We do not notice an advantage of the ’2048’
architecture over the ’1024’ architecture.

In conclusion, patch architectures with a single hidden layer and a sufficient number
of nodes show the best patching performance on average in all scenarios. 

%
%
%

\subsubsection{Conclusion on Patch Architecture}

\begin{table}[h!]
\begin{centering}
  \begin{tabular}{ | l | l | l | l |}
    \hline
		Base Archetype:&\multicolumn{1}{c|}{FC-NN}&\multicolumn{1}{c|}{CNN}&\multicolumn{1}{c|}{ResNet} \\ \hline
		\textit{MNIST} & fc1 & pool1 & p1 \\\hline
    \textit{NIST} & fc2 & conv5 & p2  \\
    \hline  
  \end{tabular}
\caption{Engagement layer choice for further experiments.}
\label{Engagement Layer Choice for further Experiments}
\end{centering}
\end{table}

After considering our findings on patch architecture selection, we decide that we use
a single hidden layer with 512 nodes as our patch architecture in further experiments,
since this configuration showed good performance in all evaluated scenarios. 
We also fixed a distinctive engagement layer for each base classifier (Tab.~\ref{Engagement Layer Choice for further Experiments}).

For selecting patch architectures to deal with non-stationary environments, we recommend shallow one-hidden-layer architectures with a number of nodes between 512
and 2048, due to their fast adaptation capabilities and sufficient representation power.

\clearpage
\subsection{Exclusive and Inclusive Patch Training}
\label{Theoretical Advantage of exclusive over inclusive Patch Network Training}

In previous sections, we trained the patch network on all instances from each batch. In other words,
 the patch is trained to approximate the new concept after the drift. We call training on all
instances of the data stream \textit{inclusive training}, since the patch is not only trained with
instances from the error region of the base classifier.

The intuition behind the patching algorithm is, that a secondary model improves the
base classifier in error-prone regions of the instance space. After reporting the performance on a batch, we assume that the labels become available, therefore we can train
the patch merely on instances, which a misclassified by the base network. We call this
training scheme \textit{exclusive training}, since the patch is exclusively trained on instances
from the error region of the base classifier.

In this section, we compare \textit{inclusive} and \textit{exclusive} \textit{training} by obtaining theoretical
performance boundaries. The base network and the patch network form a classifier
ensemble. In order to obtain the theoretical performance boundaries, we assume perfect ensemble usage. This means, an instance counts as correctly classified, if either the patch or the
base network correctly predicts the true label of the instance.

\begin{figure}[H]
	\centering
		\subfigure[\label{fig:thesis_inclusive_vs_exclusive_mnist_appear_dense_run4_exclusive} Exclusive training]{\includegraphics[scale=0.35]{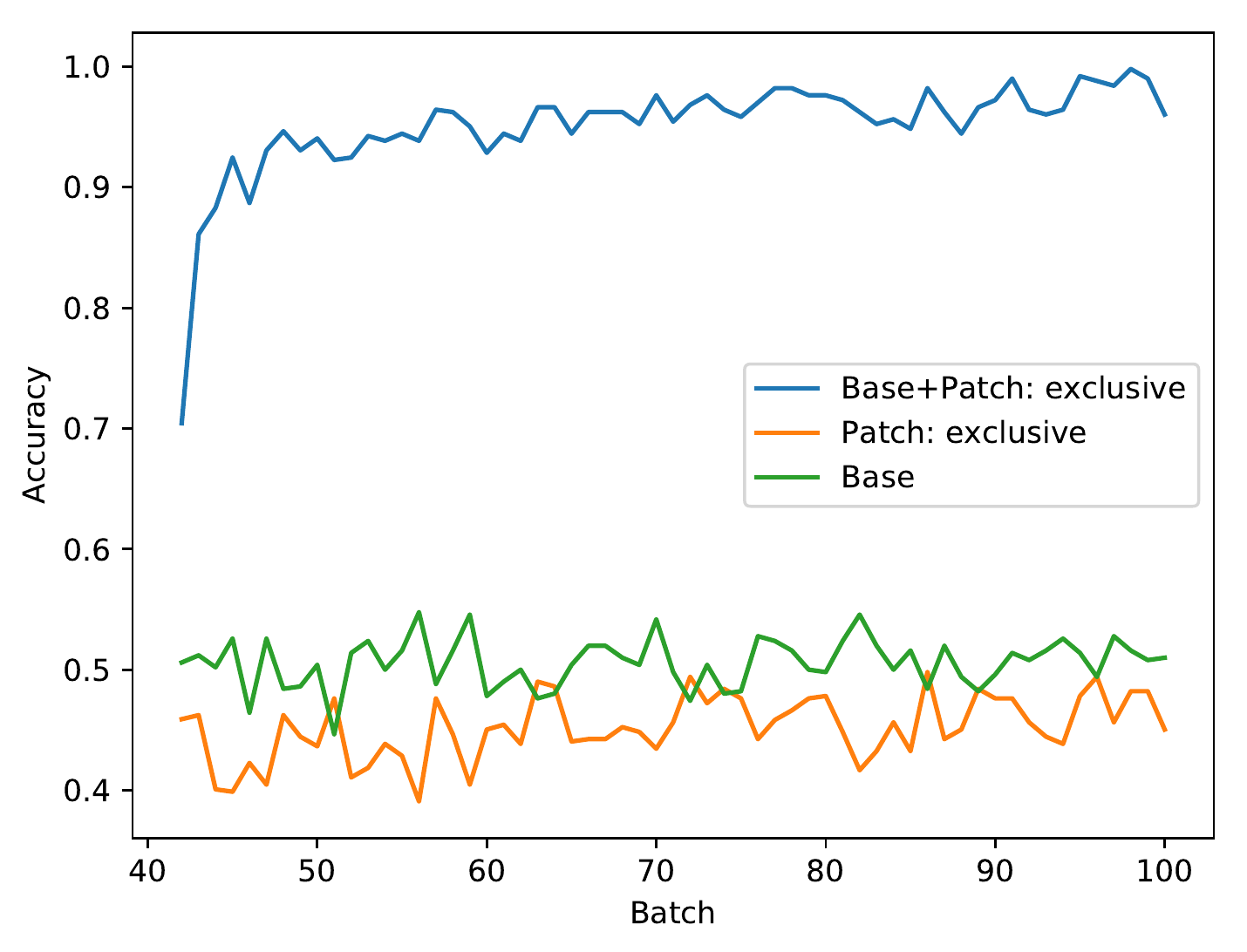}}
		\subfigure[\label{fig:thesis_inclusive_vs_exclusive_mnist_appear_dense_run4_inclusive} Inclusive training]{\includegraphics[scale=0.35]{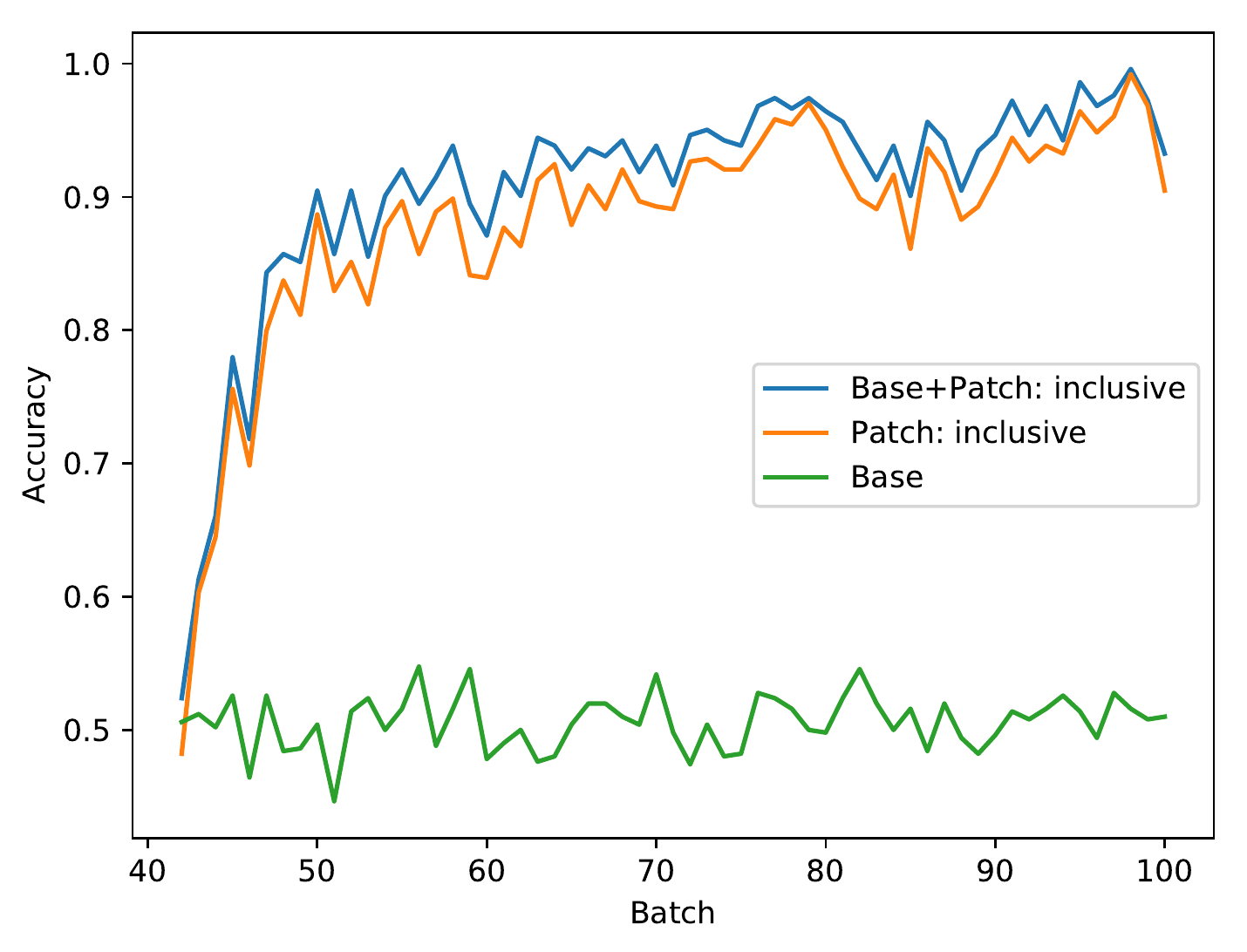}}\\
		\subfigure[\label{fig:thesis_inclusive_vs_exclusive_mnist_appear_dense_run4_incl_vs_excl} Inclusive versus exclusive training]{\includegraphics[scale=0.35]{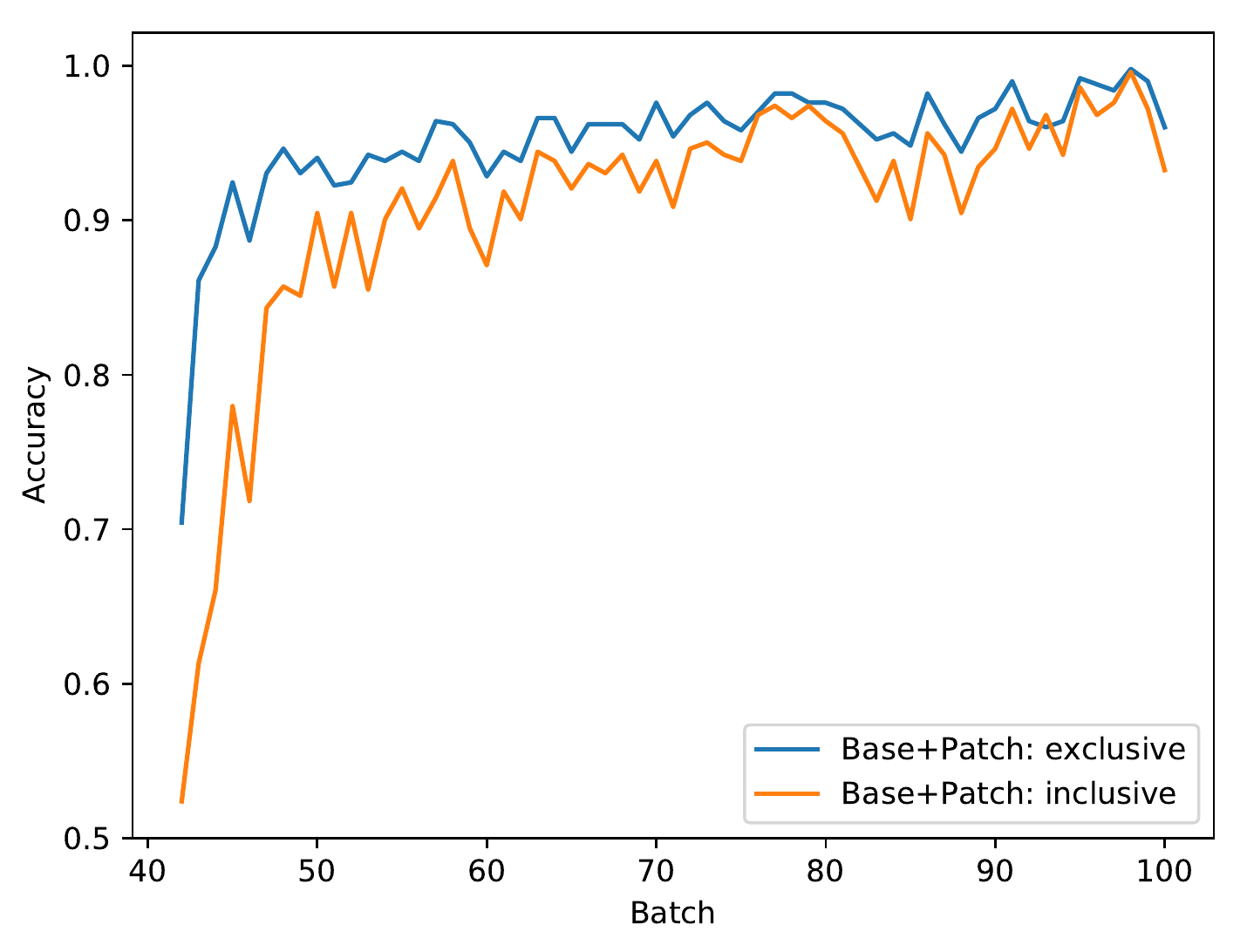}}
\caption{\textbf{Comparison of accuracy curves for inclusive and exclusive training of the patch.} 
}
\label{fig:Comparison of accuracy curves for inclusive and exclusive training of the patch}
\end{figure}  

\newpage
Figure~\ref{fig:Comparison of accuracy curves for inclusive and exclusive training of the patch} shows the accuracy comparison between \textit{exclusive} and \textit{inclusive} \textit{training}.
In Figure~(a) and (b) the accuracies of the unaltered base network, the patch network
and the combined classifiers, assuming perfect ensemble usage, are presented for the
\textit{exclusive} and \textit{inclusive} patch. The accuracies are obtained by evaluating the respective
model on the data from the next batch. The base classifier shows an accuracy of approximately 50\,\%. The accuracy of the \textit{exclusive} patch is around 42\,\%. But the combined
ensemble shows a huge accuracy increase. Since the patch is exclusively trained on instances from the error region of the base network, the classification capabilities of both
models complement one another.

The \textit{inclusive} patch (Fig.~\ref{fig:thesis_inclusive_vs_exclusive_mnist_appear_dense_run4_inclusive}) and the respective ensemble have comparable accuracy. However, the \textit{inclusive} patch is able to classify data from the whole instance
space, hence the classification capabilities of the base network and the patch are overlapping, which results in higher accuracy for the patch. However, the total accuracy of this type of ensemble is lower, as shown in Figure~\ref{fig:thesis_inclusive_vs_exclusive_mnist_appear_dense_run4_incl_vs_excl}. Here we compare the theoretical
accuracy boundary for the classifier ensemble consisting of the base network and the \textit{inclusive}/\textit{exclusive} patch. The \textit{exclusive} patch combined with the base classifier achieves
a higher accuracy than the \textit{inclusive} patch combined with the base network. 

\begin{table}[H]
\begin{centering}
\caption{\textbf{Evaluation of inclusive and exclusive patch network combined with the base network.} 
}
\label{tab:Evaluation of inclusive and exclusive patch network combined with the base network}

  \begin{tabular}{ |l|  c  c | c  c | c  c | }
    \hline
		Evaluation Measure: & \multicolumn{2}{ c|}{Average Accuracy} & \multicolumn{2}{ c|}{Final Accuracy} & \multicolumn{2}{ c|}{Recovery Speed}\\ \hline
		Dataset & excl. & incl.& excl. & incl.& excl. & incl.\\ \hline
		$\mathbf{\textit{MNIST}_{\textit{appear}}}$ & \textbf{94.92} & 92.52 & \textbf{99.18} & 98.52 & \textbf{6.6} & 7.8\\
		$\mathbf{\textit{MNIST}_{\textit{flip}}}$ & \textbf{94.55} & 94.01 & \textbf{97.99} & 97.59 & \textbf{5.2} & \textbf{5.2}\\
		$\mathbf{\textit{MNIST}_{\textit{remap}}}$ & \textbf{94.05} & 93.36 & 97.69 & \textbf{97.7}8 & 6.6 & \textbf{5.6}\\
		$\mathbf{\textit{MNIST}_{\textit{rotate}}}$ & \textbf{77.07} & 75.22 & \textbf{76.7}1 & 76.68 & ---- & ----\\ \hline
		$\mathbf{\textit{NIST}_{\textit{appear}}}$ & \textbf{95.68} & 92.42 & \textbf{97.17} & 94.92 & \textbf{1.8} & 3.8\\
		$\mathbf{\textit{NIST}_{\textit{flip}}}$ & \textbf{88.29} & 87.62 & \textbf{94.34} & 94.06 & \textbf{10.4} & 10.6\\
		$\mathbf{\textit{NIST}_{\textit{remap}}}$ & \textbf{85.55} & 84.46 & \textbf{93.21} & 92.9 & \textbf{9.2}& 12.8\\
		$\mathbf{\textit{NIST}_{\textit{rotate}}}$ & \textbf{62.75} & 62.09 & \textbf{68.06} & 66.83 & ---- & ----\\
    \hline  
  \end{tabular}

\end{centering}
\end{table}

To substantiate this observation, we conducted experiments to obtain evaluation measures for all datasets and classifiers. The results are shown in Table~\ref{tab:Evaluation of inclusive and exclusive patch network combined with the base network}. In this table
we compare the \textit{inclusive} and the \textit{exclusive} ensemble for all datasets. The \textit{exclusive}
patch ensemble outperforms the \textit{inclusive} ensemble in average accuracy, final accuracy
and recovery speed.

\begin{table}[h]
\begin{centering}
  \begin{tabular}{ | l | l | l | l |}
    \hline
		\multicolumn{1}{|l|}{Datasets}&\multicolumn{1}{c|}{FC-NN}&\multicolumn{1}{c|}{CNN}&\multicolumn{1}{c|}{ResNet} \\ \hline
$\mathbf{\textit{MNIST}_{\textit{appear}}}$ & 50.58 & 50.78 & 50.89 \\
$\mathbf{\textit{MNIST}_{\textit{flip}}}$ & 29.55 & 35.9 & 39.67 \\
$\mathbf{\textit{MNIST}_{\textit{remap}}}$ & 41.99 & 33.6 & 35.34 \\
$\mathbf{\textit{MNIST}_{\textit{rotate}}}$ & 46.58 & 51.64 & 52.87 \\
$\mathbf{\textit{MNIST}_{\textit{transfer}}}$ & 0.0 & 0.0 & 0.0 \\ \hline
$\mathbf{\textit{NIST}_{\textit{appear}}}$ & 64.89 & 69.86 & 68.3 \\
$\mathbf{\textit{NIST}_{\textit{flip}}}$ & 14.36 & 17.35 & 15.68 \\
$\mathbf{\textit{NIST}_{\textit{remap}}}$ & 13.33 & 13.05 & 13.64 \\
$\mathbf{\textit{NIST}_{\textit{rotate}}}$ & 32.16 & 38.71 & 39.45 \\
$\mathbf{\textit{NIST}_{\textit{transfer}}}$ & 0.0 & 0.0 & 0.0 \\
    \hline  
  \end{tabular}
\caption{\textbf{Average accuracy of base classifiers after the occurrence of the concept drift.} We evaluate the base network on every batch after the concept drift and report the accuracy. All values are stated in percent. The base classifier is not trained after the initialization phase. All values are averaged over five runs with varying random seed. The average standard deviation is 1.35\,\% ($\mathbf{\textit{MNIST}_{\textit{transfer}}}$ and $\mathbf{\textit{NIST}_{\textit{transfer}}}$ were excluded for obtaining this value).}
\label{Average accuracy of base classifiers after the occurrence of the concept drift}
\end{centering}
\end{table}

The performance difference between the \textit{inclusive} and the \textit{exclusive} ensemble depends on the classification capabilities of the classifier after the occurrence of the concept drift. The average accuracy of base classifiers after the occurrence of the concept
drift for all datasets are listed in Table~\ref{Average accuracy of base classifiers after the occurrence of the concept drift}. The performance difference is highest for
$\mathbf{\textit{NIST}_{\textit{appear}}}$ and $\mathbf{\textit{MNIST}_{\textit{appear}}}$. These are the datasets with the highest base classifier
accuracy after the concept drift. For datasets with lower performing base classifiers, we
observe a smaller performance difference between the inclusive and exclusive ensemble.

\subsubsection{On the Influence of the Error Region Size of the Base Classifier}
The performance increase from the exclusive ensemble is caused by the fact that a
reduced sub-problem is easier to solve than a more complex problem. The feature
space is divided by the error region of the base classifier. The exclusive patch merely
has to classify instances from the error region, which is a sub-problem. Therefore, the
exclusive patch classifies instances inside the error region with higher accuracy than the
inclusive patch.

The difference between inclusive and exclusive training is dependent on the capabilities of the base network to correctly classify instances after the concept drift. The size of the error region is larger for low performing base classifiers, hence the respective sub-problem for the exclusive patch is also large. A better base classifier performance after the drift results in a smaller sub-problem for the patch network. The smaller the problem, the higher the performance of the patch network on the sub-region of the instance space. 

We propose that the lower performance difference between the inclusive and exclusive ensemble for high base accuracies, is due to accuracy saturation. The benefit from the exclusive training is large, since the error region is small, but the overall amount of misclassifications by the base network is low. Hence, only on rare occasions the patch network gets the chance to correct the base classifier.   

We conclude that an exclusive training on instances from the error region of the base network, assuming perfect ensemble usage, leads to a patching performance increase, since the sub-problem in the instance space is easier to solve for the patch network than the comprehensive problem.

\subsubsection{Semi-Exclusive Patch Training}
\label{sec:semiexclusivetraining}
The advantage of inclusive training is the robustness towards a poor error region
estimator. To obtain the theoretical accuracy boundaries, we assumed perfect ensemble
usage. In a real world scenario, the error region estimator creates imperfect predictions.
Since the inclusive patch is capable of classifying instances from the whole instance
space satisfactorily, an error region estimator, which directs most instances to the patch
for classification results in a good performance. The exclusive patch is relying more on
a well-tuned error region estimator.

We introduce semi-exclusive training in order to leverage the robustness of exclusive training towards a fault-prone error region estimator. Semi-exclusive patches are trained on the union of the true error region and the estimated error region. Instances from the true error region of the base classifier can be easily identified after the availability of the true labels. The estimated error region is the prediction of the error estimator network on the current batch. An illustration of inclusive, exclusive, and semi-exclusive patch training is shown in Figure~\ref{fig:Illustration of inclusive, exclusive and semi-exclusive patch training}.

\begin{figure}[H]
	\centering
		\subfigure[\label{fig:inclusive_training} Inclusive training]{\includegraphics[scale=0.24]{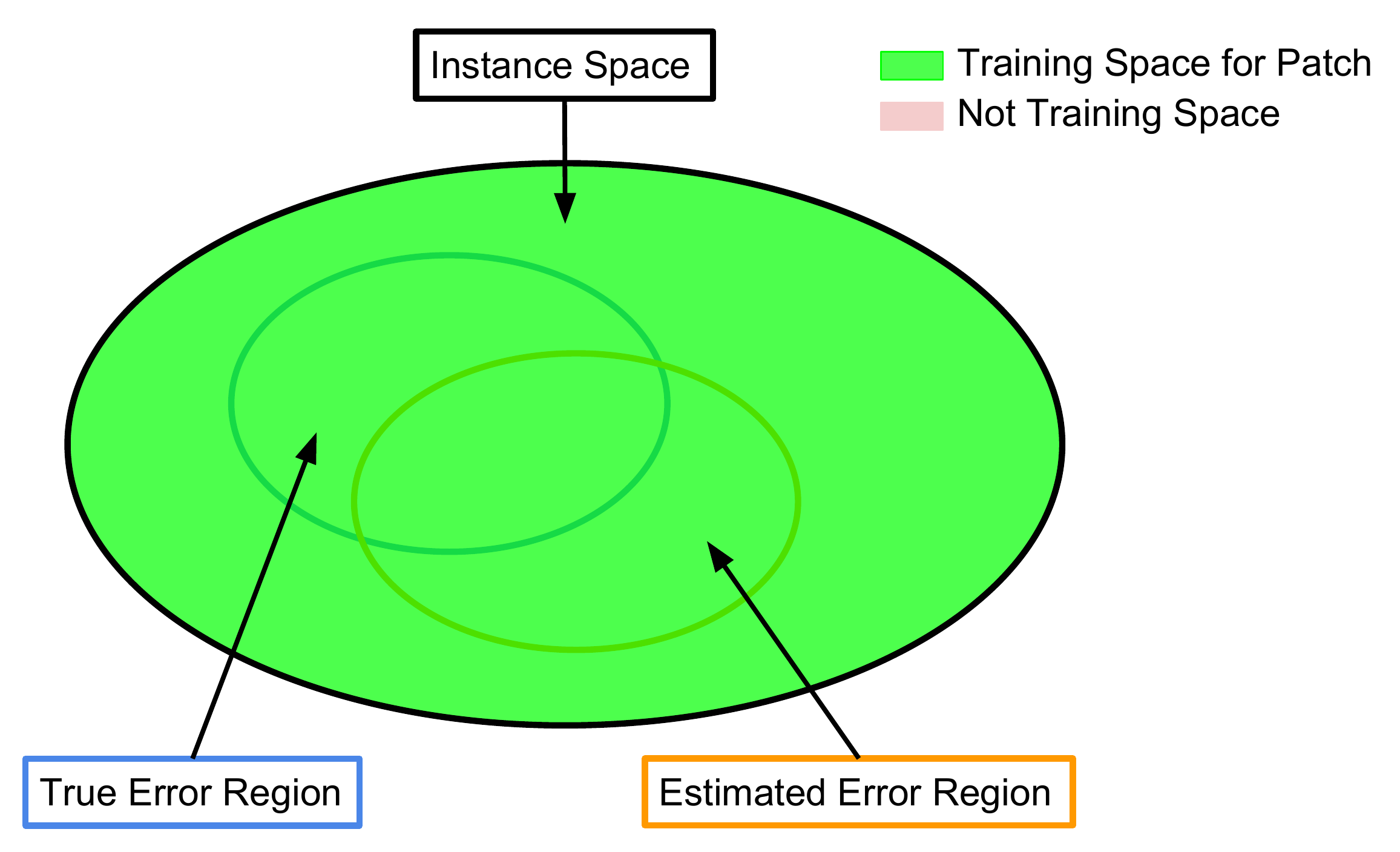}}\quad
		\subfigure[\label{fig:exclusive_training} Exclusive training]{\includegraphics[scale=0.24]{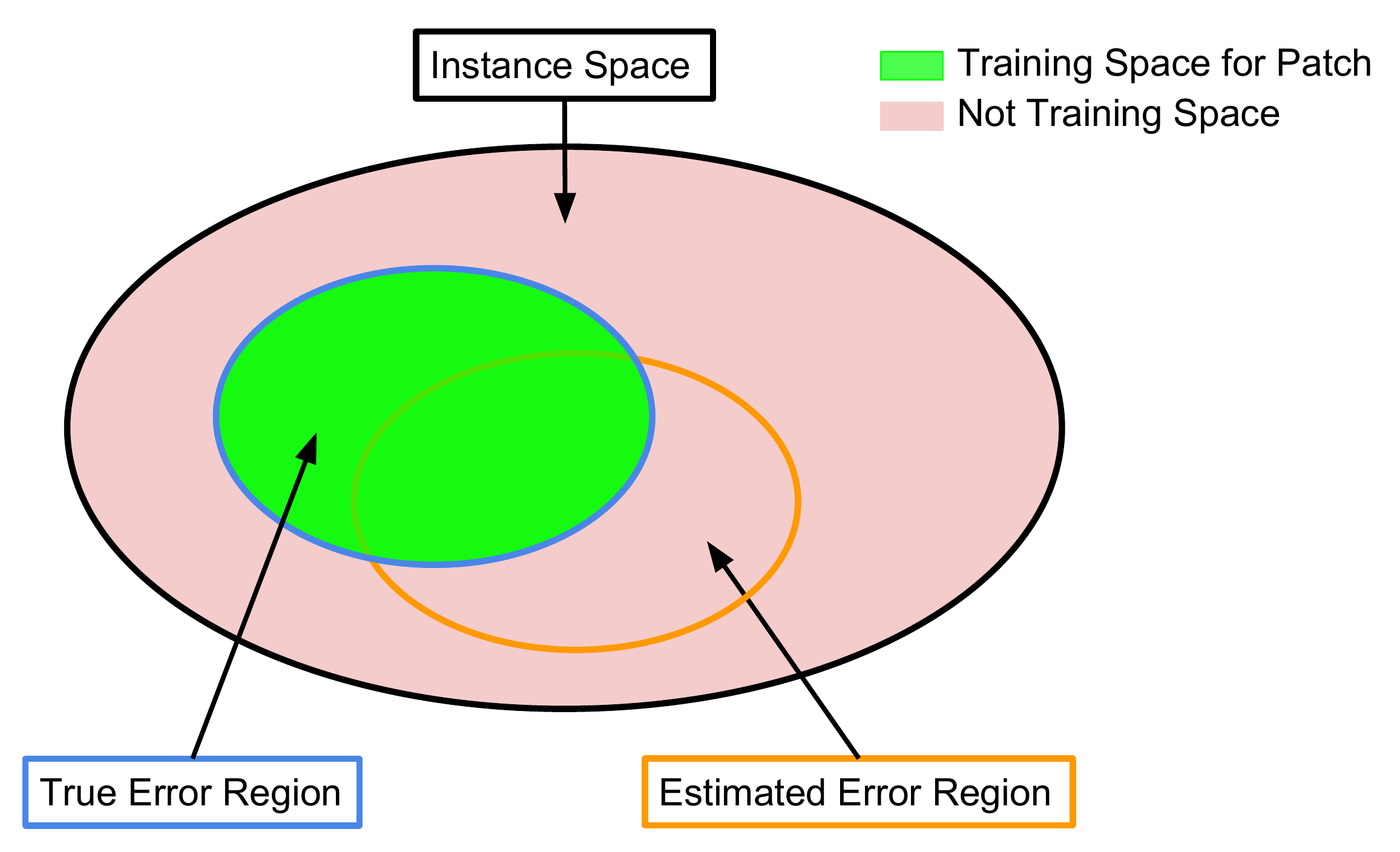}}\\
		\subfigure[\label{fig:semi_exclusive_training} Semi-exclusive training]{\includegraphics[scale=0.24]{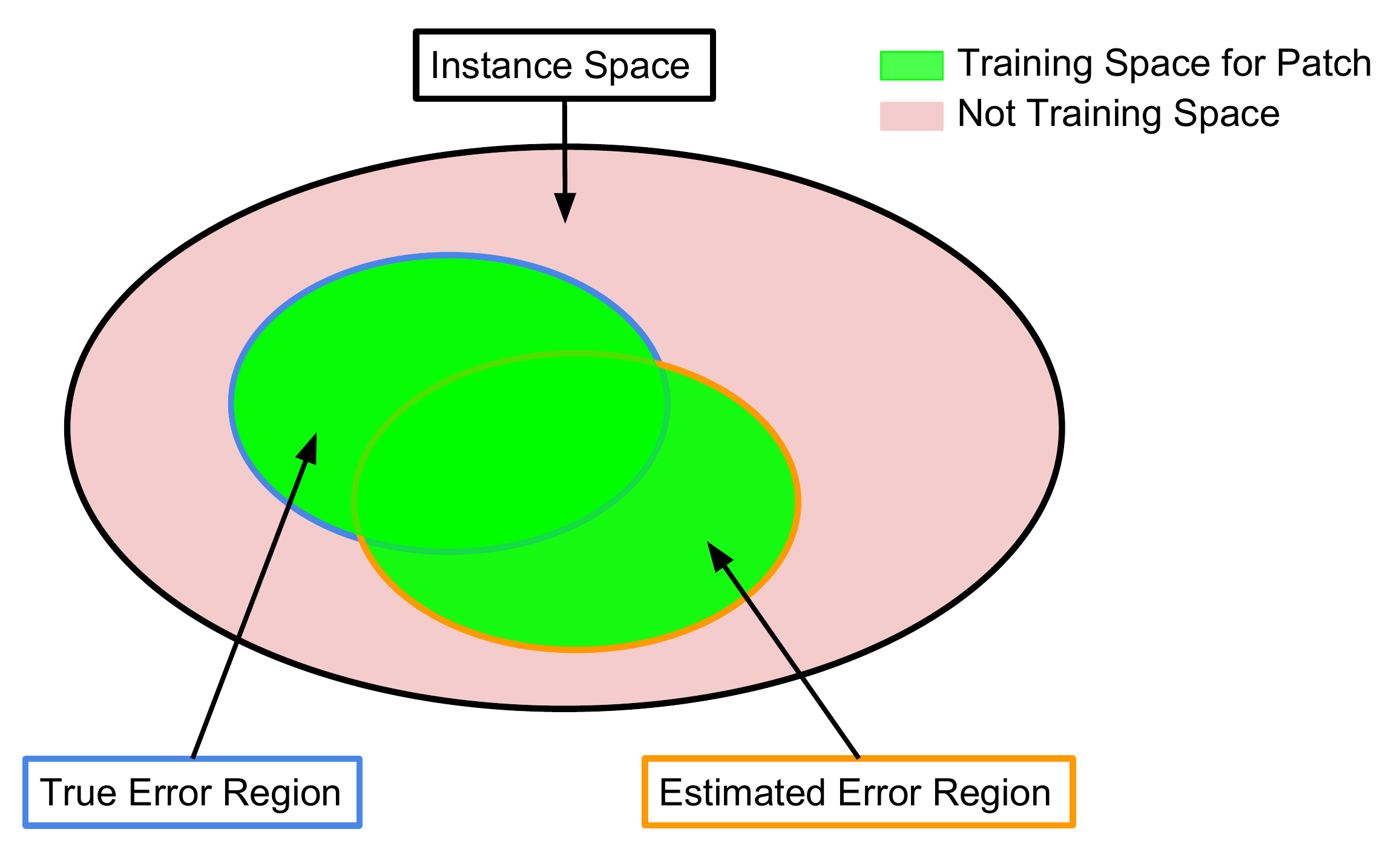}}
\caption{\textbf{Illustration of inclusive, exclusive and semi-exclusive patch training.} The inclusive patch is trained on instances from the whole instance space (Fig.~(a)). Figure~(b) shows the training space for exclusive training The exclusive patch is merely trained on instances from the true error region of the base classifier. The semi-exclusive patch (Fig.~(c)) is trained on instances from the union of the true error region and the estimated error region.}
\label{fig:Illustration of inclusive, exclusive and semi-exclusive patch training}
\end{figure}  

After the occurrence of a concept drift in the data stream, the error region estimator starts to learn the error region of the base classifier. During the first batches of the data stream, the error region estimator is still error-prone, because the amount of training at this time is not sufficient to classify the error region of instances well. If the patch is exclusively trained and receives an instance, which is not inside the error region of the base network, the patch likely fails to correctly classify this instance, because it was not trained to classify training examples from this region of the instance space. 

Semi-exclusive training is an approach to tackle this problem. We train the patch on instances from the union of the error region of the base classifier and the error region estimation. The patch network is now also trained on the instances, which the error region estimator erroneously assigned to the error region of the base classifier. Thus, the capabilities of the patch network are increased. In case the error estimator erroneously directs an instance to the patch network, the probability of a successful classification is increased, since the patch network is additionally trained on instances from this region of the instance space. 

Inclusive patch training leads to the highest robustness towards an error-prone error region
estimator. The inclusive patch is trained to classify all instances in the instance space. However, inclusive training loses the advantage of focussing only on a sub-region of the instance space, which we showed to increase theoretical performance.

The semi-exclusive training scheme combines the benefits of exclusive and inclusive training. Higher robustness is achieved by additionally training the patch on instances from the estimated error region. Moreover, the advantage of only model on a sub-region of the instance space is also preserved. After sufficient training, the true error region of the base classifier and the estimated error region, predicted by the error region estimator, should converge towards each other. Hence, in theory the full benefit of exclusive training (i.e. solving a sub-problem) can be achieved, with the convergence of the estimated error region towards the true error region of the base classifier. 

\subsection{Variants of Patching}
\label{sec:patchingvariants}
Based on our previous findings, we will now establish different variations of NN-Patching. The NN-Patching models variate in the patch training scheme and the modelled error region by the error estimator. The three patch training schemes (i.e. inclusive, exclusive, and semi-exclusive training) are discussed, and an examination of the difference between modelling the error region of the base classifier and modelling the error region of the patch network can be found in Section~\ref{sec:modellingerrorregions}.
\begin{description}[font=\normalfont]
\item[\textbf{NN-Patching$\mathbf{_{incl,noEE}}$}] After the detection of a concept drift, a patch network is initialized and trained on the arriving batches. The patch is trained on all instances, hence this method uses inclusive training. An error estimator network is not used in this setup. All instances are directed to the patch for classification. The abbreviation 'noEE' means no error estimator. 
\item[\textbf{NN-Patching$\mathbf{_{incl,baseEE}}$}] This NN-Patching variant also uses an inclusive patch (i.e. the patch is trained on all instances from the batch). In addition, an error region estimator is used to predict errors of the base classifier. After the arrival of a new batch, the error region estimator predicts the error region of each instance. If the error estimator predicts a successful classification by the base network, the instance is directed to the base classifier. Otherwise, the instance is directed to the patch for classification. The abbreviation 'baseEE' means base error estimator.
\item[\textbf{NN-Patching$\mathbf{_{semi,baseEE}}$}] The patch is semi-exclusively trained. This means, the patch is trained on instances lying in the union of the true error region and the estimated error region. The error estimator network is trained to predict a correct/incorrect classification by the base network. Based on the prediction of the error estimator on an instance, the instance is either directed to the patch network or the base network for classification.
\item[\textbf{NN-Patching$\mathbf{_{excl,baseEE}}$}] The patch is exclusively trained on instances from the error region of the base classifier. The error estimator network is trained to predict a correct/incorrect classification by the base network. Based on the prediction of the error estimator on an instance, the instance is either directed to the patch network or the base network for classification.
\item[\textbf{NN-Patching$\mathbf{_{incl,patchEE}}$}] The patch is trained inclusively. The error estimator predicts an either correct or incorrect classification by the patch network. Therefore, not the error region of the base classifier is modelled by the error estimator, but the error region of the patch network. In case the error estimator predicts a successful classification by the patch, the instance is directed to the patch for classification. Otherwise, the instance is directed to the base network. The abbreviation 'patchEE' means patch error estimator. 
\item[\textbf{NN-Patching$\mathbf{_{semi,patchEE}}$}] The error region estimator predicts an either correct or incorrect classification by the patch network. The patch is trained on the union of the true error region of the base classifier and the estimated non-error region of the patch network. We additionally train the patch on the non-error region of the patch network, since an instance, which is lying in the error region of the patch, would be directed to the base classifier instead of the patch network. Based on the prediction of the error estimator, instances are either directed to the patch network or the base network for classification. 
\item[\textbf{NN-Patching$\mathbf{_{excl,patchEE}}$}] The patch is exclusively trained on the error region of the base classifier. The error estimator network is trained to predict a correct/incorrect classification by the patch network. Based on the prediction of the error estimator, instances are either directed to the patch network or the base network for classification.   
\end{description}

\subsection{Improving Error Region Estimation}
We demonstrated the theoretical advantage of exclusive training (Sec.~\ref{Theoretical Advantage of exclusive over inclusive Patch Network Training}) under the assumption of perfect ensemble usage. In this section, the ensemble is controlled by a neural network: the \textit{error region estimator}. The \textit{error region estimator} network gets the same data input as the patch network and has the same architecture. Therefore, the input for the \textit{error region estimator} is the output of the engagement layer and the network consists of one hidden layer with 512 nodes followed by an output layer. 

The \textit{error region estimator} is trained to predict, if an instance lies in the error region of the base classifier. Deciding whether the base network is capable of classifying an instance is a two class problem. The respective target vector is the true error region of the instance. The base classifier predicts the class of each instance. The predictions are compared with the true class labels. If the prediction matches the label, the instance does not belong to the error region of the base network. In case of an erroneous prediction, the instance lies in the error region.

In Section~\ref{sec:regularization} we discuss the importance of regularizing patch and \textit{error estimator}. After that, we elaborate on the effect of of different base classifier capabilities after the drift on the inclusive, exclusive and semi-exclusive patching performance (Sec.~\ref{Effect of different Base Network Capabilities after Drift on inclusive, exclusive and semi-exclusive Patching Performance}). Finally, in Section~\ref{sec:modellingerrorregions} we discuss the possibility of modelling the error region of the patch network instead of modelling the error region of the base network.

\subsubsection{Regularizing Patch and Error Estimator Network}
\label{sec:regularization}

Overfitting reduces the classification performance of models. Regularization counteracts overfitting. In this section, we investigate the effect of dropout on the patching performance. Therefore, we apply different dropout probabilities to the patch and error estimator network. In Table~\ref{tab:Effect of different Dropout Strengths on the Patching Performance for FC-NN base classifiers} the results are presented for FC-NN base classifiers and in Table~\ref{tab:Effect of different Dropout Strengths on the Patching Performance for CNN base classifiers} for CNN base classifiers.

\begin{table}[h]
\begin{centering}
\caption{\textbf{Effect of different Dropout Probabilities on the Patching Performance for FC-NN Base Classifiers.} The table shows how regularizing the patch and error region estimator effects the evaluation performance. The respective patch and error estimator architecture is Input~-~Dropout(d1)~-~FC(512)~-~Dropout(d2)~-~Output. The dropout probabilities d1, d2 are stated in the first column. All values are averaged over all datasets.  Moreover, for each dataset the values are averaged over 10 runs. The base classifier architecture is FC-NN.}
\label{tab:Effect of different Dropout Strengths on the Patching Performance for FC-NN base classifiers}

  \begin{tabular}{ |l | l l l | l l l |}
    \hline
		\multicolumn{7}{ |c|}{Base Classifier: Fully-Connected Architecture}\\ \hline \hline
		Model:&\multicolumn{3}{ c|}{\textit{NN-Patching}$_{\textit{incl,noEE}}$}&\multicolumn{3}{ c|}{\textit{NN-Patching}$_{\textit{incl,baseEE}}$}\\ \hline
		Dropout Probs.& A.Acc.&F.Acc.&R.Spd.& A.Acc.&F.Acc.&R.Spd.\\ \hline
No Dropout       & 77.3 & 81.89 & \textbf{20.73}& 76.35 & 80.54 & \textbf{19.98} \\
d1=0.25,d2=0.25  & 77.4 & 81.95 & 20.83& 76.37 & 80.66 & 20.07 \\
d1=0.25,d2=0.5   & \textbf{77.46} & \textbf{82.05} & 20.98& \textbf{76.43} & \textbf{80.81} & 20.04 \\
d1=0.5,d2=0.25   & 77.27 & 81.84 & 21.07& 76.25 & 80.7 & 20.22 \\
d1=0.5,d2=0.5    & 77.11 & 81.9 & 21.4& 76.12 & 80.68 & 20.63\\
    \hline  
    \hline
		Model: &\multicolumn{3}{ c|}{\textit{NN-Patching}$_{\textit{semi,baseEE}}$}&\multicolumn{3}{ c|}{\textit{NN-Patching}$_{\textit{excl,baseEE}}$}\\ \hline
		Dropout Probs.& A.Acc.&F.Acc.&R.Spd.& A.Acc.&F.Acc.&R.Spd.\\ \hline
No Dropout      &  76.3 & 80.56 & 20.71& 74.84 & 78.7 & \textbf{24.24} \\
d1=0.25,d2=0.25  & \textbf{76.37} & 80.57 & 20.73 & 74.87 & 78.79 & 24.33\\
d1=0.25,d2=0.5   & 76.35 & \textbf{80.63} & 20.75  & \textbf{74.92} & \textbf{78.81} & \textbf{24.24} \\
d1=0.5,d2=0.25   & 76.13 & 80.49 & \textbf{20.7} & 74.63 & 78.65 & 24.49\\
d1=0.5,d2=0.5   &  76.09 & 80.53 & 21.0 & 74.57 & 78.58 & 24.44 \\	\hline	
  \end{tabular}

\end{centering}
\end{table}  

We evaluated four different dropout settings and the non-regularized patch and error estimator for reference. The architecture of the patch and error estimator is Input~-~Dropout(d1)~-~FC(512)~-~Dropout(d2)~-~Output. The used dropout probabilities d1 and d2 are stated in the first column of the table. 

For a FC-NN classifier the performance increase through dropout is marginal. In some cases, the application of dropout has a negative impact on the recovery speed. This is expected, since gradient updates under dropout only train a subnet of the neural network. Hence, neural networks with dropout layers train slower than without dropout~\cite{JMLR:v15:srivastava14a}.   	

\begin{table}[h]
\begin{centering}
\caption{\textbf{Effect of different Dropout Probabilities on the Patching Performance for CNN Base Classifiers.} The table shows how regularizing the patch and error region estimator effects the evaluation performance. The respective patch and error estimator architecture is Input~-~Dropout(d1)~-~FC(512)~-~Dropout(d2)~-~Output. The dropout probabilities d1, d2 are stated in the first column. All values are averaged over all datasets. Moreover, for each dataset the values are averaged over 10 runs. The base classifier architecture is CNN.}
\label{tab:Effect of different Dropout Strengths on the Patching Performance for CNN base classifiers}

  \begin{tabular}{ |l | l l l | l l l |}
    \hline
		\multicolumn{7}{ |c|}{Base Classifier: Convolutional Architecture}\\ \hline \hline
		Model:&\multicolumn{3}{ c|}{\textit{NN-Patching}$_{\textit{incl,noEE}}$}&\multicolumn{3}{ c|}{\textit{NN-Patching}$_{\textit{incl,baseEE}}$}\\ \hline
		Dropout Probs.& A.Acc.&F.Acc.&R.Spd.& A.Acc.&F.Acc.&R.Spd.\\ \hline
		No Dropout  & 87.01 & 90.76 & 10.64 & 85.85 & 88.97 & 6.46 \\
d1=0.25,d2=0.25 & 87.36 & 91.06 & \textbf{9.56} & 86.08 & 89.03 & 6.32  \\
d1=0.25,d2=0.5  & \textbf{87.42} & \textbf{91.17} & 9.69 & \textbf{86.14} & \textbf{89.34} & 6.32  \\
d1=0.5,d2=0.25  & 87.35 & 90.94 & 10.07 & 85.99 & 88.97 & \textbf{6.27}  \\
d1=0.5,d2=0.5   & 87.21 & 91.07 & 10.43 & 85.94 & 88.97 & 6.28  \\ \hline
    \hline  
    \hline
		Model: &\multicolumn{3}{ c|}{\textit{NN-Patching}$_{\textit{semi,baseEE}}$}&\multicolumn{3}{ c|}{\textit{NN-Patching}$_{\textit{excl,baseEE}}$}\\ \hline
		Dropout Probs.& A.Acc.&F.Acc.&R.Spd.& A.Acc.&F.Acc.&R.Spd.\\ \hline
		No Dropout  & 85.82 & 88.81 & 6.46 & 84.6 & 87.16 & 7.22 \\
d1=0.25,d2=0.25 & 86.01 & 89.08 & 6.27 & 84.77 & 87.36 & 7.16 \\
d1=0.25,d2=0.5  & \textbf{86.12} & \textbf{89.16} & \textbf{6.21} & \textbf{84.82} & \textbf{87.41} & \textbf{6.96} \\
d1=0.5,d2=0.25   & 86.0 & 88.89 & 6.46 & 84.66 & 87.18 & 7.2 \\
d1=0.5,d2=0.5    & 86.02 & 88.82 & 6.67 & 84.69 & 87.12 & 7.4 \\ \hline
  \end{tabular}

\end{centering}
\end{table}  

However, dropout often has a positive impact on recovery speed and otherwise the degradation in recovery speed is insignificant. In terms of average and final accuracy applying dropout is always beneficial. The positive effect is rather small for FC-NNs, but for CNNs the benefit is more significant.  

\begin{figure}[H]
\centering
\includegraphics[scale=0.5]{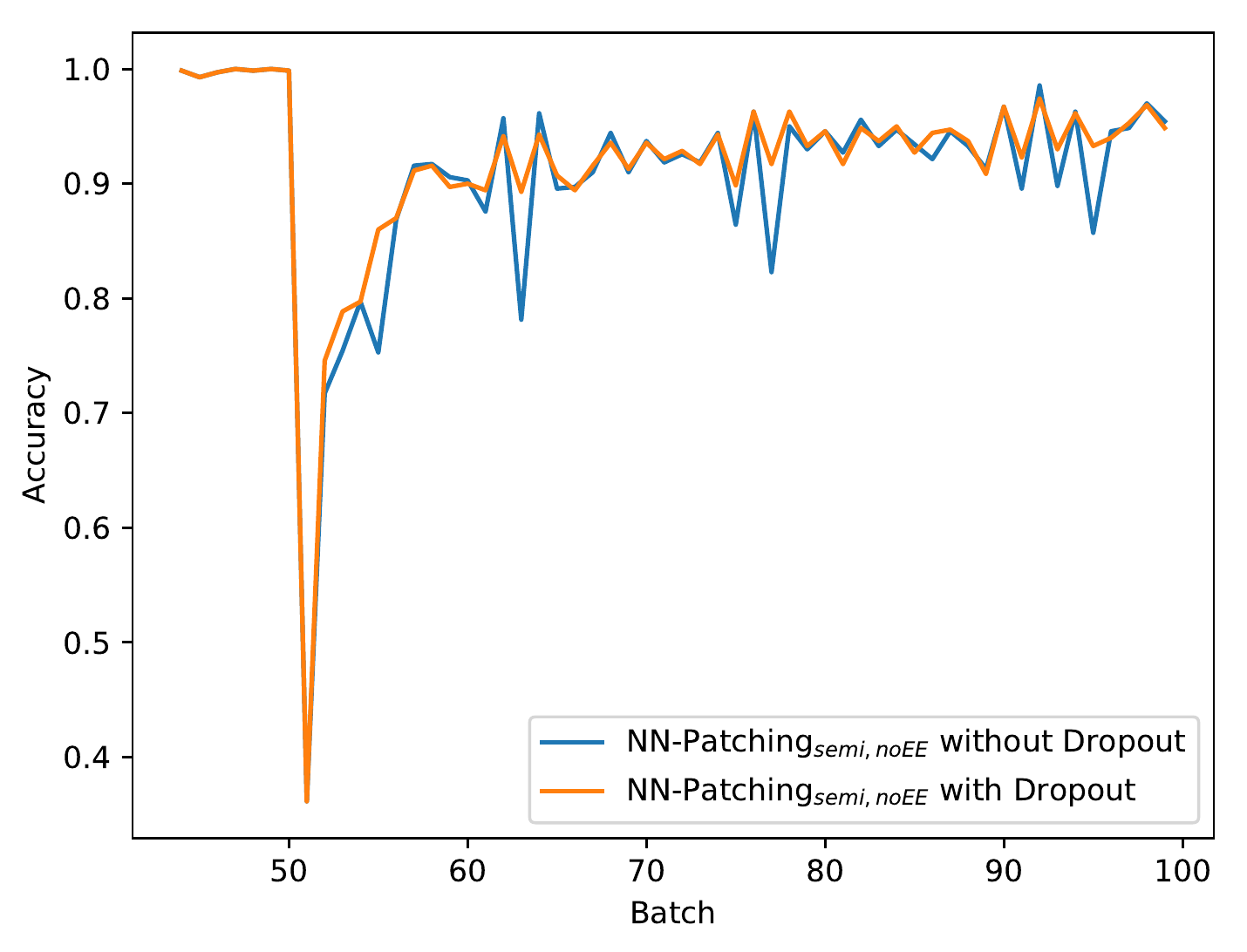}
\caption{\textbf{Reduction of inconsistent performance through regularization.} The figure shows the accuracy progressions for \textit{NN-Patching}$_{\textit{semi,baseEE}}$ with and without dropout. The dropout architecture for patch and error estimator network is 
Input - Dropout(0.25) - FC(512) - Dropout(0.5) - Output. 
The used dataset is $\mathbf{\textit{MNIST}_{\textit{remap}}}$ and the base classifier is a CNN.} 
\label{fig:thesis_dropout_architecture_conv_mnist_remap_run_0}
\end{figure}

Dropout forces the network to generalize more. Therefore, the negative impact of outlier instances in the training data on the model performance is less significant. This effect can be observed as a reduction of inconsistencies in the accuracy progression for regularized models (Fig.~\ref{fig:thesis_dropout_architecture_conv_mnist_remap_run_0}). The benefit of generalization overcomes the disadvantage of slower training in most cases. 

We conducted the regularization experiments for models with inclusive, semi-exclusive and exclusive trained patch networks. The benefit of dropout on these ensemble methods is comparable to the benefit of patching without error estimator. Therefore, we also investigated the possibility that it is only beneficial to regularize the patch instead of regularizing patch and error estimator. However, this is not the case. The absence of regularization in the error estimator network resulted in a performance decrease.

Qualitative differences in the benefit of dropout between \textit{NN-Patching}$_{\textit{incl,baseEE}}$, \textit{NN-Patching}$_{\textit{semi,baseEE}}$ and \textit{NN-Patching}$_{\textit{excl,baseEE}}$ are not observed (i.e. the total performance increase is comparable for all evaluation measures). 

Moreover, we checked if the application of dropout changed the optimal patch architecture depth. We found out that shallow patch architectures with one hidden layer are also superior with applied regularization. 

We conclude, that it is overall beneficial to regularize patch and error estimator. The optimal dropout rate is dependent on various circumstances (e.g. amount of training data, network architecture). This section should provide an short overview to get an intuition on what to expect from regularizing NN-Patching models. 

The presented results show that using the patch and error estimator architecture Input~-~Dropout(0.25)~-~FC(512)~-~Dropout(0.5)~-~Output leads to the best overall performance. Thus, we use this regularized architecture for further experiments.

\subsubsection{Effect of different Base Classifier Capabilities after Drift on the inclusive, exclusive and semi-exclusive Patching Performance}
\label{Effect of different Base Network Capabilities after Drift on inclusive, exclusive and semi-exclusive Patching Performance}

In Section~\ref{Theoretical Advantage of exclusive over inclusive Patch Network Training} we observe the theoretical advantage of exclusive over inclusive patch training. In this section, we evaluate the 
models in a more realistic scenario. Perfect ensemble usage is no longer assumed, instead the ensemble is now controlled by an error region estimator.

We evaluate four neural network patching models (i.e \textit{NN-Patching}$_{\textit{incl,noEE}}$, \textit{NN-Patching}$_{\textit{incl,baseEE}}$,\textit{NN-Patching}$_{\textit{semi,baseEE}}$, and \textit{NN-Patching}$_{\textit{excl,baseEE}}$) on the altered \\
$\mathbf{\textit{NIST}_{\textit{appear}}}$ and $\mathbf{\textit{MNIST}_{\textit{appear}}}$ datasets. 
Hence, we observe the patching performances of the different models with respect to the base network classification capability after the drift. The evaluation on $\mathbf{\textit{MNIST}_{\textit{appear}}}$ based datasets is shown in Table~\ref{tab:Comparison of the inclusive, semi-exclusive, exclusive ensemble and the stand-alone inclusive patch on altered $MNIST_{appear}$ datasets with different base classifier capabilities after the drift}. The results for $\mathbf{\textit{NIST}_{\textit{appear}}}$ based datasets are presented in Table~\ref{tab:Comparison of the inclusive, semi-exclusive, exclusive ensemble and the stand-alone inclusive patch on altered $NIST_{appear}$ datasets with different base classifier capabilities after the drift}.

\begin{table}[h]
\caption{\textbf{Comparison of the inclusive, semi-exclusive, exclusive ensemble and the stand-alone inclusive patch on altered $\mathbf{\textit{MNIST}_{\textit{appear}}}$ datasets with different base classifier capabilities after the drift.}  All values are averaged over 10 runs with varying random seed. The base network architecture is CNN. Average accuracy and final accuracy are stated in percent. The average accuracy after the drift for the untrained base classifier is stated below each dataset name.}
\label{tab:Comparison of the inclusive, semi-exclusive, exclusive ensemble and the stand-alone inclusive patch on altered $MNIST_{appear}$ datasets with different base classifier capabilities after the drift}
\scriptsize
\begin{centering}

  \begin{tabular}{ |l | l  l l | l l  l | l l l | }
    \hline
		Dataset: & \multicolumn{3}{ c|}{$\mathbf{\textit{MNIST}_{\textit{appear,0\text{-}1}}}$}& \multicolumn{3}{ c|}{$\mathbf{\textit{MNIST}_{\textit{appear,0\text{-}2}}}$}& \multicolumn{3}{ c|}{$\mathbf{\textit{MNIST}_{\textit{appear,0\text{-}3}}}$} \\ \hline
		Baseline after Drift: & \multicolumn{3}{ l|}{Avg.Acc. = 22.48\,\%}& \multicolumn{3}{ l|}{Avg.Acc. = 33.23\,\%}& \multicolumn{3}{ l|}{Avg.Acc. = 41.12\,\%} \\ \hline
		Model & A.Acc & F.Acc & R.Spd & A.Acc & F.Acc & R.Spd & A.Acc & F.Acc & R.Spd \\ \hline
		\textit{NN-Patching}$_{\textit{incl,noEE}}$ & \textbf{81.7}8 &\textbf{93.34} & 26.5 & \textbf{91.02} & \textbf{97.85} & 12.3 & \textbf{92.49} & \textbf{98.28} & 10.3 \\
\textit{NN-Patching}$_{\textit{incl,baseEE}}$ & 81.22 & 92.69 & \textbf{26.0} & 90.41 & 97.75 & 13.2 & 91.85 & 98.09 & 11.0 \\
\textit{NN-Patching}$_{\textit{semi,baseEE}}$ & 81.37 & 92.62 & 27.3 & 90.5 & 97.71 & \textbf{11.8} & 91.84 & 98.16 & \textbf{10.2} \\
\textit{NN-Patching}$_{\textit{excl,baseEE}}$ & 80.72 & 92.71 & 30.8 & 90.07 & 97.4 & 12.1 & 90.9 & 97.67 & 10.4 \\
\hline\hline
Dataset: & \multicolumn{3}{ c|}{$\mathbf{\textit{MNIST}_{\textit{appear}}}$}& \multicolumn{3}{ c|}{$\mathbf{\textit{MNIST}_{\textit{appear,0\text{-}5}}}$}& \multicolumn{3}{ c|}{$\mathbf{\textit{MNIST}_{\textit{appear,0\text{-}6}}}$} \\ \hline
Baseline after Drift: & \multicolumn{3}{ l|}{Avg.Acc. = 50.76\,\%}& \multicolumn{3}{ l|}{Avg.Acc. = 60.28\,\%}& \multicolumn{3}{ l|}{Avg.Acc. = 70.15\,\%} \\ \hline
Model & A.Acc & F.Acc & R.Spd & A.Acc & F.Acc & R.Spd & A.Acc & F.Acc & R.Spd \\ \hline
\textit{NN-Patching}$_{\textit{incl,noEE}}$ & \textbf{92.9} & 98.27 & \textbf{6.8} & \textbf{93.87} & \textbf{98.46} & 7.0 & 94.08 & 98.49 & 7.4 \\
\textit{NN-Patching}$_{\textit{incl,baseEE}}$ & 92.07 & 98.25 & 8.1 & 93.3 & 98.29 & \textbf{6.8} & \textbf{94.28} & \textbf{98.54} & \textbf{4.4} \\
\textit{NN-Patching}$_{\textit{semi,baseEE}}$ & 91.91 & \textbf{98.32} & 9.1 & 92.92 & 98.29 & 7.3 & 94.07 & 98.44 & 5.3 \\
\textit{NN-Patching}$_{\textit{excl,baseEE}}$ & 90.1 & 97.83 & 10.4 & 91.47 & 98.06 & 7.7 & 93.05 & 98.22 & 5.8 \\
  \hline
	\end{tabular}
	\end{centering}
  \begin{tabular}{ |l | l  l l | l l  l | }
	\hline
Dataset: & \multicolumn{3}{ c|}{$\mathbf{\textit{MNIST}_{\textit{appear,0\text{-}7}}}$}& \multicolumn{3}{ c|}{$\mathbf{\textit{MNIST}_{\textit{appear,0\text{-}8}}}$}\\ \hline
Baseline after Drift: & \multicolumn{3}{ l|}{Avg.Acc. = 79.31\,\%}& \multicolumn{3}{ l|}{Avg.Acc. = 88.77\,\%} \\ \hline
Model & A.Acc & F.Acc & R.Spd & A.Acc & F.Acc & R.Spd \\ \hline
\textit{NN-Patching}$_{\textit{incl,noEE}}$ & 94.18 & 98.45 & 6.5 & 94.56 & 98.71 & 6.1 \\
\textit{NN-Patching}$_{\textit{incl,baseEE}}$ & 94.38 & 98.49 & 5.1 & 96.71 & 99.12 & 1.2 \\
\textit{NN-Patching}$_{\textit{semi,baseEE}}$ & \textbf{94.48} & \textbf{98.56} & \textbf{4.6} & \textbf{96.79} & \textbf{99.19} & \textbf{1.0} \\
\textit{NN-Patching}$_{\textit{excl,baseEE}}$ & 93.55 & 98.35 & 6.1 & 96.59 & 99.13 & 5.7 \\
    \hline  
  \end{tabular}

\end{table}

Both tables illustrate similar trends. If the base network is only capable of successfully classifying  65\,\% or less of the instances after the occurrence of the concept drift, \textit{NN-Patching}$_{\textit{incl,noEE}}$  results in highest performance for average and final accuracy. In terms of recovery speed, already a lower base classifier accuracy leads to a faster recovery speed for the ensemble methods \textit{NN-Patching}$_{\textit{incl,baseEE}}$ and \textit{NN-Patching}$_{\textit{semi,baseEE}}$ than the stand-alone patch. The exclusive approach is not preferable in any case. 

The performance difference of \textit{NN-Patching}$_{\textit{incl,noEE}}$ for the different datasets only comes from the changing quality of features in the engagement layer. The performance of \textit{NN-Patching}$_{\textit{incl,noEE}}$ increases simultaneously with the base classifier accuracy after the drift, since the concept before the drift becomes closer to the new concept after the drift. Therefore, also the latent features in the engagement layer become more relevant with respect to the target task.

\begin{table}[!htbp]
\caption{\textbf{Comparison of the inclusive, semi-exclusive, exclusive ensemble and the stand-alone inclusive patch on altered $\mathbf{\textit{NIST}_{\textit{appear}}}$ datasets with different base classifier capabilities after the drift.}  All values are averaged over 10 runs with varying random seed. The base network architecture is CNN. Average accuracy and final accuracy are stated in percent. The average accuracy after the drift for the untrained base classifier is stated below each dataset name.}
\label{tab:Comparison of the inclusive, semi-exclusive, exclusive ensemble and the stand-alone inclusive patch on altered $NIST_{appear}$ datasets with different base classifier capabilities after the drift}

\scriptsize
\begin{centering}
  \begin{tabular}{ |l | l  l l | l l  l | l l l | }
    \hline
		Dataset: & \multicolumn{3}{ c|}{$\mathbf{\textit{NIST}_{\textit{appear,S\text{-}Z}}}$ }& \multicolumn{3}{ c|}{$\mathbf{\textit{NIST}_{\textit{appear,Q\text{-}Z}}}$}& \multicolumn{3}{ c|}{$\mathbf{\textit{NIST}_{\textit{appear,O\text{-}Z}}}$} \\ \hline
		Baseline after Drift: & \multicolumn{3}{ l|}{Avg.Acc. = 21.64\,\%}& \multicolumn{3}{ l|}{Avg.Acc. = 27.11\,\%}& \multicolumn{3}{ l|}{Avg.Acc. = 32.59\,\%} \\ \hline
				Model & A.Acc & F.Acc & R.Spd & A.Acc & F.Acc & R.Spd & A.Acc & F.Acc & R.Spd \\ \hline
		\textit{NN-Patching}$_{\textit{incl,noEE}}$ & \textbf{85.36} & \textbf{91.2} & \textbf{29.2} & \textbf{86.6} & \textbf{91.91} & \textbf{20.7} & \textbf{87.7} & \textbf{92.53} & \textbf{17.9} \\
\textit{NN-Patching}$_{\textit{incl,baseEE}}$ & 84.56 & 90.56 & 33.1 & 85.46 & 91.0 & 22.8 & 86.14 & 91.51 & 20.2 \\
\textit{NN-Patching}$_{\textit{semi,baseEE}}$ & 84.57 & 90.59 & 35.0 & 85.64 & 90.64 & 26.0 & 86.16 & 91.42 & 20.5 \\
\textit{NN-Patching}$_{\textit{excl,baseEE}}$ & 82.41 & 89.03 & 52.4 & 82.24 & 88.84 & 48.9 & 82.45 & 89.57 & 46.6 \\
\hline\hline
Dataset: & \multicolumn{3}{ c|}{$\mathbf{\textit{NIST}_{\textit{appear,M\text{-}Z}}}$}& \multicolumn{3}{ c|}{$\mathbf{\textit{NIST}_{\textit{appear,K\text{-}Z}}}$}& \multicolumn{3}{ c|}{$\mathbf{\textit{NIST}_{\textit{appear,I\text{-}Z}}}$} \\ \hline
Baseline after Drift: & \multicolumn{3}{ l|}{Avg.Acc. = 38.28\,\%}& \multicolumn{3}{ l|}{Avg.Acc. = 43.57\,\%}& \multicolumn{3}{ l|}{Avg.Acc. = 48.85\,\%} \\ \hline
		Model & A.Acc & F.Acc & R.Spd & A.Acc & F.Acc & R.Spd & A.Acc & F.Acc & R.Spd \\ \hline
\textit{NN-Patching}$_{\textit{incl,noEE}}$ & \textbf{88.31} & \textbf{92.7} & \textbf{16.6} & \textbf{88.51} & \textbf{92.96} & \textbf{16.1} & \textbf{89.11} & \textbf{93.26} & 15.6 \\
\textit{NN-Patching}$_{\textit{incl,baseEE}}$ & 86.91 & 91.84 & 18.4 & 87.24 & 92.01 & 18.0 & 87.72 & 92.68 & 16.3 \\
\textit{NN-Patching}$_{\textit{semi,baseEE}}$ & 87.06 & 91.86 & 18.8 & 87.27 & 92.04 & 16.5 & 87.74 & 92.86 & \textbf{15.9} \\
\textit{NN-Patching}$_{\textit{excl,baseEE}}$ & 83.75 & 90.24 & 37.4 & 84.17 & 90.04 & 36.5 & 84.88 & 91.68 & 31.0 \\
\hline\hline
Dataset: & \multicolumn{3}{ c|}{$\mathbf{\textit{NIST}_{\textit{appear,G\text{-}Z}}}$}& \multicolumn{3}{ c|}{$\mathbf{\textit{NIST}_{\textit{appear,E\text{-}Z}}}$}& \multicolumn{3}{ c|}{$\mathbf{\textit{NIST}_{\textit{appear,C\text{-}Z}}}$} \\ \hline
Baseline after Drift: & \multicolumn{3}{ l|}{Avg.Acc. = 53.97\,\%}& \multicolumn{3}{ l|}{Avg.Acc. = 59.39\,\%}& \multicolumn{3}{ l|}{Avg.Acc. = 64.5\,\%} \\ \hline
Model & A.Acc & F.Acc & R.Spd & A.Acc & F.Acc & R.Spd & A.Acc & F.Acc & R.Spd \\ \hline
\textit{NN-Patching}$_{\textit{incl,noEE}}$ & \textbf{89.66} & \textbf{93.63} & 13.6 & \textbf{90.37} & \textbf{94.0} & 10.2 & \textbf{91.07} & \textbf{94.39} & 6.7 \\
\textit{NN-Patching}$_{\textit{incl,baseEE}}$ & 88.2 & 92.53 & \textbf{12.4} & 89.07 & 93.07 & \textbf{8.5} & 90.3 & 93.55 & \textbf{5.1} \\
\textit{NN-Patching}$_{\textit{semi,baseEE}}$ & 88.24 & 92.58 & 12.9 & 89.05 & 93.05 & \textbf{8.5} & 90.26 & 93.61 & 6.0 \\
\textit{NN-Patching}$_{\textit{excl,baseEE}}$ & 85.02 & 91.18 & 30.7 & 86.73 & 91.68 & 16.1 & 88.46 & 92.5 & 10.0 \\
\hline\hline
Dataset: & \multicolumn{3}{ c|}{$\mathbf{\textit{NIST}_{\textit{appear}}}$ }& \multicolumn{3}{ c|}{$\mathbf{\textit{NIST}_{\textit{appear,8\text{-}Z}}}$}& \multicolumn{3}{ c|}{$\mathbf{\textit{NIST}_{\textit{appear,6\text{-}Z}}}$} \\ \hline
Baseline after Drift: & \multicolumn{3}{ l|}{Avg.Acc. = 69.86\,\%}& \multicolumn{3}{ l|}{Avg.Acc. = 74.95\,\%}& \multicolumn{3}{ l|}{Avg.Acc. = 80.52\,\%} \\ \hline
Model & A.Acc & F.Acc & R.Spd & A.Acc & F.Acc & R.Spd & A.Acc & F.Acc & R.Spd \\ \hline
\textit{NN-Patching}$_{\textit{incl,noEE}}$ & 91.44 & \textbf{94.51} & 6.7 & 91.69 & \textbf{94.51} & 5.9 & 91.59 & \textbf{94.62} & 5.5 \\
\textit{NN-Patching}$_{\textit{incl,baseEE}}$ & 91.55 & 93.85 & 4.7 & 91.77 & 94.07 & 5.4 & 91.95 & 94.5 & \textbf{5.1} \\
\textit{NN-Patching}$_{\textit{semi,baseEE}}$ & \textbf{91.59} & 93.96 & \textbf{3.9} & \textbf{91.83} & 94.14 & \textbf{5.2} & \textbf{92.03} & 94.52 & \textbf{5.1} \\
\textit{NN-Patching}$_{\textit{excl,baseEE}}$ & 90.35 & 93.17 & 6.4 & 90.69 & 93.84 & 5.9 & 91.1 & 93.98 & 6.2 \\
 \hline
\end{tabular}
\end{centering}
  \begin{tabular}{ |l | l  l l | l l  l | }
    \hline
Dataset: & \multicolumn{3}{ c|}{$\mathbf{\textit{NIST}_{\textit{appear,4\text{-}Z}}}$}& \multicolumn{3}{ c|}{$\mathbf{\textit{NIST}_{\textit{appear,2\text{-}Z}}}$}\\ \hline
Baseline after Drift: & \multicolumn{3}{ l|}{Avg.Acc. = 85.43\,\%}& \multicolumn{3}{ l|}{Avg.Acc. = 90.29\,\%}\\ \hline
Model & A.Acc & F.Acc & R.Spd & A.Acc & F.Acc & R.Spd \\ \hline
\textit{NN-Patching}$_{\textit{incl,noEE}}$ & 91.72 & 94.63 & 5.3 & 91.79 & 94.67 & --- \\
\textit{NN-Patching}$_{\textit{incl,baseEE}}$ & 93.02 & 94.69 & 2.6 & 93.06 & 94.73 & --- \\
\textit{NN-Patching}$_{\textit{semi,baseEE}}$ & \textbf{93.04} & \textbf{94.7} & \textbf{2.4} & \textbf{93.24} & \textbf{94.78} & --- \\
\textit{NN-Patching}$_{\textit{excl,baseEE}}$ & 92.73 & 94.56 & 2.5 & 92.85 & 94.64 & --- \\
    \hline  
  \end{tabular}

\end{table}


In both experiment series presented here, it is not beneficial to choose the exclusive over the inclusive ensemble. The theoretical advantage can not be preserved, since the error region estimator misclassifies too many instances.

However, the semi-inclusive training results in a significant improvement over the inclusive ensemble. The higher robustness of the model, towards the error region estimator, increases the performance for all evaluation measures. Moreover, in case of very high base classifier capabilities after the drift, the semi-inclusive ensemble often outperforms the inclusive ensemble and the inclusive patch without error region estimator.   

\paragraph{On the Benefit of Base Classifier Usage to leverage Recovery Speed.}
Right after the concept drift, the patch network is still adapting to the new concept. The classification capabilities of the patch are limited, since more training is required. Hence, shortly after the occurrence of a concept drift, the base classifier is strongest in comparison to the patch classifier. The use of the base classifier is more beneficial during this time than after further training of the patch. This is because, the patch capabilities increase after every seen batch, whereas the base classifier capabilities remain constant. However, if the error region estimator is not capable of predicting the error region of the base classifier sufficiently well shortly after the drift, the early advantage of the base classifier over the patch would be negated, due to inadequate ensemble usage.
 
\begin{figure}[H]
	\centering
		\subfigure[\label{fig:thesis_patch_vs_ee_convergence_speed_conv_nist_appear_7_run_4} $\mathbf{\textit{NIST}_{\textit{appear,4\text{-}Z}}}$]{\includegraphics[scale=0.35]{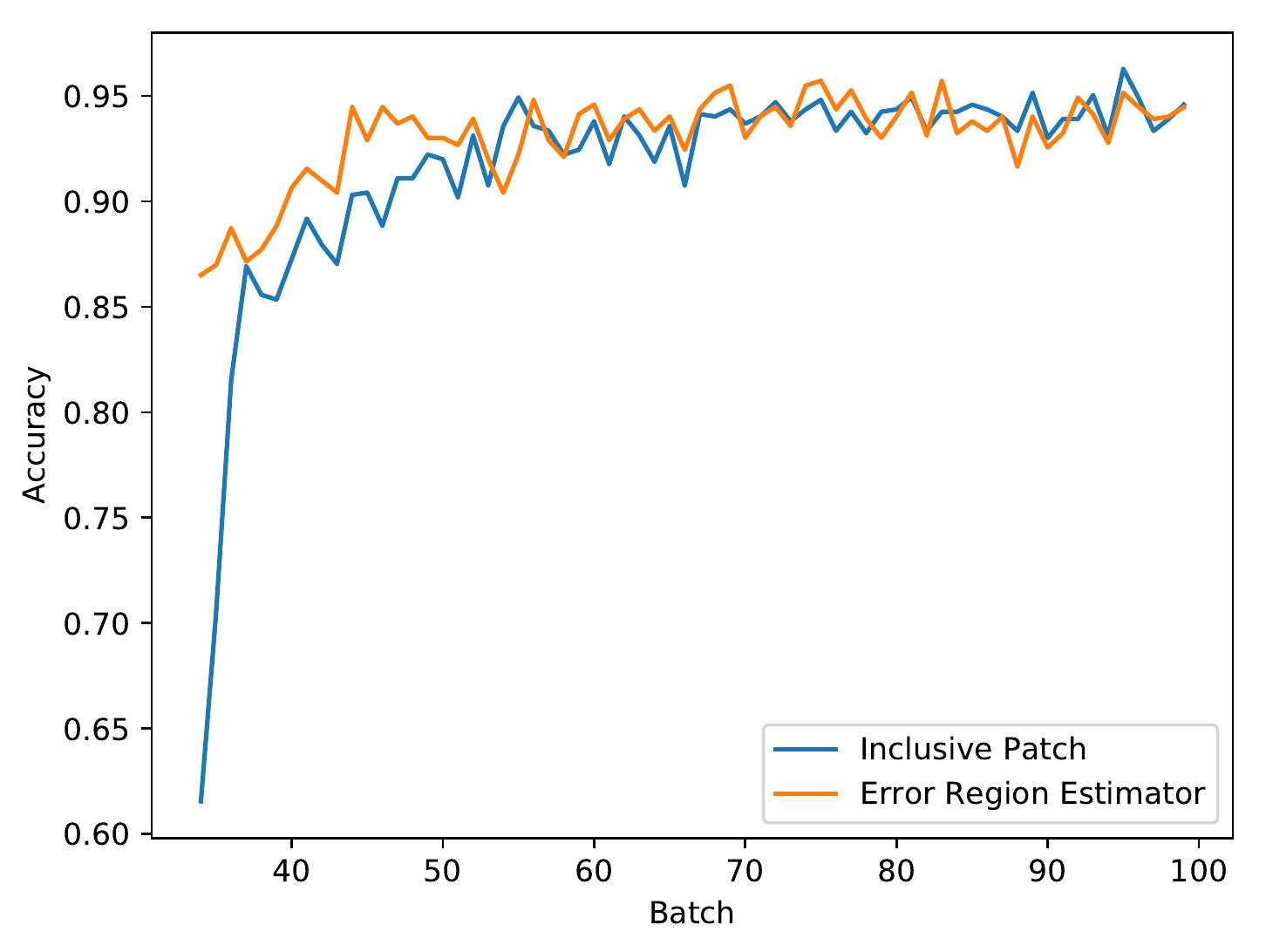}}
		\subfigure[\label{fig:thesis_patch_vs_ee_convergence_speed_conv_nist_appear_run_4} $\mathbf{\textit{NIST}_{\textit{appear}}}$]{\includegraphics[scale=0.35]{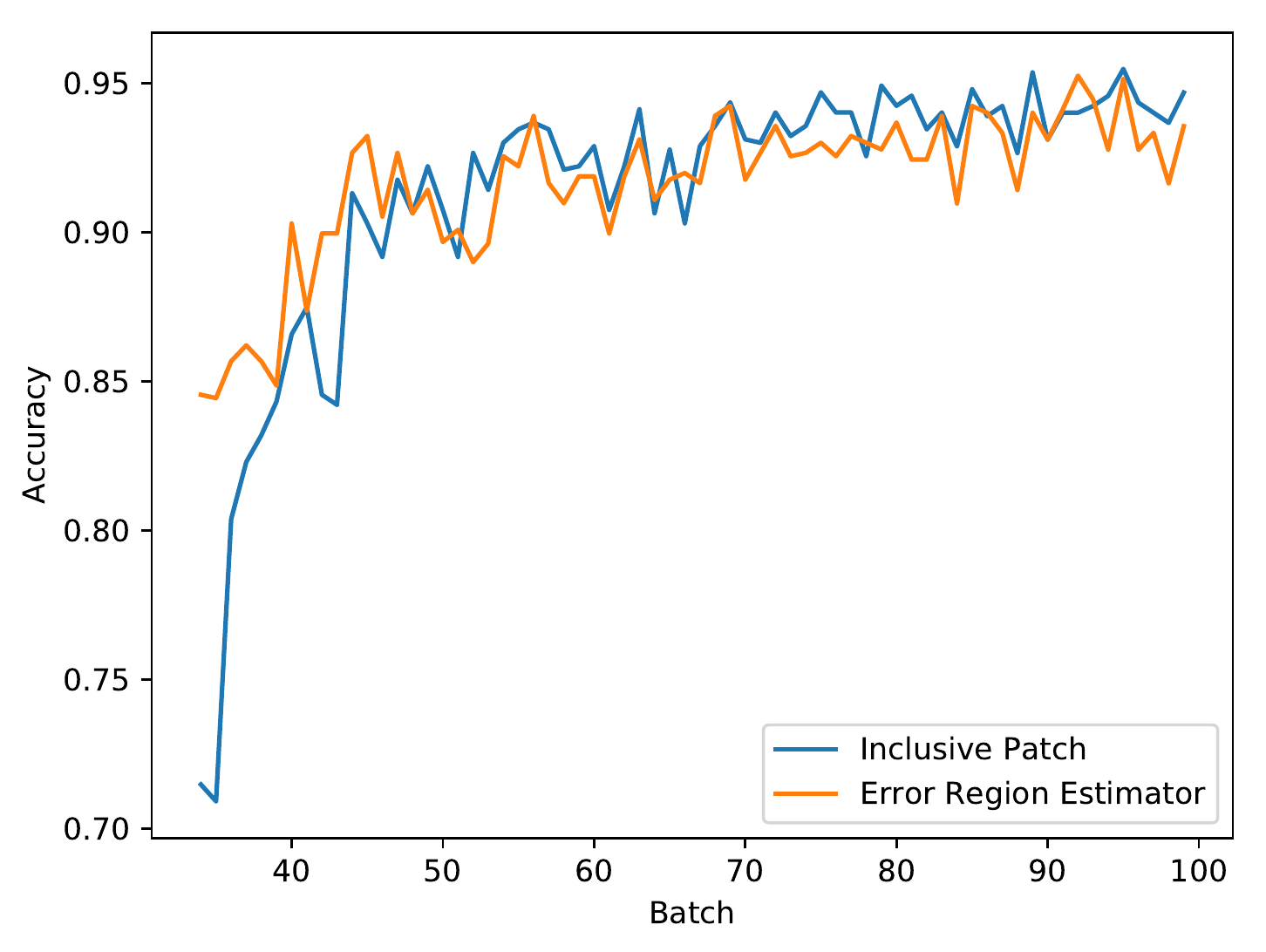}}
\caption{\textbf{Comparison of accuracy curves for an inclusive patch and an error region estimator network.} The concept drift occurred one batch before the first shown datapoint. The accuracy of the inclusive patch is obtained by predicting the labels for each batch and compare them to the true labels. The error region estimator accuracy is obtained by predicting the error region for each batch. The prediction is then compared with the true error region of the base classifier. The base classifier is a CNN in both cases.}
\label{fig:Comparison of accuracy curves for an inclusive patch and an error region estimator network}
\end{figure}

Therefore, we need to further review the prediction quality of the error estimator network, especially on the first batches after the drift. In Figure~\ref{fig:Comparison of accuracy curves for an inclusive patch and an error region estimator network} we show accuracy curves for patch and error estimator network. In Figure~\ref{fig:thesis_patch_vs_ee_convergence_speed_conv_nist_appear_7_run_4} the patch network has an accuracy of 61.5\,\% after training on the batch, where the concept drift occurred (i.e the first shown datapoint). In contrast, the error region estimator starts with an accuracy of over 85\,\%. 

In Figure~\ref{fig:thesis_patch_vs_ee_convergence_speed_conv_nist_appear_run_4} the error estimator also achieves a significantly higher accuracy on the first batches after the drift. We propose that this phenomenon is due to two reasons. First, the error region estimator has to solve only a two class problem instead of the 36 class problem for the patch network. Hence, the baseline for the error estimator network is higher than the baseline for the patch network. Second, the baseline for the error estimator is further increased as a result of an unevenly distributed error region problem (i.e. predicting majority class). If the error region and no error region priors are unevenly distributed in favour of the no error region class, directing the majority of instances to the base network for classification already results in a satisfying solution. 

The dependency of the error estimator performance on the batches right after the concept drift with respect to the base classifier accuracy after the drift is shown in Figure~\ref{fig:Final accuracy, starting accuracy and baseline accuracy for the error estimator network over the base classifier accuracy after the drift}. The graph shows that the start accuracy (i.e. average accuracy on the first 5 batches after the drift) of the error estimator qualitative follows the course of the baseline. The higher the baseline, the better the start accuracy of the error estimator.

The error region estimator often converges faster than the patch network. This observation combined with the fact that the base classifier is strongest in comparison to the patch shortly after the drift, explains why robust ensembles, under preconditions on the base classifier capabilities after the drift, outperform \textit{NN-Patching}$_{\textit{incl,noEE}}$ in recovery speed and average accuracy.

\paragraph{On the Final Accuracy of the Error Estimator Network}
We discussed the fast convergence speed of the error estimator network in comparison to the patch. If we further look at the course of the accuracy graph (Fig.~\ref{fig:Comparison of accuracy curves for an inclusive patch and an error region estimator network}), we notice that the error estimator at the end of the data stream shows comparable (Fig.~\ref{fig:thesis_patch_vs_ee_convergence_speed_conv_nist_appear_7_run_4}) or slightly worse accuracy (Fig.~\ref{fig:thesis_patch_vs_ee_convergence_speed_conv_nist_appear_run_4}) in comparison to the patch network. 

However, the two class error region problem should be easier to solve than the 36 class classification task (i.e. digits and uppercase letters), if the inter-class distances are comparable. Since this is not the case, the error region problem must be significantly harder to decide than a decision task between two digits or letters. 

\begin{figure}[H]
	\centering
		\subfigure[\label{fig:error_est_performance_and_baseline_MNIST} On altered $\mathbf{\textit{MNIST}_{\textit{appear}}}$ datasets]{\includegraphics[scale=0.35]{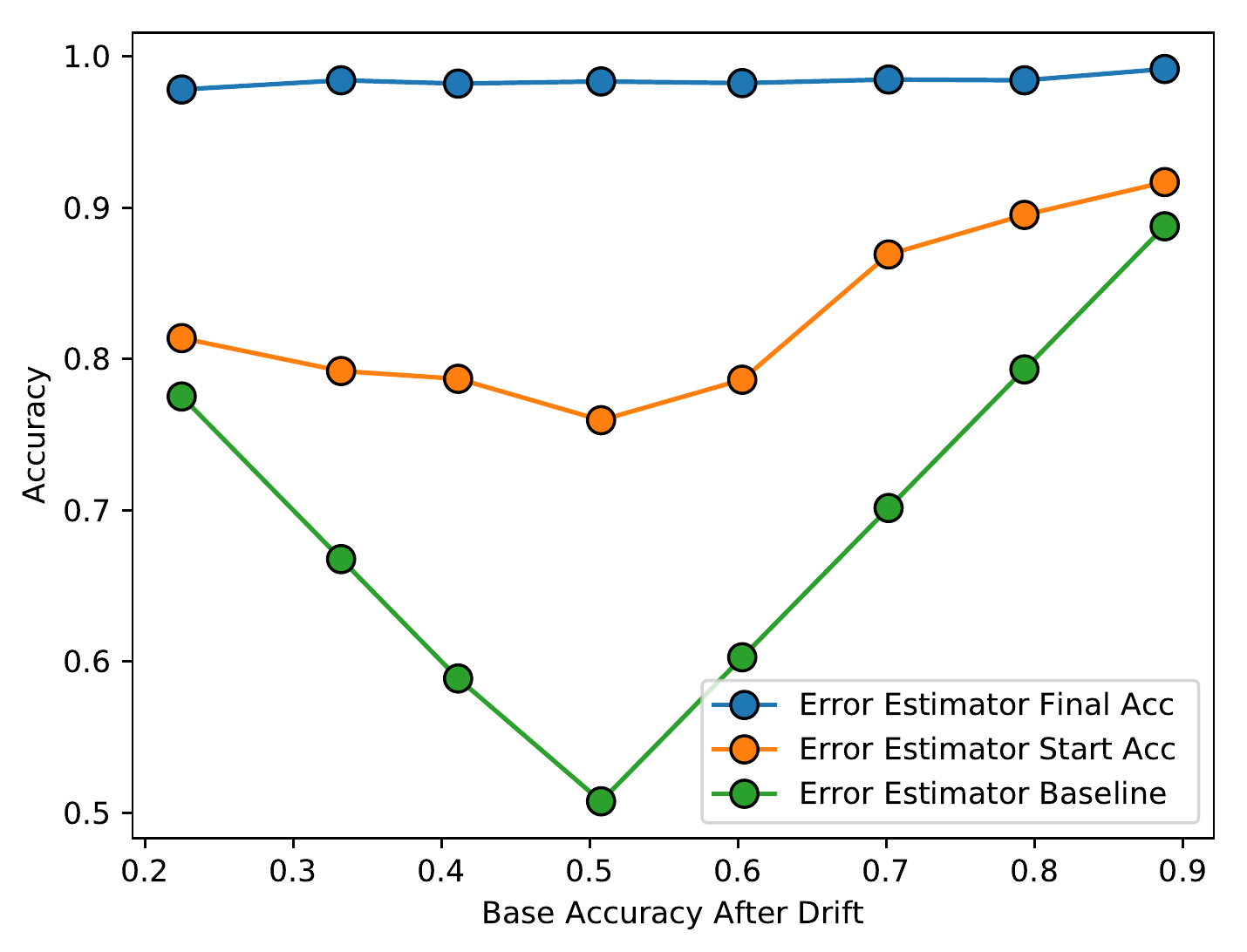}}
		\subfigure[\label{fig:error_est_performance_and_baseline_NIST} On altered $\mathbf{\textit{NIST}_{\textit{appear}}}$ datasets]{\includegraphics[scale=0.35]{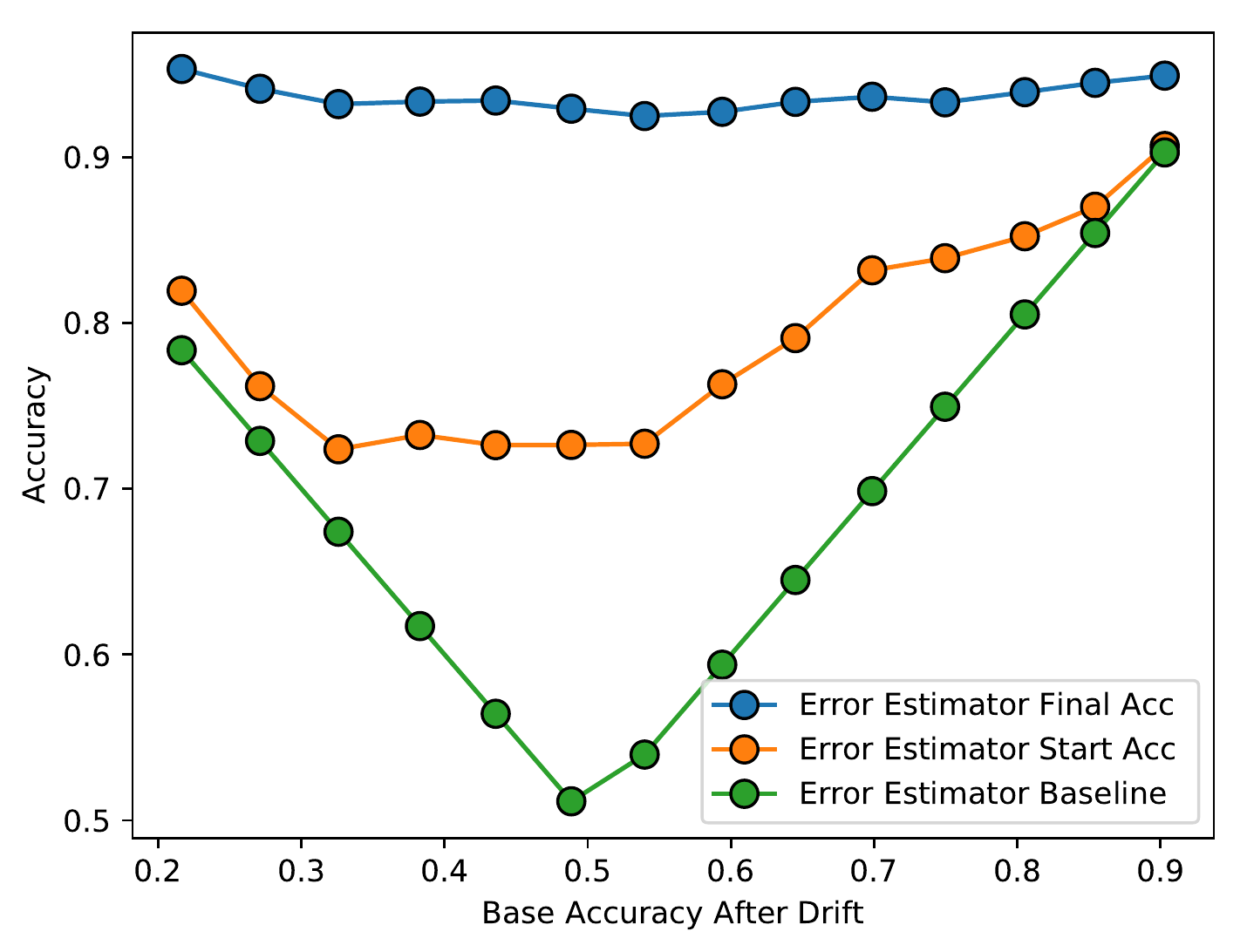}}
\caption{\textbf{Final accuracy, start accuracy and baseline accuracy for the error estimator network over the base classifier accuracy after the drift.} Each data point in an accuracy progression refers to one dataset. The final accuracy is obtained as the average accuracy over the last five batches. The start accuracy is calculated as the average accuracy on the first five batches after the drift. The baseline of the error estimator is the base classifier accuracy after the drift. If the base classifier accuracy is less than 0.5, the baseline is calculated as one minus that value, since this is the accuracy of an error estimator, which always predicts the majority class. All presented values are averaged over 10 runs. The base classifier is a CNN in both cases.}
\label{fig:Final accuracy, starting accuracy and baseline accuracy for the error estimator network over the base classifier accuracy after the drift}
\end{figure}

Figure~\ref{fig:Final accuracy, starting accuracy and baseline accuracy for the error estimator network over the base classifier accuracy after the drift} shows the progression of the final accuracy of the error region estimator network for different base classifier accuracies after the drift. We already discussed the significant dependency of the start accuracy with respect to the baseline. However, the final accuracy shows a low dependence with respect to the base classifier accuracy. From the figures, the final accuracy of the error estimator might seem satisfactory. However, the accuracy of the inclusive patch without error estimation is also high. Our models reach final accuracies significantly over 90\,\% for all datasets. The performance of the error estimator on the overall performance of the classifier ensemble is a linear relation. Even if the error estimator predicts erroneous on merely a low percentage of instances, the resulting misclassifications make the ensemble usage disadvantageous.

The experiments (Tab.~\ref{tab:Comparison of the inclusive, semi-exclusive, exclusive ensemble and the stand-alone inclusive patch on altered $MNIST_{appear}$ datasets with different base classifier capabilities after the drift},~\ref{tab:Comparison of the inclusive, semi-exclusive, exclusive ensemble and the stand-alone inclusive patch on altered $NIST_{appear}$ datasets with different base classifier capabilities after the drift}) show that the ensemble usage is still beneficial under some preconditions, but the theoretical advantage of the exclusive and semi-exclusive training over \textit{NN-Patching}$_{\textit{incl,noEE}}$ is lost almost completely due to erroneous decisions by the error estimator.

\paragraph{High Base Classifier Capability after Drift increases Ensemble Robustness against erroneous Error Estimators} 
The difference in final accuracy between \\ \textit{NN-Patching}$_{\textit{incl,noEE}}$, \textit{NN-Patching}$_{\textit{incl,baseEE}}$, and \textit{NN-Patching}$_{\textit{semi,baseEE}}$ is not significant for base classifier accuracies greater than 75\,\%. This indicates, that the performance increase through solving only a sub-problem in the instance space for semi-exclusive training is small. The theoretical performance increase is lost due to insufficiently accurate error region estimation. The notable advantage of \textit{NN-Patching}$_{\textit{incl,baseEE}}$ and \textit{NN-Patching}$_{\textit{semi,baseEE}}$ in average accuracy comes from the faster recovery because of the ensemble usage.

These advantages of the classifier ensembles can only be observed for high base network capabilities after the occurrence of the concept drift. Since the difference in final accuracy between \textit{NN-Patching}$_{\textit{incl,baseEE}}$ and \textit{NN-Patching}$_{\textit{semi,baseEE}}$ is marginal, we can exclude the benefit of solving a sub-problem as the source of the advantage. 

The real reason is, that a stronger base classifier increases the robustness against an erroneous error region estimation. The inclusive or semi-inclusive training enhanced the robustness of the ensemble in comparison to exclusive training, since we increased the sub-region of the instance space, where the classification capabilities of the patch are sufficient. Hence, if an instance is incorrectly directed to the patch (i.e. the base network could correctly classify the instance), the probability of a successful classification by the patch, regardless of the erroneous error region prediction, is increased.

If an instance is directed to the base network, the probability of a successful classification is independent from the capability of the patch. In this case, the success or failure of the classification is merely dependent on the capabilities of the base classifier. Thus, a higher coverage of the instance space by the base classifier increases the probability of a successful classification and therefore the robustness of the ensemble.

\subsection{Modelling the Error Region of the Patch Network}
\label{sec:modellingerrorregions}
Modelling the error region of the base classifier is a difficult task for the error estimator. In this section, we discuss the option of modelling the error region of the patch instead. 

There are theoretical advantages and disadvantages for both options. The error region of the base classifier is a stationary concept, whereas the error region of the patch is changing over time. In general, it should be advantageous to train on data following a stationary concept.

However, the base classifier has a much higher model complexity in comparison to the error estimator. Therefore, the expressiveness of the error estimator network might not be sufficient to represent such a complex concept satisfactorily. The patch and the error estimator, on the other hand, share a comparable expressiveness. Hence, it is a fair assumption that it is easier for the error estimator to model the error region of the patch. 

Moreover, right after a concept drift the patch has low classification capabilities. In Section~\ref{Effect of different Base Network Capabilities after Drift on inclusive, exclusive and semi-exclusive Patching Performance} we discussed the benefit of using the base classifier especially during the first batches after the occurrence of a concept drift. Since the error region of the patch is large in this phase, more instances are directed to the base network for classification. After further training, the error region of the patch network shrinks and therefore more instances are classified by the patch.

\begin{figure}[H]
\centering
\includegraphics[scale=0.70]{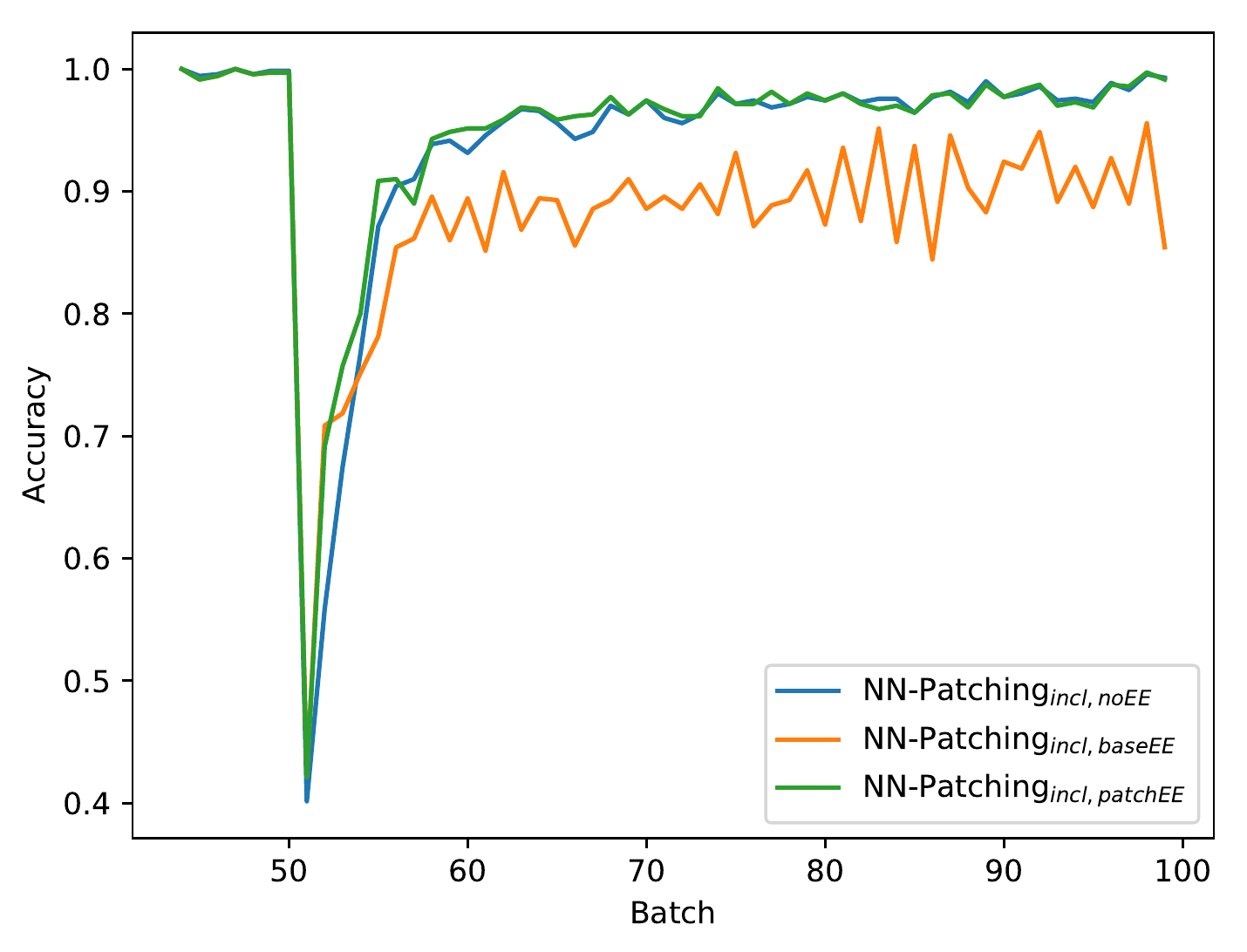}
\caption{\textbf{Comparison of base classifier error estimator, patch error estimator and stand-alone patch without error estimator.} The figure shows the accuracy difference between modelling the patch error region, the base classifier error region, and the performance of the patch without error estimator. The concept drift occurs at batch 50. The patches are trained inclusively. The used dataset is $\mathbf{\textit{MNIST}_{\textit{remap}}}$ and the base classifier is a CNN.} 
\label{fig:Comparison of base classifier error estimator, patch error estimator and stand-alone patch without error estimator}
\end{figure}

A problem of modelling the base classifier error region is that not enough instances are directed to the patch network in the final phase. Hence, ensembles with a base network error estimator often show a significantly decreased final accuracy in comparison to \textit{NN-Patching}$_{\textit{incl,noEE}}$. In Figure~\ref{fig:Comparison of base classifier error estimator, patch error estimator and stand-alone patch without error estimator} we show the accuracy curves for an inclusive ensemble with base classifier error estimator, an inclusive ensemble with patch error estimator, and a stand-alone patch without error estimator. The graph shows a deficit in final accuracy for \textit{NN-Patching}$_{\textit{incl,baseEE}}$ due to the base network error estimator. 

Furthermore, the accuracy curve of \textit{NN-Patching}$_{\textit{incl,baseEE}}$ shows an oscillating behavior. Although, the error region of the base network follows a stationary concept, the error estimator is not capable of modelling  the error region well. This is another indication of the difficulty of the base error region task. In contrast, the patch error estimator does not suffer from that problem and shows comparable performance with \textit{NN-Patching}$_{\textit{incl,noEE}}$ in final accuracy.


\clearpage
\section{Initializing Neural Networks}
\label{On the Performance Difference between Transfer Weights and Random Initialisation}
One difference between neural network patching and transfer learning methods, such as \textit{Freezing}, is the initialization. In neural network patching the patch and error estimator are initialized with random weights (i.e Glorot Uniform Initialization), whereas \textit{Freezing} adopts the weights from the base classifier. 

In this section, we investigate the performance difference between the transferred weights in comparison to random initialization. The used model for this series of experiments is \textit{Freezing}. In \textit{Freezing} only the last layers of of network are trainable. The first layers of the network are non-trainable and can be ignored during backpropagation. More precisely, every layer up to and including the engagement layer is non-trainable. The rest of the network is trainable. The used engagement layers can be checked in Table~\ref{Engagement Layer Choice for further Experiments}. 

We evaluate two variants of \textit{Freezing}. The original \textit{Freezing} approach, where all weights are inherited from the base classifier, and a second variant, where all weights up to and including the engagement layer are inherited from the base classifier. However, in the second variant the trainable layers do not adopt the weights, instead they are randomly initialized using Glorot Uniform Initialization. All values in the Tables~\ref{FC-NN random transfer}-\ref{ResNet random transfer} are averaged over 10 runs with varying random seed. 

\subsection{Fully-Connected Base Network}
In Table~\ref{FC-NN random transfer} the results for \textit{Freezing} with transfer weights and \textit{Freezing} with random initialization of the trainable layers are presented for the FC-NN base classifier. We recall that the engagement layer for the FC-NN is the first FC-layer of the network for \textit{MNIST} datasets and the second FC-layer for \textit{NIST} datasets. Most of the layers are trainable and therefore a large part of the network is randomly initialised in case of the \textit{Freezing} variant with random initialization.

\begin{table}[h]
\caption{Performance difference between transfer weights and random initialization for FC-NN base classifiers.}
\label{FC-NN random transfer}

\smaller
\begin{tabular}{ |l | l l l  | l l l | l l l |}
    \hline
		\multicolumn{10}{|c|}{Fully-Connected Neural Network}\\ \hline
		Dataset: & \multicolumn{3}{ c|}{$\mathbf{\textit{MNIST}_{\textit{appear}}}$}& \multicolumn{3}{ c|}{$\mathbf{\textit{MNIST}_{\textit{flip}}}$}& \multicolumn{3}{ c|}{$\mathbf{\textit{MNIST}_{\textit{remap}}}$}\\ \hline
		Model & A.Acc & F.Acc & R.Spd & A.Acc & F.Acc & R.Spd & A.Acc & F.Acc & R.Spd  \\ \hline
Transfer weights & \textbf{79.88} & \textbf{94.58} & \textbf{32.0} & \textbf{83.91} & \textbf{94.23} & \textbf{15.9} & \textbf{88.47} & \textbf{94.93} & \textbf{13.7} \\
Random weights   & 79.87 & 94.41 & 33.1 & 82.4 & 92.89 & 18.5 & 86.9 & 94.35 & 15.6 \\    \hline 
	\end{tabular} \begin{tabular}{ |l | l l l  | l l l |}
    \hline
		Dataset: & \multicolumn{3}{ c|}{$\mathbf{\textit{MNIST}_{\textit{rotate}}}$}& \multicolumn{3}{ c|}{$\mathbf{\textit{MNIST}_{\textit{transfer}}}$}\\ \hline
		Model & A.Acc & F.Acc & R.Spd & A.Acc & F.Acc & R.Spd   \\ \hline
Transfer weights & \textbf{69.08} & \textbf{70.12}  & --- & 71.22 & 93.25 & 28.2 \\
Random weights   & 66.51 & 68.62 & --- & \textbf{77.52} & \textbf{94.41} & \textbf{20.3} \\		\hline  
\end{tabular}
\\ \begin{tabular}{ |l | l l l  | l l l | l l l |}
    \hline
		Dataset: & \multicolumn{3}{ c|}{$\mathbf{\textit{NIST}_{\textit{appear}}}$}& \multicolumn{3}{ c|}{$\mathbf{\textit{NIST}_{\textit{flip}}}$}& \multicolumn{3}{ c|}{$\mathbf{\textit{NIST}_{\textit{remap}}}$}\\ \hline
		Model & A.Acc & F.Acc & R.Spd & A.Acc & F.Acc & R.Spd & A.Acc & F.Acc & R.Spd  \\ \hline
Transfer weights & \textbf{78.26}  & \textbf{85.6} & \textbf{30.0} & \textbf{58.54} & \textbf{78.28} & \textbf{58.4} & \textbf{64.58} & \textbf{79.29}  & -- \\
Random weights & 75.18 & 84.65 & 34.1 & 57.54 & 78.06 & 58.7 & 64.08 & 77.46 & -- \\		\hline 
	\end{tabular} 
\begin{tabular}{ |l | l l l | l l l |} 
    \hline
		Dataset: & \multicolumn{3}{ c|}{$\mathbf{\textit{NIST}_{\textit{rotate}}}$} & \multicolumn{3}{ c|}{$\mathbf{\textit{NIST}_{\textit{transfer}}}$}\\ \hline
		Model & A.Acc & F.Acc & R.Spd & A.Acc & F.Acc & R.Spd  \\ \hline
Transfer weights & \textbf{36.51} & \textbf{37.77}  & --- & 36.57 & 55.86  & --- \\
Random weights & 34.35 & 35.12 & --- & \textbf{48.04} & \textbf{66.27} & --- \\
		\hline  
	\end{tabular}
	
\end{table}

The \textit{Freezing} variant with transfer weights shows superior results on all datasets except $\mathbf{\textit{MNIST}_{\textit{transfer}}}$ and $\mathbf{\textit{NIST}_{\textit{transfer}}}$ for the FC-NN base classifier. In case of the transfer-datasets, the base network classifies all instances incorrectly after the occurrence of the concept drift. The use of transfer weights highly decreases performance for all evaluation measures. 

On all other datasets the usage of transfer weights is preferable. For $\mathbf{\textit{MNIST}_{\textit{appear}}}$ the performance increase is rather small, but besides for this particular dataset the performance increase of transfer weights over random weights is large. 

\subsection{Convolutional Base Network}
In Table~\ref{CNN random transfer} the results are shown for the convolutional base architectures. The engagement layer of the CNN is the pooling layer for \textit{MNIST} and the last convolutional layer for \textit{NIST}. The remaining trainable layers of CNN architecture for \textit{NIST} are 
\begin{verbatim}
MaxPooling - Dropout(0.25) - FC(256) - 
Dropout(0.5) - Softmax.
\end{verbatim}
Since the pooling layer has no parameters, only the FC-layer and the output layer is trained. For the CNN base classifier on \textit{MNIST} the trainable network part is Dropout(0.25)~-~FC(128)~-~Dropout(0.5)~-~Softmax. In both cases the trainable network part is closely related to the optimal patch architecture, which we elaborated on in previous sections. 

\begin{table}[h]

\caption{Performance difference between transfer weights and random initialization for CNN base classifiers.}
\label{CNN random transfer}

\smaller
\begin{tabular}{ |l | l l l  | l l l | l l l |}
    \hline
		\multicolumn{10}{|c|}{Convolutional Neural Network}\\ \hline
		Dataset: & \multicolumn{3}{ c|}{$\mathbf{\textit{MNIST}_{\textit{appear}}}$}& \multicolumn{3}{ c|}{$\mathbf{\textit{MNIST}_{\textit{flip}}}$}& \multicolumn{3}{ c|}{$\mathbf{\textit{MNIST}_{\textit{remap}}}$}\\ \hline
		Model & A.Acc & F.Acc & R.Spd & A.Acc & F.Acc & R.Spd & A.Acc & F.Acc & R.Spd  \\ \hline
Transfer weights & \textbf{92.82} & \textbf{98.24} & \textbf{7.7} & 87.22 & 95.91 & 13.9 & 91.51 & 97.55  & 7.4 \\
Random weights   & 92.27 & 98.1 &  8.3 & \textbf{93.51} & \textbf{97.48} & \textbf{7.4} & \textbf{94.12} & \textbf{98.29} & \textbf{5.2} \\  
\hline 
	\end{tabular}
 
\begin{tabular}{ |l | l l l  | l l l |}
    \hline
		Dataset: & \multicolumn{3}{ c|}{$\mathbf{\textit{MNIST}_{\textit{rotate}}}$}& \multicolumn{3}{ c|}{$\mathbf{\textit{MNIST}_{\textit{transfer}}}$}\\ \hline
		Model & A.Acc & F.Acc & R.Spd & A.Acc & F.Acc & R.Spd   \\ \hline
Transfer weights & \textbf{74.61} & \textbf{79.55}  & --- & 91.08 & 97.92  & 5.7 \\
Random weights & 72.49 & 77.79 & ---- & \textbf{94.03} & \textbf{98.54} & \textbf{4.8} \\\hline  
\end{tabular}
\\ \begin{tabular}{ |l | l l l  | l l l | l l l |}
    \hline
		Dataset: & \multicolumn{3}{ c|}{$\mathbf{\textit{NIST}_{\textit{appear}}}$}& \multicolumn{3}{ c|}{$\mathbf{\textit{NIST}_{\textit{flip}}}$}& \multicolumn{3}{ c|}{$\mathbf{\textit{NIST}_{\textit{remap}}}$}\\ \hline
		Model & A.Acc & F.Acc & R.Spd & A.Acc & F.Acc & R.Spd & A.Acc & F.Acc & R.Spd  \\ \hline
Transfer weights & \textbf{91.4} & \textbf{94.95}  & \textbf{7.5} & 71.3 & 88.4 & 36.8 & 85.92  & 95.21  & 13.7 \\
Random weights & 90.97 & 94.63 & 8.4 & \textbf{88.76} & \textbf{92.89} & \textbf{9.6} & \textbf{92.24} & \textbf{96.55} & \textbf{7.2} \\	\hline 
	\end{tabular} \begin{tabular}{ |l | l l l  | l l l |} 
    \hline
		Dataset: & \multicolumn{3}{ c|}{$\mathbf{\textit{NIST}_{\textit{rotate}}}$} & \multicolumn{3}{ c|}{$\mathbf{\textit{NIST}_{\textit{transfer}}}$}\\ \hline
		Model & A.Acc & F.Acc & R.Spd & A.Acc & F.Acc & R.Spd  \\ \hline
Transfer weights & 52.42 & 58.48  & --- & 80.28 & 91.46 & 32.4 \\
Random weights & \textbf{54.68} & \textbf{60.73} & --- & \textbf{85.83} & \textbf{93.04} & \textbf{22.5} \\
		\hline  
	\end{tabular}
	
\end{table}

The CNN result table shows that transfer weights are only beneficial on $\mathbf{\textit{MNIST}_{\textit{appear}}}$, $\mathbf{\textit{NIST}_{\textit{appear}}}$ and $\mathbf{\textit{MNIST}_{\textit{rotate}}}$. On the majority of datasets using transfer weights results in a performance decrease. The performance degradation due to transfer weights is particularly large for $\mathbf{\textit{MNIST}_{\textit{flip}}}$, $\mathbf{\textit{NIST}_{\textit{flip}}}$, $\mathbf{\textit{MNIST}_{\textit{remap}}}$, $\mathbf{\textit{NIST}_{\textit{remap}}}$, $\mathbf{\textit{MNIST}_{\textit{transfer}}}$ and $\mathbf{\textit{NIST}_{\textit{transfer}}}$.
 
\subsection{Residual Base Network}
We present the results for the ResNet base archetype (Tab.~\ref{ResNet random transfer}). The engagement layer for the ResNet architecture on \textit{NIST} and \textit{MNIST} is the second to last layer of the network, the resulting trainable part is \begin{verbatim}Dropout(0.5)~-~Softmax\end{verbatim}. The trainable network part only consists of one trainable layer, hence this is a linear classifier. 

\begin{table}[h]
\smaller
\begin{tabular}{ |l | l l l  | l l l | l l l |}
    \hline
		\multicolumn{10}{|c|}{Residual Neural Network}\\ \hline
		Dataset: & \multicolumn{3}{ c|}{$\mathbf{\textit{MNIST}_{\textit{appear}}}$}& \multicolumn{3}{ c|}{$\mathbf{\textit{MNIST}_{\textit{flip}}}$}& \multicolumn{3}{ c|}{$\mathbf{\textit{MNIST}_{\textit{remap}}}$}\\ \hline
		Model & A.Acc & F.Acc & R.Spd & A.Acc & F.Acc & R.Spd & A.Acc & F.Acc & R.Spd  \\ \hline
Transfer weights & \textbf{89.2} & \textbf{96.42}  & \textbf{12.1} & 63.01 & 75.9 & ---   & 65.26 & 83.07  & 44.2 \\
Random weights   & 87.78 & 94.86 & 17.4 & \textbf{90.75} & \textbf{95.77 }& \textbf{15.7} & \textbf{82.77} & \textbf{91.06} & \textbf{18.6} \\  
\hline 
	\end{tabular}
	\begin{tabular}{ |l | l l l  | l l l |}
    \hline
		Dataset: & \multicolumn{3}{ c|}{$\mathbf{\textit{MNIST}_{\textit{rotate}}}$}& \multicolumn{3}{ c|}{$\mathbf{\textit{MNIST}_{\textit{transfer}}}$}\\ \hline
		Model & A.Acc & F.Acc & R.Spd & A.Acc & F.Acc & R.Spd   \\ \hline
Transfer weights & 62.84 & 67.61  & --- & 73.71 & 91.54  & 30.6 \\
Random weights & \textbf{63.06} & \textbf{68.73} & --- &\textbf{80.18} & \textbf{92.59} & \textbf{14.6} \\ \hline  
\end{tabular}
\\ 
\begin{tabular}{ |l | l l l  | l l l | l l l |}
    \hline
		Dataset: & \multicolumn{3}{ c|}{$\mathbf{\textit{NIST}_{\textit{appear}}}$}& \multicolumn{3}{ c|}{$\mathbf{\textit{NIST}_{\textit{flip}}}$}& \multicolumn{3}{ c|}{$\mathbf{\textit{NIST}_{\textit{remap}}}$}\\ \hline
		Model & A.Acc & F.Acc & R.Spd & A.Acc & F.Acc & R.Spd & A.Acc & F.Acc & R.Spd  \\ \hline
Transfer weights & \textbf{87.44} & \textbf{91.26}  & \textbf{11.3} & 49.43 & 72.61 & --- & 53.94 & 70.3 & 43.1 \\
Random weights   & 84.29 & 89.14 & 15.3  & \textbf{85.95} & \textbf{90.1} & \textbf{13.8} & \textbf{60.58 }& \textbf{80.9} & \textbf{27.6} \\ \hline 
	\end{tabular} \begin{tabular}{ |l | l l l  | l l l |} 
    \hline
		Dataset: & \multicolumn{3}{ c|}{$\mathbf{\textit{NIST}_{\textit{rotate}}}$} & \multicolumn{3}{ c|}{$\mathbf{\textit{NIST}_{\textit{transfer}}}$}\\ \hline
		Model & A.Acc & F.Acc & R.Spd & A.Acc & F.Acc & R.Spd  \\ \hline
Transfer weights & 45.64 & 51.72  & --- & 66.51 & 80.27  & --- \\
Random weights & \textbf{47.25} & \textbf{52.75} & --- & \textbf{72.81} & \textbf{82.59} & --- \\
		\hline  
	\end{tabular}
	\caption{Performance difference between transfer weights and random initialization for ResNet base classifiers.}
\label{ResNet random transfer}
\end{table}

In case of the ResNet base classifier, the random initialization is preferable for all datasets except $\mathbf{\textit{MNIST}_{\textit{appear}}}$ and $\mathbf{\textit{NIST}_{\textit{appear}}}$. We propose that due to the low complexity of the trainable network part, the \textit{Freezing} model for our ResNets are particularly prone to inadequate initialization, since the loss function contains many poor local minima~\cite{DBLP:journals/corr/ChoromanskaHMAL14}. Furthermore, the improper initialization increases the probability of finding one of the poor local minima.  

\subsection{Conclusion on Transfer Weights versus Random Initialization}
We notice that the benefit of transfer weights or random initialization is dependent on the base classifier accuracy after the drift.
The untrained base classifier shows the highest accuracy after the drift on $\mathbf{\textit{NIST}_{\textit{appear}}}$ and $\mathbf{\textit{MNIST}_{\textit{appear}}}$, which are also the datasets where \textit{Freezing} with transfer weights is most beneficial. 
On $\mathbf{\textit{NIST}_{\textit{transfer}}}$ and $\mathbf{\textit{MNIST}_{\textit{transfer}}}$ the base classifier accuracy after the drift is 0\,\%, hence on these datasets random initialization always increases performance.

We conclude, that the accuracy of the untrained base classifier after the drift is a good indication, whether a random initialization or a weight transfer is preferable. 

Finally, we discuss what the findings of this section mean in terms of neural network patching. The trainable part of the CNNs and our patch architecture are comparable. Therefore, these results are a good indication on the effect of a previous training of the patch network on instances before the concept drift. If the base classifier strength highly decreases after the occurrence of a drift, a random initialization of the patch network is preferable. However, if the previous concept and the new concept after the drift are sufficiently related, then pre-training on data following original concept results in a performance increase. 

A simple approach to utilize this knowledge, is to train the patch on the instances from the previous concept until the patch accuracy is saturated. Then on occurrence of a concept drift, the base classifier accuracy on the new concept can be obtained after the availability of the labels. If the base network accuracy is high enough (our experiments indicate this is the case if the base accuracy is greater than 50\,\%), the pre-trained weights are used. Correspondingly, if the base classifier accuracy is too low, the pre-trained weights are discarded by reinitializing the patch according to Glorot Uniform Initialization.

\clearpage
\section{Comparison of Neural Network Patching and Transfer Learning Methods}
\label{Final Results}
In this section, we compare NN-Patching variants and transfer learning methods. Overall we evaluated 10 different models. Among the models are seven neural network patching variants, two transfer learning methods, and the baseline. The experiments are conducted for all \textit{NIST} and \textit{MNIST} datasets. Moreover, each dataset is evaluated with a FC-NN, a CNN and ResNet as base classifier. Each base classifier dataset combination is executed 10 times with varying random seed. All values are averaged over these 10 runs. Due to the large amount of information, the tables with the exact results are presented in the appendix (Appendix~\ref{Appendix A}). The results are organized in a total of six tables. For each base classifier archetype respectively a table for the results on \textit{NIST} and \textit{MNIST} is presented. The patch and error estimator network architecture is \begin{verbatim}Input - Dropout(0.25) - FC(512) - Dropout(0.5) - Output.\end{verbatim} The used engagement layers are specified in Table~\ref{Engagement Layer Choice for further Experiments}. The average standard deviation for the experiments is shown in Table~\ref{Average standard deviation by dataset}.

\begin{table}[!htbp]
\begin{centering}
  \begin{tabular}{ | l | l  l  l  l  l |}
    \hline
		\multicolumn{6}{|c|}{Average Standard Deviation}\\ \hline
		\multicolumn{1}{|l|}{Datasets}& A.Acc & F.Acc & R.Spd & Ad.Rk & F.Rk\\ \hline
$\mathbf{\textit{MNIST}_{\textit{appear}}}$ &0.16 & 0.08 & 0.85 & 0.24 & 0.32 \\
$\mathbf{\textit{MNIST}_{\textit{flip}}}$&0.25 & 0.2 & 0.59 & 0.13 & 0.17\\
$\mathbf{\textit{MNIST}_{\textit{remap}}}$&0.44 & 0.38 & 0.85 & 0.24 & 0.19\\
$\mathbf{\textit{MNIST}_{\textit{rotate}}}$&0.49 & 0.44 & --- & 0.26 & 0.26\\
$\mathbf{\textit{MNIST}_{\textit{transfer}}}$&0.19 & 0.04 & 0.46 & 0.28 & 0.32\\ \hline
$\mathbf{\textit{NIST}_{\textit{appear}}}$&0.13 & 0.17 & 0.49 & 0.23 & 0.25\\
$\mathbf{\textit{NIST}_{\textit{flip}}}$&0.23 & 0.17 & 0.62 & 0.12 & 0.16\\
$\mathbf{\textit{NIST}_{\textit{remap}}}$&0.31 & 0.16 & 0.49 & 0.22 & 0.31\\
$\mathbf{\textit{NIST}_{\textit{rotate}}}$&0.31 & 0.36 & --- & 0.12 & 0.15\\
$\mathbf{\textit{NIST}_{\textit{transfer}}}$&0.25 & 0.17 & 0.91 & 0.18 & 0.39\\
    \hline  
  \end{tabular}
\caption{\textbf{Average standard deviation by dataset.} The standard deviation is calculated over the 10 runs for each model respectively. Moreover, the standard deviations are averaged over all models and base classifiers. The values for average and final accuracy are stated in percent.}  
\label{Average standard deviation by dataset}
\end{centering}
\end{table}

In Section~\ref{Result Overview}, we discuss the results for each base classifier archetype individually. Moreover, in Section~\ref{Modelling the Patch Error Region versus the Base Classifier Error Region}, we compare the results with respect to differences for modelling the error region of the base network or the patch. Thereafter, we elaborate on differences due to inclusive, exclusive and semi-exclusive patch training (Sec.~\ref{On inclusive, exclusive and semi-exclusive Patch Training with Error Estimator}). In Section~\ref{Performance differences due to Transfer weights and Random Initialization}, the effects of random initialization and transfer weights on the results are discussed. 

If neural networks are subsequently trained on different tasks, the model loses the capability of solving previous tasks. The property of neural networks to forget previously learned information upon learning new information is called catastrophic forgetting. In Section~\ref{Catastrophic Forgetting} we investigate the capability of neural network patching in order to deal with reoccurring concepts, and therefore catastrophic forgetting. At last, we summarize the findings of this section (Sec.~\ref{Summarization}).

\subsection{Result Overview}
\label{Result Overview}
In order to evaluate the results we use meta-tables. These tables can be generated from the full result tables in the Appendix~\ref{Appendix A}. They state the number of times each model shows top~1 performance among the 10 competitors. The meta-tables give us an insightful overview over the results. We present one meta-table for each base network archetype, which summarizes the results for the respective base classifier on \textit{NIST} and \textit{MNIST} datasets. However, in order to discuss more specific phenomena, we have to refer to the full result tables.
\subsubsection{Fully-Connected Base Architecture}
\label{fc1}
In Table~\ref{Number of Top 1 model performances on $NIST$ and $MNIST$ for FC-NN base classifiers} the meta-table for fully-connected base classifiers is given.

\begin{table}[h]
\caption{Number of top~1 model performances on \textit{NIST} and \textit{MNIST} for FC-NN base classifiers.}
\label{Number of Top 1 model performances on $NIST$ and $MNIST$ for FC-NN base classifiers}
\begin{centering}
  \begin{tabular}{ |l | c | c | c | c | c|}
    \hline
			\multicolumn{1}{|l|}{Base Archetype: FC-NN}&\multicolumn{5}{ c|}{Number of Top~1 Performances}\\ \hline
		Model & A.Acc & F.Acc & R.Spd & Ad.Rk & F.Rk\\ \hline
		\textit{NN-Patching}$_{\textit{incl,noEE}}$ & \textbf{7} & \textbf{5} & \textbf{3} & \textbf{6} & 4 \\   
\textit{NN-Patching}$_{\textit{incl,baseEE}}$  & -- & 1 & 1 & -- & -- \\  
\textit{NN-Patching}$_{\textit{semi,baseEE}}$ & 1 & -- & 1 & 1 & -- \\
\textit{NN-Patching}$_{\textit{excl,baseEE}}$ & -- & -- & -- & -- & -- \\
\textit{NN-Patching}$_{\textit{incl,patchEE}}$& 2 & -- & 2 & 1 & -- \\
\textit{NN-Patching}$_{\textit{semi,patchEE}}$& -- & -- & -- & -- & -- \\
\textit{NN-Patching}$_{\textit{excl,patchEE}}$& -- & -- & 1 & -- & -- \\
\textit{Freezing}                    & -- & -- & -- & -- & -- \\
\textit{Base}$_{\textit{update}}$            & 1 & \textbf{5} & 1 & 2 & \textbf{6} \\
    \hline  
  \end{tabular}

\end{centering}
\end{table}

For the FC-NN base classifier the patch without error estimator (\textit{NN-Patching}$_{\textit{incl,noEE}}$) accomplishes most top~1 performances in average accuracy, final accuracy, recovery speed and adaptation rank. However, retraining the whole base classifier (Base$_{update}$) ties the top~1 performances in final accuracy and has most top~1 performances in terms of final rank. 

Strong performances of \textit{Base}$_{\textit{update}}$in final accuracy and final rank are expected due to the large representation power of the base network in comparison to the patch. Therefore, after sufficient training the base classifier can represent the new concept more accurate than the patch network. In contrast, a many-layered network is not suited for fast adaptation to a new concept, hence \textit{Base}$_{\textit{update}}$ often lacks performance in recovery speed and adaptation rank.

\textit{Freezing} never achieves the top performances. The engagement layer for FC-NNs is first (\textit{MNIST}) or second layer (\textit{NIST}) of the network, thus most network layers are trainable. \textit{Freezing} does not lead to a better recovery speed than Base$_{update}$, since the difference in the amount of trainable layers is only one or two layers. Therefore, the trainable network for \textit{Freezing} is still large. Updating all weights in the base classifier is preferable in case of the FC-NN. The higher representation power by additionally training the first network layers is more important than the low benefit of training a shallower network. This is particularly observed for the FC-NN base classifier, since fully-connected layers extract lower quality features than convolutional layers. 

Moreover, sometimes patching variants utilizing an error estimator network show best performance in recovery speed. This is, because using the base classifier predictions is beneficial especially on the first few batches after the concept drift. 

However, the full advantage for the patching variants with an error estimator can not be obtained for FC-NN base classifiers due to the low quality features generated by the fully-connected layers.

\subsubsection{Convolutional Base Architecture}
\label{cnn1}

In contrast to the FC-NN, the convolutional layers in CNNs generate better and more transferable features. The top~1 performances on \textit{NIST} and \textit{MNIST} with CNN base classifiers are stated in Table~\ref{Number of Top 1 model performances on $NIST$ and $MNIST$ for CNN base classifiers}.

\begin{table}[h]
\caption{Number of top~1 model performances on \textit{NIST} and \textit{MNIST} for CNN base classifiers.}
\label{Number of Top 1 model performances on $NIST$ and $MNIST$ for CNN base classifiers}
\begin{centering}
  \begin{tabular}{ |l | c | c | c | c | c|}
    \hline
			\multicolumn{1}{|l|}{Base Archetype: CNN}&\multicolumn{5}{ c|}{Number of Top~1 Performances}\\ \hline
		Model & A.Acc & F.Acc & R.Spd & Ad.Rk & F.Rk\\ \hline
		\textit{NN-Patching}$_{\textit{incl,noEE}}$ & 2 & \textbf{4} & 2 & \textbf{4} & \textbf{4} \\   
\textit{NN-Patching}$_{\textit{incl,baseEE}}$  & 1 & -- & -- & -- & -- \\  
\textit{NN-Patching}$_{\textit{semi,baseEE}}$ & 1 & 2 & \textbf{3} & 1 & -- \\
\textit{NN-Patching}$_{\textit{excl,baseEE}}$ & -- & -- & 1 & -- & -- \\
\textit{NN-Patching}$_{\textit{incl,patchEE}}$& \textbf{4} & 1 & \textbf{3} & 2 & \textbf{4} \\
\textit{NN-Patching}$_{\textit{semi,patchEE}}$& 1 & 3 & 1 & 1 & -- \\
\textit{NN-Patching}$_{\textit{excl,patchEE}}$& -- & -- & -- & -- & -- \\
\textit{Freezing}                    & 1 & 1 & -- & 2 & 1 \\
\textit{Base}$_{\textit{update}}$            & -- & 1 & -- & -- & 1 \\
    \hline  
  \end{tabular}

\end{centering}
\end{table}

The features in the engagement layer of the CNN are higher quality, therefore the patching variants utilizing an error estimator become more viable, since the performance of the error estimator correlates with the quality of features in the engagement layer. The error estimation is required to have a high accuracy, otherwise the amount of follow up errors due to an erroneous error region prediction make patching with an error estimator disadvantageous. \textit{NN-Patching}$_{\textit{incl,patchEE}}$ shows most top~1 performances for average accuracy, recovery speed, and final rank. However, \textit{NN-Patching}$_{\textit{incl,noEE}}$ still excels in final accuracy and adaptation rank. Furthermore, the semi-exclusive patching variants occasionally show top~1 performances. \textit{NN-Patching}$_{\textit{semi,baseEE}}$ is tied on the top~1 performances in recovery speed.

Training the full base classifier results in a lower performance in comparison to other competitors for the CNN than for the FC-NN base classifier. This is, because of the slow adaptation capabilities and the fact that the other models improve due to the availability of better features in the engagement layer. However, \textit{Freezing} achieves better performance. For the CNN base classifiers, the engagement layer is the last convolutional or pooling layer of the network. This leaves the last two layers of the network trainable for \textit{Freezing}. The trainable part of the network is comparable to our patch architecture. Therefore, \textit{Freezing} often accomplishes fast adaptation, although some NN-Patching variants perform even better.   

\subsubsection{Residual Base Architecture}
\label{resnet1}

The meta-table for the ResNet architectures is presented in Table~\ref{Number of Top 1 model performances on $NIST$ and $MNIST$ for ResNet base classifiers}.

\begin{table}[h]
\begin{centering}
  \begin{tabular}{ |l | c | c | c | c | c|}
    \hline
			\multicolumn{1}{|l|}{Base Archetype: ResNet}&\multicolumn{5}{ c|}{Number of Top~1 Performances}\\ \hline
		Model & A.Acc & F.Acc & R.Spd & Ad.Rk & F.Rk\\ \hline
	\textit{NN-Patching}$_{\textit{incl,noEE}}$ & \textbf{6} & 1 & 3 & 3 & 1 \\   
\textit{NN-Patching}$_{\textit{incl,baseEE}}$ & 1 & -- & -- & -- & 1 \\  
\textit{NN-Patching}$_{\textit{semi,baseEE}}$ & -- & -- & -- & -- & -- \\
\textit{NN-Patching}$_{\textit{excl,baseEE}}$ & -- & -- & -- & 1 & -- \\
\textit{NN-Patching}$_{\textit{incl,patchEE}}$& -- & 2 & \textbf{4} & 1 & 2 \\
\textit{NN-Patching}$_{\textit{semi,patchEE}}$& -- & 1 & -- & 1 & -- \\
\textit{NN-Patching}$_{\textit{excl,patchEE}}$& -- & -- & -- & -- & -- \\
\textit{Freezing}                    & -- & -- & -- & -- & -- \\
\textit{Base}$_{\textit{update}}$            & 3 & \textbf{6} & 1 & \textbf{4} & \textbf{6} \\
    \hline  
  \end{tabular}
\caption{Number of top~1 model performances on \textit{NIST} and \textit{MNIST} for ResNet base classifiers.}
\label{Number of Top 1 model performances on $NIST$ and $MNIST$ for ResNet base classifiers}
\end{centering}
\end{table}

\begin{figure}[H]
\centering
\includegraphics[scale=0.45]{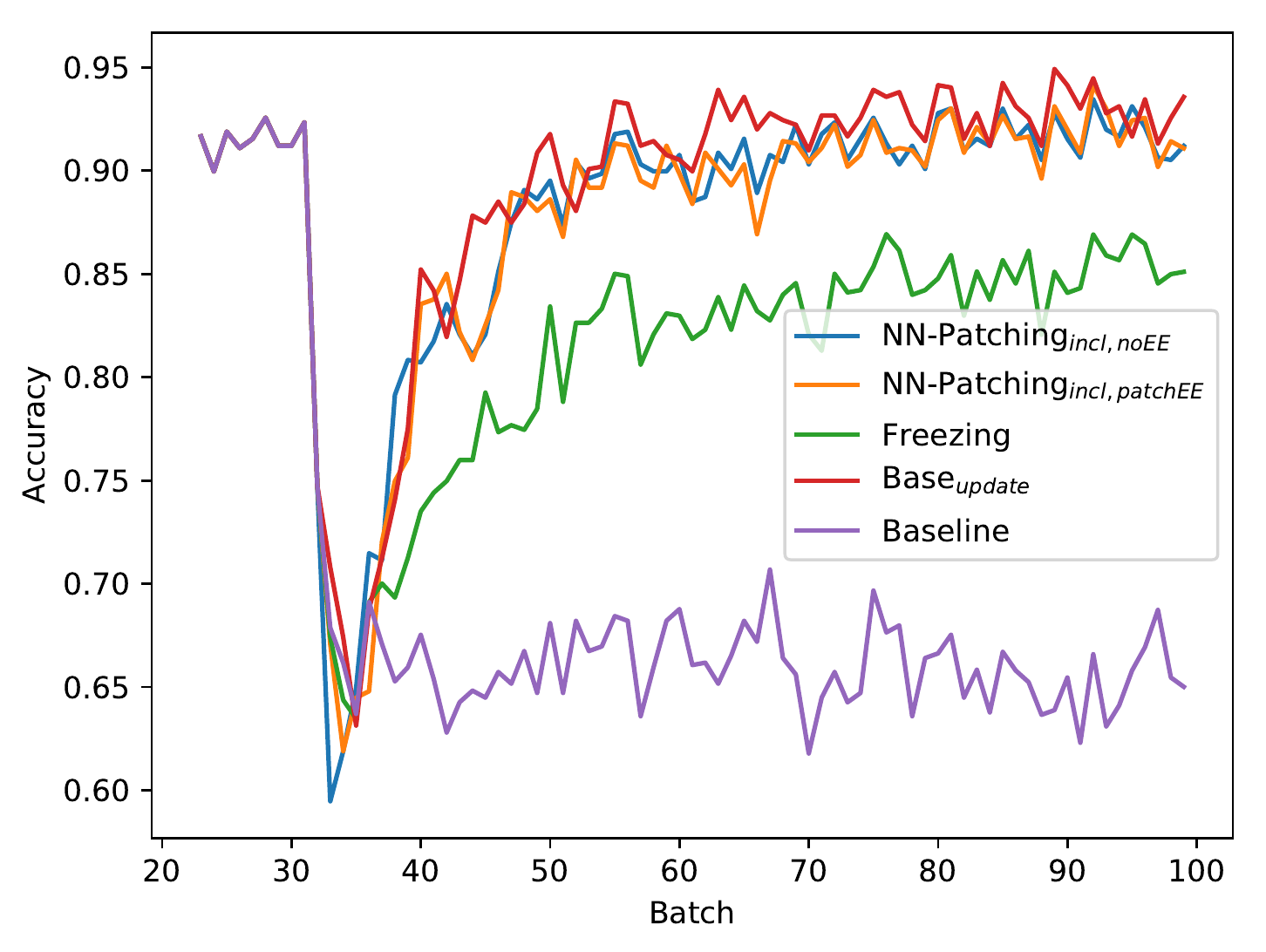}
\caption{\textbf{Comparison of different models on $\mathbf{\textit{NIST}_{\textit{appear}}}$ with a CNN base classifier.} \textit{Base}$_{\textit{update}}$ achieves a higher accuracy than the NN-Patching variants. \textit{Freezing} results in a low performance due to the linear classifier limitations.} 
\label{resnet_base_up}
\end{figure}

For the ResNet, it is often beneficial to train the full base classifier. \textit{Base}$_{\textit{update}}$has most top~1 performances in final accuracy, adaptation rank, and final rank. Surprisingly, re-training the full ResNet often results in fast adaptation. Especially on the $\mathbf{\textit{NIST}_{\textit{appear}}}$ dataset (Tab.~\ref{Comparison of neural network patching and transfer learning techniques on $NIST$ with ResNet base classifiers}), where transfer weights are preferable over random initialization, \textit{Base}$_{\textit{update}}$achieves a better performance than every other competitor for all evaluation measures (Fig.~\ref{resnet_base_up}). The fast learning capabilities of the ResNet are not observed during the initial learning phase. In this initial training process, we observe that the FC-NN and CNN require less epochs to saturate the accuracy. In contrast, the pre-trained ResNet adapts fast. However, updating the whole ResNet is costly in terms of computational resources.

The most top~1 performances for average accuracy are achieved by \\ 
\textit{NN-Patching}$_{\textit{incl,noEE}}$. Moreover, the inclusive patch with patch error region estimator (\textit{NN-Patching}$_{\textit{incl,patchEE}}$) shows most top~1 performances for recovery speed. This again indicates the benefit of using the patch/base classifier ensemble especially on the first batches after the occurrence of the concept drift in order to leverage recovery speed.  

\subsection{Modelling the Patch Error Region versus the Base Classifier Error Region}
\label{Modelling the Patch Error Region versus the Base Classifier Error Region}
In Section~\ref{sec:modellingerrorregions} we elaborated on the theoretical benefits that come from modelling the patch error region instead of the base error region. In this section, we investigate, which variant is superior across all datasets and base classifiers. Therefore, we sum up the top~1 performances from neural patching variants that use a base error estimator (i.e. \textit{NN-Patching}$_{\textit{incl,baseEE}}$, \textit{NN-Patching}$_{\textit{semi,baseEE}}$, and \textit{NN-Patching}$_{\textit{excl,baseEE}}$) or a patch error estimator (i.e \textit{NN-Patching}$_{\textit{incl,patchEE}}$, \textit{NN-Patching}$_{\textit{semi,patchEE}}$, and \textit{NN-Patching}$_{\textit{excl,patchEE}}$) respectively. The cumulative top~1 performances for patch error estimator and base error estimator are presented in Table~\ref{Comparison of Top 1 model performances on for NN-Patching variants with base error estimator and patch error estimator}. 

\begin{table}[h]
\caption{\textbf{Comparison of top~1 model performances for NN-Patching variants with base error estimator and patch error estimator.} We further summarize the top~1 performance tables by respectively adding up the performances of the patching variants with base error region estimator and patch error estimator. This table includes all base archetypes.}
\label{Comparison of Top 1 model performances on for NN-Patching variants with base error estimator and patch error estimator}

\begin{centering}
  \begin{tabular}{ |l | c | c | c | c | c|}
    \hline
			\multicolumn{6}{ |c|}{Number of Top~1 Performances}\\ \hline
		Models with & A.Acc & F.Acc & R.Spd & Ad.Rk & F.Rk\\ \hline
Base Error Estimator & 4 & 3 & 6 & 3 & 1 \\  
Patch Error Estimator& \textbf{7} &\textbf{7} & \textbf{11} & \textbf{6} & \textbf{6} \\
    \hline  
  \end{tabular}

\end{centering}
\end{table}

The table indicates that in most cases it is preferable to use an patch error estimator instead of the base error estimator. The variants with patch error estimator accomplish more top~1 performances for all evaluation measures compared to the base error estimator variants.  

However, on specific datasets modelling the error region of the base classifier is still beneficial. The transfer-datasets ($\mathbf{\textit{MNIST}_{\textit{transfer}}}$ and $\mathbf{\textit{NIST}_{\textit{transfer}}}$) are special, because the base classifier accuracy is 0\,\% after the drift. In other words, the base classifier prediction on instances after the concept drift is always incorrect. The task for the error estimator is to direct all instances to the patch network for classification. The error region of the base classifier instantly comprises all instances from the data batches after the concept drift. The optimal solution to forward all instances to the patch is easily found by the base error estimator. In contrast, the capabilities of the patch network grow with training on the batches after the drift. The patch does not classify all instances correctly after the occurrence of a concept drift. Hence, on the first batches after the drift the patch error estimator tries to represent the capabilities of the patch, which does not take into account that the base network is not capable of correctly classifying a single instance (Fig.~\ref{fig:Comparison of patching with base error estimator, patch error estimator and stand-alone patch without error estimator on $NIST_{transfer}$}). Moreover, the direction of an instance to the base classifier always results in an incorrect classification, thus the allocation of instances to the base classifier is fully penalized. If the patch network adapts slowly to the new concept, the performance loss even manifests in a final accuracy decrease. Strong performance decreases of patching variants with patch error estimator on the transfer-datasets, due to the described phenomenon, can be observed in Table~\ref{Comparison of neural network patching and transfer learning techniques on $NIST$ with FC-NN base classifiers},~\ref{Comparison of neural network patching and transfer learning techniques on $MNIST$ with ResNet base classifiers}, and~\ref{Comparison of neural network patching and transfer learning techniques on $NIST$ with ResNet base classifiers}

Further, we note that \textit{NN-Patching}$_{\textit{incl,noEE}}$ is the optimal model to deal with the transfer-datasets, since all instances are directed to the patch network by design.

\begin{figure}[H]
\centering
\includegraphics[scale=0.5]{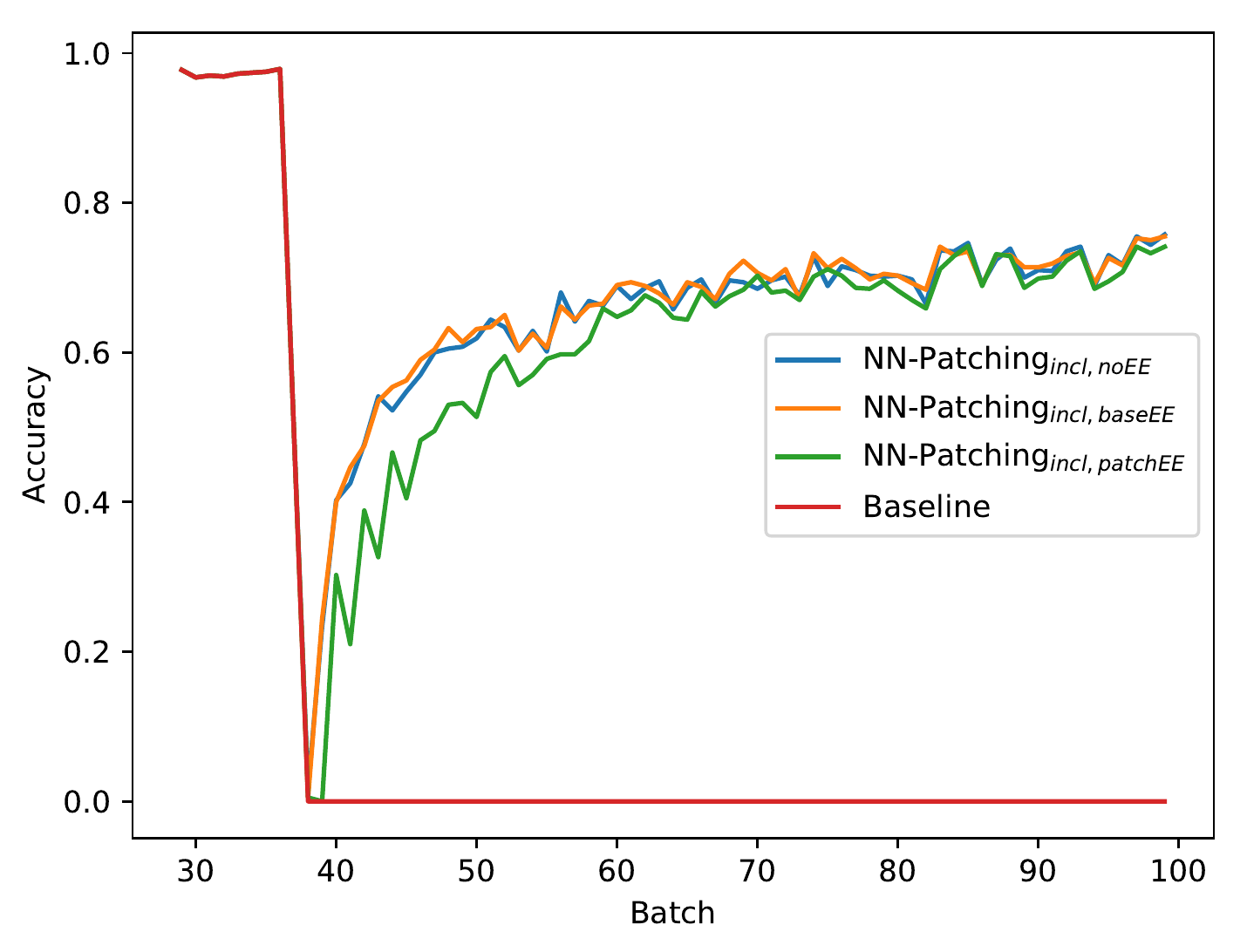}
\caption{\textbf{Comparison of patching with base error estimator, patch error estimator, and stand-alone patch without error estimator on $\mathbf{\textit{NIST}_{\textit{transfer}}}$.} \textit{NN-Patching}$_{\textit{incl,noEE}}$ and \textit{NN-Patching}$_{\textit{incl,baseEE}}$ show comparable performance, hence the base error estimator quickly learns to refer all instances to the patch for classification. \textit{NN-Patching}$_{\textit{incl,patchEE}}$ achieves lower performance than the other two competitors, due to the modelling of the patch error region. The concept drift occurred at batch 35. The base classifier is a FC-NN.} 
\label{fig:Comparison of patching with base error estimator, patch error estimator and stand-alone patch without error estimator on $NIST_{transfer}$}
\end{figure}

Another example, where the patch error estimator shows deficiency in comparison to the base error estimator, are the rotate-datasets. In these datasets every new batch represents a new concept (i.e. different degree of rotation). The patch network tries to adapt to the new concept. Due to the adaptation to a new concept for every arriving batch, the error region of the patch network changes rapidly. Therefore, the concept underlying the training data for the patch error estimator is also quickly changing. This results in a poor error estimation performance for patch error estimators, whereas the stationary concept of the base classifier error region is easier to learn. This phenomenon can, just as the performance decrease on transfer-datasets, also be observed in Table~\ref{Comparison of neural network patching and transfer learning techniques on $NIST$ with FC-NN base classifiers},~\ref{Comparison of neural network patching and transfer learning techniques on $MNIST$ with ResNet base classifiers}, and~\ref{Comparison of neural network patching and transfer learning techniques on $NIST$ with ResNet base classifiers}.

In order to conclude this section, we recognize that these effects scale with patch performance. The better the patch performance, the better the patch error estimation. The magnitude of the described effects are dependent on the quality and transferability of features in the engagement layer and the difficulty of the dataset. Hence, these phenomenons can be observed particularly well in the full result tables based on weaker base classifiers. 

\subsection{Training Schemas for Patching with Error Estimator}
\label{On inclusive, exclusive and semi-exclusive Patch Training with Error Estimator}
Here, we evaluate, which patch training scheme accomplishes the best results across all datasets and base classifiers. Analogous to the table in the previous section about patch error region estimator and base error region estimator, we sum up the cumulative top~1 performances by inclusive, exclusive, and semi-exclusive patch training variants respectively (Tab.~\ref{Comparison of Top 1 model performances on for NN-Patching variants with inclusive, exclusive and semi-exclusive patch training with error estimation}).  

\begin{table}[h]
\begin{centering}
  \begin{tabular}{ |l | c | c | c | c | c |}
    \hline
			\multicolumn{6}{ |c|}{Number of Top~1 Performances}\\ \hline
		Models with & A.Acc & F.Acc & R.Spd & Ad.Rk & F.Rk\\ \hline
Inclusive Training & \textbf{8} & 4 & \textbf{10} & \textbf{4} & \textbf{7} \\  
Semi-Exclusive Training& 3 & \textbf{6} & 5 & \textbf{4} & -- \\
Exclusive Training& -- & -- & 2 & 1 & -- \\
    \hline  
  \end{tabular}
\caption{\textbf{Comparison of top~1 model performances for NN-Patching variants with inclusive, exclusive, and semi-exclusive patch training.} We obtain this table by summing up the top~1 performances of NN-Patching models by inclusive, exclusive, and semi-exclusive training across all base archetypes. \textit{NN-Patching}$_{\textit{incl,noEE}}$ is excluded, since no error estimator is used.}
\label{Comparison of Top 1 model performances on for NN-Patching variants with inclusive, exclusive and semi-exclusive patch training with error estimation}
\end{centering}
\end{table}

Inclusive training accomplishes most top~1 performances in average accuracy, recovery speed, and final rank. Adaptation rank is tied with semi-exclusive training and in terms final accuracy semi-inclusive training outperforms the inclusive training scheme. 

Exclusive training shows low performance for all evaluation measures. Thus, exclusive training is not a suitable training scheme for neural network patching. The reason for this is the lack of robustness towards a poor error estimator as thoroughly discussed in Section~\ref{sec:semiexclusivetraining} and~\ref{Effect of different Base Network Capabilities after Drift on inclusive, exclusive and semi-exclusive Patching Performance}. Semi-exclusive training successfully counteracts this problem. 

In most cases, inclusive training is the best choice, due to the highest robustness towards a poor error estimator. The theoretical performance advantage of semi-exclusive training, due to the fact that the patch can focus on modelling merely a sub-problem of the instance space as elaborated in Section~\ref{Theoretical Advantage of exclusive over inclusive Patch Network Training}, can not be concluded from the data. Semi-inclusive training shows best performance in final accuracy, which indicates a performance increase due to the sub-problem advantage. However, inclusive training dominates in terms of final rank, which contradicts the existence of such a gain in performance. 

\subsection{Performance Differences due to Transfer Weights and Random Initialization}
\label{Performance differences due to Transfer weights and Random Initialization}
In Section~\ref{On the Performance Difference between Transfer Weights and Random Initialisation}, we investigated the difference of \textit{Freezing} with random initialization and transfer weights and discussed the findings with respect to improving neural network patching. We concluded that only if the base network classification accuracy after the occurrence of the drift is greater than approximately 50\,\%, the transfer weights are preferable over the random initialization (i.e. Glorot Uniform Initialization). If this is not the case, using the transfer weights often results in significant performance decreases in comparison to the random initialization. From all used datasets in the experiments only $\mathbf{\textit{MNIST}_{\textit{appear}}}$, $\mathbf{\textit{NIST}_{\textit{appear}}}$, and $\mathbf{\textit{MNIST}_{\textit{rotate}}}$ satisfy this requirement (for $\mathbf{\textit{MNIST}_{\textit{rotate}}}$ this is only true for CNN and ResNet base classifiers, since convolutional features are more transferable).     

\textit{Freezing} uses transfer weights, whereas the neural network patching variants use random initialization. Therefore, a part of the performance differences in the results can be explained by the difference in the initialization. Especially on flip-, remap- or transfer-datasets the performance of the transfer learning methods (i.e. \textit{Freezing}, Base$_{update}$) suffer from the transfer weight initialization. 

\begin{figure}[H]
	\centering
		\subfigure[\label{fig:thesis_zz_last_cnn_nist_appear_0} $\mathbf{\textit{NIST}_{\textit{appear}}}$]{\includegraphics[scale=0.59]{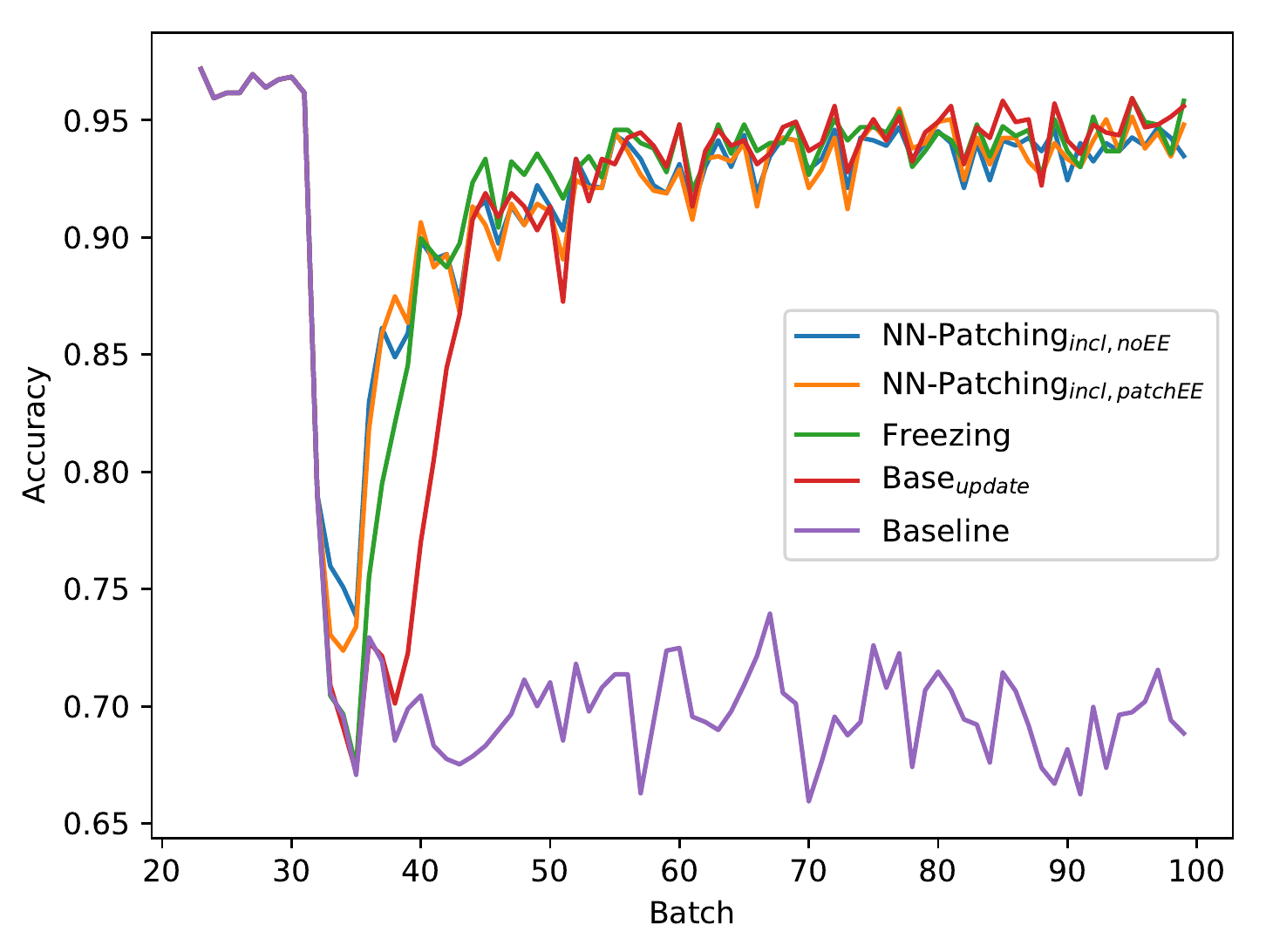}}
		\subfigure[\label{fig:thesis_zz_last_cnn_nist_flip_0} $\mathbf{\textit{NIST}_{\textit{flip}}}$]{\includegraphics[scale=0.59]{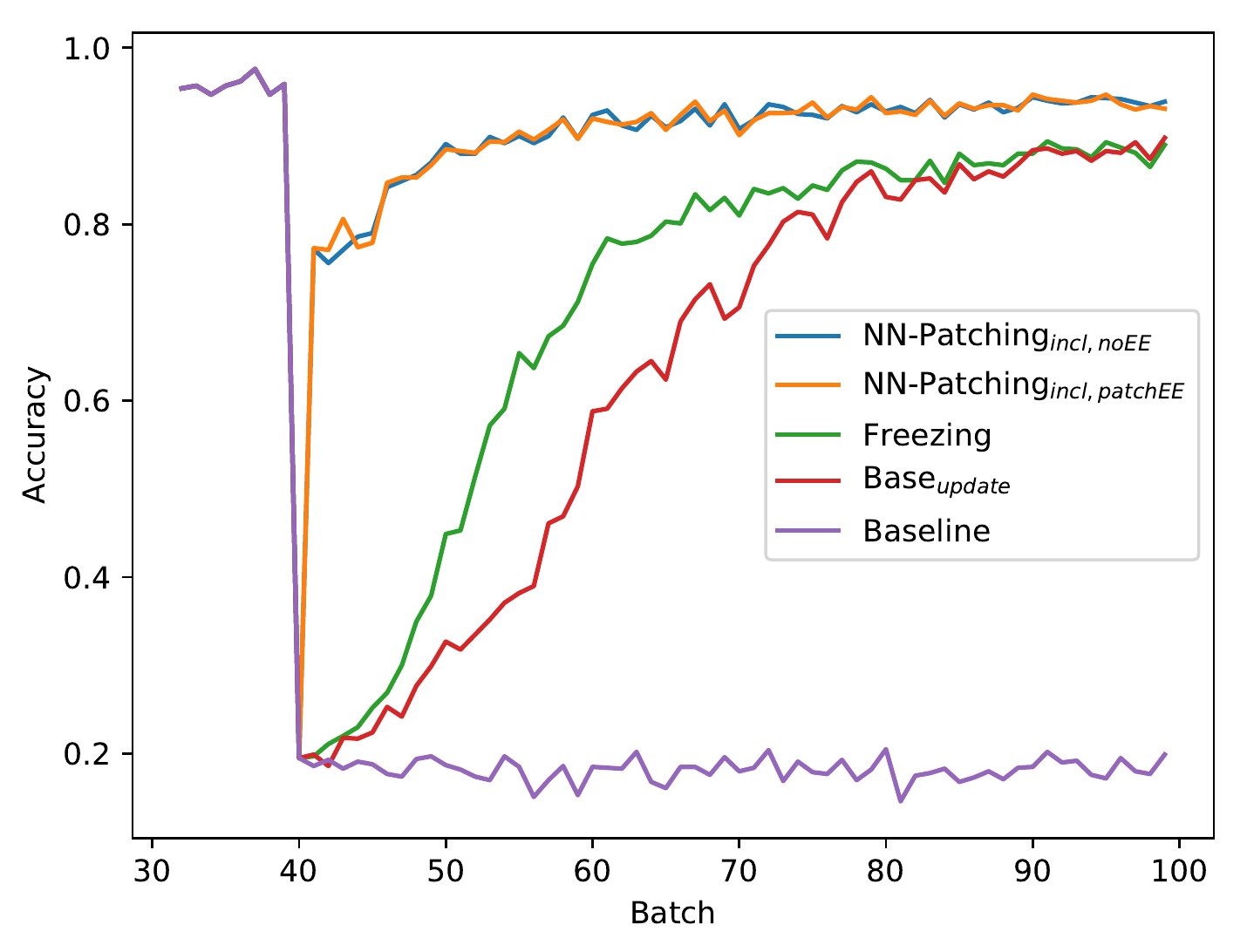}}
\caption{\textbf{Comparison of NN-Patching variants and transfer learning methods on $\mathbf{\textit{NIST}_{\textit{appear}}}$ and $\mathbf{\textit{NIST}_{\textit{flip}}}$ for CNN base classifiers.} In Figure~(a) \textit{NN-Patching}$_{\textit{incl,noEE}}$, \textit{NN-Patching}$_{\textit{incl,patchEE}}$ and \textit{Freezing} show a comparable performance. \textit{Base}$_{\textit{update}}$converges slower. In Figure~(b) the transfer learning models show significantly lower performance than the NN-Patching models.}
\label{Comparison of NN-Patching variants and transfer learning methods on $NIST_{appear}$ and $NIST_{flip}$ for CNN base classifiers}
\end{figure}

In Figure~\ref{fig:thesis_zz_last_cnn_nist_appear_0} we show an accuracy progression on $\mathbf{\textit{NIST}_{\textit{appear}}}$, where the transfer weight initialization is preferable. Thus, \textit{Freezing} shows a comparable performance as \textit{NN-Patching}$_{\textit{incl,noEE}}$ and \textit{NN-Patching}$_{\textit{incl,patchEE}}$.

On $\mathbf{\textit{NIST}_{\textit{flip}}}$ the transfer weight initialization is disadvantageous. Therefore, in Figure~\ref{fig:thesis_zz_last_cnn_nist_flip_0} we observe a highly degraded performance of the transfer learning methods in comparison to the NN-Patching the models. 

\subsection{Catastrophic Forgetting}
\label{Catastrophic Forgetting}
Neural networks have a tendency to forget previously learned information upon learning new information~\cite{Ratcliff1990ConnectionistMO}. This phenomenon is referred to as \textit{catastrophic forgetting}. In concept drift learning, \textit{catastrophic forgetting} is relevant, when a previous concept is reoccurring after some time.

Transfer learning models lose the ability to classify instances well from the original concept after adapting to different concepts in between. In contrast, neural network patching, in theory, does not suffer from this problem, since the base classifier is preserved. A simple approach to gain fast performance recovery on the reoccurrence of the original concept is to evaluate the base network on every new batch, if a concept drift is detected. Hence, on the detection of a new concept in the data stream, we check whether the new concept is identical to the original concept. If the original concept (or a closely related concept) is identified, all instances are directed to the base network for classification. 

However, this approach needs one batch to detect the original concept. Therefore, on the batch in which the concept drift occurs, the base classifier might not be fully utilized and directing instances to the patch may result in a performance decrease.  

A better approach would be to use the error estimator to detect instances in which the base classifier is confident. This would result in no recovery time at all, since the base classifier is instantly utilized. 

However, the base error estimator network also suffers from \textit{catastrophic forgetting}. Thus, after further training, the error estimator loses the ability to correctly predict the error region for instances from the original concept. This happens, although, the concept underlying the training data for the error estimator is stationary. Merely the attribute distribution Pr$\lbrack$attributes$\rbrack$ or class distribution Pr$\lbrack$class$\rbrack$ changes.

In order to tackle this problem, we introduce \textit{memory rehearsal}. \textit{Rehearsal} is motivated from the cognitive psychology and describes the process of repeating information without thinking about its meaning or connecting it to other information. \textit{Rehearsal} is used in the human brain to maintain information in the short-term memory. An intuition of \textit{maintenance rehearsal} would be to repeat a phone number mentally.~\cite{trove.nla.gov.au/work/7564959}

\begin{figure}[H]
\centering
\includegraphics[scale=0.50]{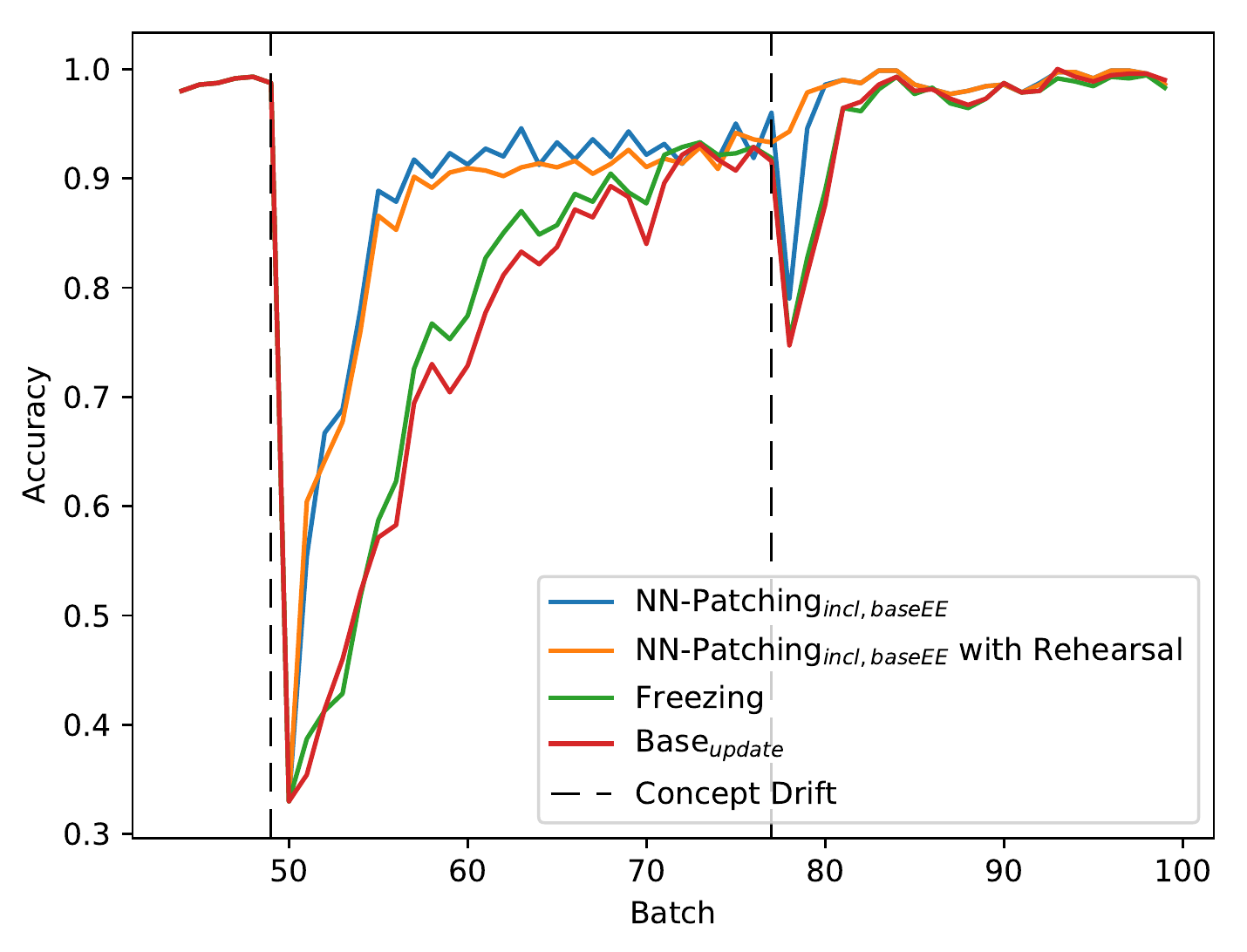}
\caption{\textbf{NN-Patching variants and transfer methods handling a reoccurring concept.} After the first concept drift the instances are flipped vertically and horizontally. After the second concept drift the data returns to the original concept ($\mathbf{\textit{MNIST}_{\textit{flip,reoccur}}}$). \textit{NN-Patching}$_{\textit{incl,baseEE}}$, \textit{Freezing}, and \textit{Base}$_{\textit{update}}$suffer equally from catastrophic forgetting, whereas \textit{NN-Patching}$_{\textit{incl,baseEE}}$ with rehearsal instantly utilizes the base classifier and overcomes the problem of catastrophic forgetting.} 
\label{fig:thesis_catastrophic_forgetting_cnn_mnist_flip_reoccur}
\end{figure}

\textit{Rehearsal} in terms of neural networks is to additionally retrain on data from the concept, which we want to maintain, in every training cycle~\cite{Ratcliff1990ConnectionistMO}\cite{articleccccc}. We implemented this by storing batches from the original concept and training the error estimator on the instances from the arriving batch mixed with instances from a batch, which is following the original concept (the instances are mixed 1:1).   

In order to evaluate the capabilities of the models dealing with reoccurring concepts, we alter the $\mathbf{\textit{MNIST}_{\textit{flip}}}$ dataset. The derived dataset, containing a reoccurring concept, comprises 70k instances and is named $\mathbf{\textit{MNIST}_{\textit{flip,reoccur}}}$. The dataset is divided into 100 batches. The first 35k instances are unaltered. From instance 35k-55k the data is flipped vertically and horizontally. Instances 55k-70k are again from the original \textit{MNIST} dataset. Therefore, the dataset has two concept drifts. At the first change point, the instances are flipped vertically and horizontally. At the second change point, the data returns to the original concept.

In Figure~\ref{fig:thesis_catastrophic_forgetting_cnn_mnist_flip_reoccur} we show that \textit{NN-Patching}$_{\textit{incl,baseEE}}$, \textit{Freezing}, and \textit{Base}$_{\textit{update}}$suffer equally from \textit{catastrophic forgetting}. In contrast, \textit{NN-Patching}$_{\textit{incl,baseEE}}$ with \textit{rehearsal} overcomes this problem. However, the addition of \textit{rehearsal} slightly decreases the accuracy of the model, when adapting after the first change point, since there is a trade-off between representing the original concept and the new concept.

\clearpage
\subsection{Conclusion on Neural Network Patching and Transfer Learning}
\label{Summarization}
In Section~\ref{Result Overview}, we discuss the results for each each classifier archetype individually. We conclude that the usage of the base classifier/patch ensemble with error estimator is only beneficial, if the quality of features in the engagement layer is high. The ensembles particularly perform well in recovery speed, since the usage of the base classifier leverages the model performance especially on the recent batches after the drift. The CNN generates the most transferable features.

Updating the whole base classifier often results in a better final accuracy and final rank than the other models due to the large model complexity, but lacks in recovery speed. However, for a ResNet base classifier it is especially beneficial to train the whole base network, since this often leads to the best performance for all evaluation measures (even in recovery speed). 

\textit{Freezing} shows a similar performance in comparison to the NN-Patching variants, if the trainable network part is comparable to the optimal patch architecture. Besides that, the performance of \textit{Freezing} is often decreased in comparison to the other models. 

In Section~\ref{Modelling the Patch Error Region versus the Base Classifier Error Region} we conclude that, in most cases, it is beneficial to have an patch error estimator instead of a base error estimator. However, if the concept underlying the data rapidly changes or the base classifier accuracy after the drift is very low, it is still preferable to model the error region of the base classifier.

Furthermore, we elaborated on the performance differences between inclusive, exclusive, and semi-exclusive patch training (Sec.~\ref{On inclusive, exclusive and semi-exclusive Patch Training with Error Estimator}). The inclusive patch training results in the best overall performance due to the robustness towards a poor error estimator. Exclusive training is not beneficial in any observed scenario. However, the semi-exclusive training scheme improves the robustness of exclusive training and often leads to a comparable performance as the inclusive training. 

In Section~\ref{Performance differences due to Transfer weights and Random Initialization} we elaborate that the performance difference between NN-Patching and transfer learning methods can partly be explained with the difference in the initialization scheme. In most cases, random initialization is preferable over transfer weights.

Finally, we discussed the capability of NN-Patching and transfer learning methods in order to deal with reoccurring concepts (Sec.~\ref{Catastrophic Forgetting}). We observe that the error estimator and transfer methods equally suffer from catastrophic forgetting. However, we managed to counteract the catastrophic forgetting of the error estimator with the use of rehearsal (i.e. partly retraining on the original concept). Therefore, the error estimator is capable of detecting the reoccurring concept and instantly directing instances, which are following the reoccurring concept, to the base network for classification.

\clearpage

\section{Conclusion}
\label{sec:conclusion}
In this work, we applied \textit{Patching} to neural networks. We conducted a variety of different experiments in order to leverage the performance and improve the understanding of \textit{neural network patching}. We discovered that \textit{neural network patching} can profit from utilizing information from the inner layers of a given base network. However, the method introduces new hyperparameters such as patch architecture and engagement layer selection. In Section~\ref{sec:networkarchitectureandengagement} we tackle these problems and obtain heuristic rules for engagement layer selection and guidelines for the selection of the patch architecture. These findings show how to select neural network architectures in order to achieve fast adaptation and give insights about the functionality and information processing in inner network layers. 

Furthermore, we evaluated different training schemes for patch training. The training schemes are: inclusive, exclusive, and semi-exclusive patch training. In theory, the exclusive and semi-exclusive training scheme could benefit from the division of the instance space into sub-problems. The benefit that comes from dividing the instance space of a learning tasks into sub-problems is empirically shown in Section~\ref{Theoretical Advantage of exclusive over inclusive Patch Network Training}. However, our results indicate that patching with error estimator in practice is a difficult task. Often the stand-alone patch network results in the best performance. 

The use of an classifier ensemble, consisting of the patch and base classifier, is only beneficial if the engagement layer contains sufficiently transferable features. Inclusive training is the best choice in most cases, because of the robustness towards fault-prone error region estimations. Although semi-exclusive training successfully leverages the robustness towards fault-prone error region estimations in comparison to exclusive training, the semi-exclusive ensemble is still usually outperformed by the inclusive ensemble.

Moreover, we explored the option of modelling the error region of the patch instead of the base classifier error region. In most cases, the model with patch error estimator leads to an improvement in comparison to the base error estimator. However, in special cases, such as the concept rapidly changing throughout the data stream or the impracticality of the base network predictions after the occurrence of the drift, it is still beneficial to use a base error estimator.

Section~\ref{On the Performance Difference between Transfer Weights and Random Initialisation} looks at the performance difference that comes from transfer weights versus random initialization. We conclude that random initialization is the preferable choice in most cases. Only if the drift task is closely related to the target task, then the transfer weights are preferable. Our results indicate that this is the case if the base classifier accuracy after the drift is greater than 50\,\%.

In Section~\ref{Final Results} we compare \textit{neural network patching} to transfer learning models. The results show that the \textit{neural network patching} variants outperform transfer learning methods in the majority of cases. The edge is particularly significant in recovery speed and adaptation rank. The results further indicate that ResNets are naturally capable of fast adaptation to new concepts.


\subsection{Future Work}
The content of this thesis offers considerable opportunities for further research. The main drawback of using the classifier ensemble (i.e. base classifier and patch) is the insufficient performance of the error estimator network. If future research leverages the performance of the error estimator, then the benefit from exclusive or semi-exclusive patch training would be significantly increased. 

Moreover, the search space for suitable patch architectures is immense. This work merely covers a small subset of possible patch architectures. Architectures with convolutional layers, batch normalization, or residual connections could result in a performance increase. In addition, we only evaluated our models on altered \textit{NIST} and \textit{MNIST}. A more comprehensive evaluation on a larger variety of datasets could substantiate our observations and lead to additional knowledge about \textit{neural network patching}.

At last, we want to mention that the fast adaptation capabilities of ResNets are promising in order to deal with non-stationary environments. A deeper analysis of this phenomenon could lead to relevant methods for neural networks in relation to concept drift learning.

\clearpage

\bibliography{./references,./ensembles,./kauschke,./jf,./referencesdavid}

\begin{thebibliography}{10}

\bibitem{BifetMOA}
Albert Bifet, Geoff Holmes, Richard Kirkby, and Bernhard Pfahringer.
\newblock {MOA Massive Online Analysis}.
\newblock {\em Journal of Machine Learning Research}, 11:1601--1604, 2010.

\bibitem{chollet2015keras}
Fran\c{c}ois Chollet et~al.
\newblock Keras.
\newblock \url{https://keras.io}, 2015.

\bibitem{DBLP:journals/corr/ChoromanskaHMAL14}
Anna Choromanska, Mikael Henaff, Micha{\"{e}}l Mathieu, G{\'{e}}rard~Ben Arous,
  and Yann LeCun.
\newblock The loss surface of multilayer networks.
\newblock {\em CoRR}, abs/1412.0233, 2014.

\bibitem{Ciresan2012}
Dan~C. Cire{\c{s}}an, Ueli Meier, and Jurgen Schmidhuber.
\newblock {Transfer Learning for Latin and Chinese Characters with Deep Neural
  Networks}.
\newblock In {\em Proceedings of the 2012 International Joint Conference on
  Neural Networks -- IJCNN'12}, pages 1--6, 2012.

\bibitem{French1999}
R~French.
\newblock {Catastrophic Forgetting in Connectionist Networks}.
\newblock {\em Trends in Cognitive Sciences}, 3(4):128--135, 1999.

\bibitem{trove.nla.gov.au/work/7564959}
E.~Bruce Goldstein, Daniel Vanhorn, Greg Francis, and 1965 Neath, Ian.
\newblock {\em Cognitive psychology : connecting mind, research, and everyday
  experience}.
\newblock Belmont, CA : Wadsworth/Cengage Learning, third edition edition,
  2011.
\newblock Previous ed.: 2008.

\bibitem{He2016}
Kaiming He, Xiangyu Zhang, Shaoqing Ren, and Jian Sun.
\newblock Deep residual learning for image recognition.
\newblock In {\em Proceedings of the 2016 IEEE Conference on Computer Vision
  and Pattern Recognition -- CVPR'16}, June 2016.

\bibitem{DBLP:journals/corr/IoffeS15}
Sergey Ioffe and Christian Szegedy.
\newblock Batch normalization: Accelerating deep network training by reducing
  internal covariate shift.
\newblock {\em CoRR}, abs/1502.03167, 2015.

\bibitem{Kauschke2018}
Sebastian Kauschke and Johannes F{\"{u}}rnkranz.
\newblock {Batchwise Patching of Classifiers}.
\newblock In {\em Proceedings of the 32nd {AAAI} Conference on Artificial
  Intelligence -- {AAAI'18}}, 2018.

\bibitem{Oquab2014}
Maxime Oquab, Leon Bottou, Ivan Laptev, and Josef Sivic.
\newblock {Learning and Transferring Mid-Level Image Representations using
  Convolutional Neural Networks}.
\newblock In {\em IEEE Conference on Computer Vision and Pattern Recognition --
  CVPR'14}, pages 1717--1724, 2014.

\bibitem{Ratcliff1990ConnectionistMO}
Roger Ratcliff.
\newblock Connectionist models of recognition memory: constraints imposed by
  learning and forgetting functions.
\newblock {\em Psychological review}, 97 2:285--308, 1990.

\bibitem{articleccccc}
Anthony Robins.
\newblock Catastrophic forgetting, rehearsal and pseudorehearsal.
\newblock {\em Connection Science}, 7:123--, 01 1995.

\bibitem{JMLR:v15:srivastava14a}
Nitish Srivastava, Geoffrey Hinton, Alex Krizhevsky, Ilya Sutskever, and Ruslan
  Salakhutdinov.
\newblock Dropout: A simple way to prevent neural networks from overfitting.
\newblock {\em Journal of Machine Learning Research}, 15:1929--1958, 2014.

\bibitem{Yosinski2014}
Jason Yosinski, Jeff Clune, Yoshua Bengio, and Hod Lipson.
\newblock {How Transferable are Features in Deep Neural Networks?}
\newblock In {\em Advances in Neural Information Processing Systems 27}, pages
  3320--3328. Curran Associates, Inc., 2014.

\bibitem{DBLP:journals/corr/ZeilerF13}
Matthew~D. Zeiler and Rob Fergus.
\newblock Visualizing and understanding convolutional networks.
\newblock {\em CoRR}, abs/1311.2901, 2013.

\end{thebibliography}
\bibliographystyle{plain}


\clearpage
\appendix
\section{Full Result Tables for the Comparison of Neural Network Patching and Transfer Learning Methods}
\label{Appendix A}

\subsection{Result Tables for the Fully-Connected Base Classifiers}

\begin{table}[!htbp]
\begin{centering}
\resizebox{\columnwidth}{!}{%
  \begin{tabular}{ |l | l l l l l | l l l l l |}
    \hline
		\multicolumn{11}{|c|}{Fully-Connected Neural Network on \textit{MNIST}}\\ \hline
		Dataset: & \multicolumn{5}{ c|}{$\mathbf{\textit{MNIST}_{\textit{appear}}}$}& \multicolumn{5}{ c|}{$\mathbf{\textit{MNIST}_{\textit{flip}}}$}\\ \hline
		Model & A.Acc & F.Acc & R.Spd & Ad.Rk & F.Rk & A.Acc & F.Acc & R.Spd & Ad.Rk & F.Rk \\ \hline
Baseline & 50.55 & 51.01 & --- & 9.39 & 10.0 & 29.64 & 29.39 & --- & 10.0 & 10.0 \\ 
\textit{NN-Patching}$_{\textit{incl,noEE}}$    & \textbf{88.05} & 95.0 & 14.4 & \textbf{3.31} & 3.98 & \textbf{90.75} & 93.63 & \textbf{6.5} & \textbf{1.71} & 3.7 \\ 
\textit{NN-Patching}$_{\textit{incl,baseEE}}$  & 87.77 & 94.79 & 16.0 & 3.84 & 5.08 & 87.97 & 90.89 & 14.5 & 4.42 & 7.58 \\
\textit{NN-Patching}$_{\textit{semi,baseEE}}$  & 87.84 & 94.9 & 16.4 & 3.68 & 4.6 & 88.11 & 90.99 & 13.9 & 4.33 & 7.48 \\ 
\textit{NN-Patching}$_{\textit{excl,baseEE}}$  & 87.24 & 93.87 & 15.0 & 3.99 & 7.36 & 86.07 & 90.59 & 19.0 & 7.03 & 8.36 \\ 
\textit{NN-Patching}$_{\textit{incl,patchEE}}$ & 87.87 & 95.12 & \textbf{13.4} & 3.6 & 3.08 & 90.68 & 93.68 & 7.3 & 1.99 & 3.36 \\
\textit{NN-Patching}$_{\textit{semi,patchEE}}$ & 86.2 & 92.81 & 15.7 & 4.54 & 7.18 & 90.22 & 93.49 & 8.0 & 3.06 & 4.12 \\
\textit{NN-Patching}$_{\textit{excl,patchEE}}$ & 86.7 & 94.16 & 16.5 & 5.29 & 6.38 & 87.35 & 91.9 & 14.2 & 6.04 & 6.46 \\ 
\textit{Freezing}                     & 79.88 & 94.58 & 32.0 & 8.45 & 5.04 & 83.91 & 94.23 & 15.9 & 8.64 & 2.28 \\
\textit{Base}$_{\textit{update}}$             & 79.37 & \textbf{95.61} & 28.8 & 8.91 & \textbf{2.3} & 85.16 & \textbf{94.63} & 14.3 & 7.77 & \textbf{1.66} \\
    \hline  
    \hline
		Dataset: & \multicolumn{5}{ c|}{$\mathbf{\textit{MNIST}_{\textit{remap}}}$}& \multicolumn{5}{ c|}{$\mathbf{\textit{MNIST}_{\textit{rotate}}}$}\\ \hline
		Model & A.Acc & F.Acc & R.Spd & Ad.Rk & F.Rk & A.Acc & F.Acc & R.Spd & Ad.Rk & F.Rk \\ \hline
Baseline & 39.67 & 40.91 & --- & 9.99 & 10.0 & 48.12 & 46.11 & --- & 7.61 & 7.94 \\ 
\textit{NN-Patching}$_{\textit{incl,noEE}}$    & 92.03 & \textbf{95.38} & 5.6 & 2.0 & 2.86 & 67.25 & \textbf{71.23} & --- & 6.18 & \textbf{2.96 }\\ 
\textit{NN-Patching}$_{\textit{incl,baseEE}}$  & 86.65 & 89.45 & 10.6 & 5.61 & 7.58 & 66.7 & 68.74 & --- & 4.85 & 4.24 \\
\textit{NN-Patching}$_{\textit{semi,baseEE}}$  & 86.65 & 89.72 & 9.7 & 5.71 & 7.38 & 66.78 & 68.91 & --- & 5.34 & 4.24 \\ 
\textit{NN-Patching}$_{\textit{excl,baseEE}}$  & 84.44 & 88.81 & 29.5 & 7.87 & 8.06 & 64.47 & 62.2 & --- & 5.34 & 7.88  \\ 
\textit{NN-Patching}$_{\textit{incl,patchEE}}$ & \textbf{92.04} & 95.36 & \textbf{4.9} & \textbf{1.79} & 2.84 & 64.78 & 68.84 & --- & 7.18 & 4.52  \\
\textit{NN-Patching}$_{\textit{semi,patchEE}}$ & 91.22 & 95.13 & 6.0 & 3.53 & 3.88 & 62.34 & 62.02 & --- & 7.02 & 7.66  \\
\textit{NN-Patching}$_{\textit{excl,patchEE}}$ & 87.41 & 91.45 & 18.4 & 6.25 & 6.42 & 61.76 & 61.41 & --- & 7.45 & 8.04 \\ 
\textit{Freezing}                     & 88.47 & 94.93 & 13.7 & 6.75 & 3.22 & 69.08 & 70.12 & --- & 2.51 & 3.72  \\
\textit{Base}$_{\textit{update}}$             & 89.68 & 95.43 & 7.0 & 5.49 & \textbf{2.76} & \textbf{71.06} & 70.3 & --- & \textbf{1.52} & 3.8 \\
    \hline 
		\end{tabular}
		}
		\end{centering}
		\resizebox{10.783cm}{!}{%
	  \begin{tabular}{ |l | l l l l l |}
    \hline
		Dataset: & \multicolumn{5}{ c|}{$\mathbf{\textit{MNIST}_{\textit{transfer}}}$} \\ \hline
		Model & A.Acc & F.Acc & R.Spd & Ad.Rk & F.Rk \\ \hline
Baseline & 0.0 & 0.0 & --- & 10.0 & 10.0 \\ 
\textit{NN-Patching}$_{\textit{incl,noEE}}$ & \textbf{91.78} & \textbf{95.39} & \textbf{5.9} & \textbf{3.6} & \textbf{3.22} \\ 
\textit{NN-Patching}$_{\textit{incl,baseEE}}$ & 91.73 & \textbf{95.39} & 6.0 & 3.81 & 4.04 \\
\textit{NN-Patching}$_{\textit{semi,baseEE}}$ & 91.72 & 95.34 & 6.0 & 4.04 & 4.36 \\ 
\textit{NN-Patching}$_{\textit{excl,baseEE}}$ &  91.69 & 95.36 & 6.0 & 4.03 & 3.88 \\ 
\textit{NN-Patching}$_{\textit{incl,patchEE}}$ & 91.63 & 95.26 & 6.0 & 4.61 & 4.84 \\
\textit{NN-Patching}$_{\textit{semi,patchEE}}$ & 91.76 & 95.31 & 6.0 & 3.93 & 4.84 \\
\textit{NN-Patching}$_{\textit{excl,patchEE}}$ & 91.74 & 95.34 & \textbf{5.9} & 3.97 & 4.56 \\ 
\textit{Freezing}                     & 71.22 & 93.25 & 28.2 & 8.34 & 8.44 \\
\textit{Base}$_{\textit{update}}$             & 67.42 & 93.7 & 28.2 & 8.66 & 6.82 \\
    \hline  
  \end{tabular}
	}
\caption{Comparison of neural network patching and transfer learning techniques on \textit{MNIST} with FC-NN base classifiers.}
\label{Comparison of neural network patching and transfer learning techniques on $MNIST$ with FC-NN base classifiers}
\end{table}

\begin{table}[!htbp]
\begin{centering}
\resizebox{\columnwidth}{!}{%
  \begin{tabular}{ |l | l l l l l | l l l l l |}
    \hline
		\multicolumn{11}{|c|}{Fully-Connected Neural Network on \textit{NIST}}\\ \hline
		Dataset: & \multicolumn{5}{ c|}{$\mathbf{\textit{NIST}_{\textit{appear}}}$}& \multicolumn{5}{ c|}{$\mathbf{\textit{NIST}_{\textit{flip}}}$}\\ \hline
		Model & A.Acc & F.Acc & R.Spd & Ad.Rk & F.Rk & A.Acc & F.Acc & R.Spd & Ad.Rk & F.Rk \\ \hline
Baseline & 65.17 & 65.89 & --- & 8.87 & 10.0 & 14.13 & 14.36 & --- & 10.0 & 10.0 \\ 
\textit{NN-Patching}$_{\textit{incl,noEE}}$   & \textbf{82.36} & 86.11 & 10.2 & 3.57 & 3.52 & \textbf{76.25} & \textbf{83.01 }& \textbf{32.1} & \textbf{2.28} & \textbf{2.0} \\ 
\textit{NN-Patching}$_{\textit{incl,baseEE}}$ & 82.13 & 85.63 & \textbf{8.4} & 3.56 & 5.02 & 75.78 & 82.16 & 38.5 & 2.63 & 4.02 \\
\textit{NN-Patching}$_{\textit{semi,baseEE}}$ & 82.1 & 85.49 & 9.5 & \textbf{3.37} & 5.58 & 75.8 & 82.1 & 37.5 & 2.68 & 4.38 \\
\textit{NN-Patching}$_{\textit{excl,baseEE}}$ & 80.86 & 84.99 & 13.5 & 5.84 & 6.4 & 74.4 & 81.16 & 43.6 & 4.91 & 6.22 \\ 
\textit{NN-Patching}$_{\textit{incl,patchEE}}$ & \textbf{82.36} & 86.13 & 11.7 & 3.86 & 3.18 & 74.81 & 82.59 & 35.7 & 4.46 & 3.2 \\
\textit{NN-Patching}$_{\textit{semi,patchEE}}$ & 81.93 & 85.75 & 10.8 & 4.69 & 4.64 & 74.7 & 82.65 & 35.7 & 4.59 & 2.7 \\
\textit{NN-Patching}$_{\textit{excl,patchEE}}$ & 79.04 & 81.37 & 17.4 & 4.85 & 9.0 & 72.41 & 80.92 & 47.8 & 6.53 & 6.66 \\ 
\textit{Freezing}                     & 78.26 & 85.6 & 30.0 & 7.81 & 5.12 & 58.54 & 78.28 & 58.4 & 8.95 & 8.68 \\
\textit{Base}$_{\textit{update}}$             & 78.37 & \textbf{86.91} & 31.0 & 8.58 & \textbf{2.54} & 59.09 & 80.06 & 55.4 & 7.98 & 7.14 \\
    \hline  
    \hline
		Dataset: & \multicolumn{5}{ c|}{$\mathbf{\textit{NIST}_{\textit{remap}}}$}& \multicolumn{5}{ c|}{$\mathbf{\textit{NIST}_{\textit{rotate}}}$}\\ \hline
		Model & A.Acc & F.Acc & R.Spd & Ad.Rk & F.Rk & A.Acc & F.Acc & R.Spd & Ad.Rk & F.Rk \\ \hline
Baseline & 13.44 & 12.52 & --- & 10.0 & 10.0 & 31.78 & 32.18 & --- & 5.09 & 6.56  \\ 
\textit{NN-Patching}$_{\textit{incl,noEE}}$    & \textbf{78.71} & 84.48 & --- & \textbf{2.39} & 3.16 & 37.73 & 40.51 & --- & 6.19 & 4.12 \\ 
\textit{NN-Patching}$_{\textit{incl,baseEE}}$  & 78.02 & 83.69 & --- & 2.95 & 4.86 & 38.49 & 40.37 & --- & 5.14 & 4.18  \\
\textit{NN-Patching}$_{\textit{semi,baseEE}}$  & 78.18 & 83.76 & --- & 3.11 & 4.66 & \textbf{38.56} & 40.33 & --- & 5.5 & 4.54  \\
\textit{NN-Patching}$_{\textit{excl,baseEE}}$  & 76.18 & 82.65 & --- & 5.6 & 6.92 & 34.76 & 33.31 & --- & 6.65 & 6.8 \\ 
\textit{NN-Patching}$_{\textit{incl,patchEE}}$ & 77.65 & 84.57 & --- & 3.97 & 2.84 & 33.54 & 35.76 & --- & 6.52 & 6.76  \\
\textit{NN-Patching}$_{\textit{semi,patchEE}}$ & 77.09 & 83.97 & --- & 3.83 & 4.16 & 32.91 & 35.5 & --- & 7.63 & 7.14  \\
\textit{NN-Patching}$_{\textit{excl,patchEE}}$ & 74.76 & 82.55 & --- & 6.28 & 6.98 & 32.19 & 34.46 & --- & 7.47 & 7.46  \\ 
\textit{Freezing}                     & 64.58 & 79.29 & --- & 8.66 & 8.66 & 36.51 & 37.77 & --- & 2.9 & 4.5 \\
\textit{Base}$_{\textit{update}}$             & 70.32 & \textbf{85.94} & \textbf{39.8} & 8.21 & \textbf{2.76} & 38.45 & \textbf{40.76} & --- & \textbf{1.91} & \textbf{2.94}  \\
    \hline 
		\end{tabular}
		}
		\end{centering}
		\resizebox{10.783cm}{!}{%
	  \begin{tabular}{ |l | l l l l l |}
    \hline
		Dataset: & \multicolumn{5}{ c|}{$\mathbf{\textit{NIST}_{\textit{transfer}}}$} \\ \hline
		Model & A.Acc & F.Acc & R.Spd & Ad.Rk & F.Rk \\ \hline
Baseline & 0.0 & 0.0 & --- & 9.89 & 10.0 \\
\textit{NN-Patching}$_{\textit{incl,noEE}}$    & \textbf{65.87} & \textbf{73.35} & --- & \textbf{1.44} & \textbf{2.36} \\ 
\textit{NN-Patching}$_{\textit{incl,baseEE}}$  & 64.55 & 72.9 & --- & 3.73 & 3.42 \\
\textit{NN-Patching}$_{\textit{semi,baseEE}}$  & 64.65 & 72.85 & --- & 3.41 & 3.48 \\
\textit{NN-Patching}$_{\textit{excl,baseEE}}$  & 64.6 & 72.96 & --- & 3.34 & 3.42 \\ 
\textit{NN-Patching}$_{\textit{incl,patchEE}}$ & 60.16 & 71.8 & --- & 6.5 & 5.32 \\
\textit{NN-Patching}$_{\textit{semi,patchEE}}$ & 61.7 & 71.81 & --- & 4.91 & 5.02 \\
\textit{NN-Patching}$_{\textit{excl,patchEE}}$ & 61.95 & 71.95 & --- & 4.99 & 4.98 \\ 
\textit{Freezing}                     & 36.57 & 55.86 & --- & 8.23 & 8.26 \\
\textit{Base}$_{\textit{update}}$             & 27.74 & 51.24 & --- & 8.55 & 8.74 \\
    \hline  
  \end{tabular}
	}
\caption{Comparison of neural network patching and transfer learning techniques on \textit{NIST} with FC-NN base classifiers.}
\label{Comparison of neural network patching and transfer learning techniques on $NIST$ with FC-NN base classifiers}
\end{table}

\clearpage
\subsection{Result Tables for the Convolutional Base Classifiers}
\begin{table}[!htbp]
\begin{centering}
\resizebox{\columnwidth}{!}{%
  \begin{tabular}{ |l | l l l l l | l l l l l |}
    \hline
		\multicolumn{11}{|c|}{Convolutional Neural Network on \textit{MNIST}}\\ \hline
		Dataset: & \multicolumn{5}{ c|}{$\mathbf{\textit{MNIST}_{\textit{appear}}}$}& \multicolumn{5}{ c|}{$\mathbf{\textit{MNIST}_{\textit{flip}}}$}\\ \hline
		Model & A.Acc & F.Acc & R.Spd & Ad.Rk & F.Rk & A.Acc & F.Acc & R.Spd & Ad.Rk & F.Rk \\ \hline
Baseline & 50.78 & 51.27 & --- & 9.91 & 10.0 & 35.6 & 35.06 & --- & 9.99 & 10.0 \\ 
\textit{NN-Patching}$_{\textit{incl,noEE}}$    & 92.9 & 98.27 & 6.8 & 3.89 & 3.58 & 94.46 & \textbf{97.9} & 5.7 & 2.55 & 1.8 \\ 
\textit{NN-Patching}$_{\textit{incl,baseEE}}$  & 92.07 & 98.25 & 8.1 & 4.89 & 4.1 & 90.34 & 93.37 & 6.1 & 4.49 & 7.18  \\  
\textit{NN-Patching}$_{\textit{semi,baseEE}}$  & 91.91 & \textbf{98.32} & 9.1 & 5.03 & 4.22 & 90.3 & 93.21 & 6.0 & 4.69 & 7.28 \\
\textit{NN-Patching}$_{\textit{excl,baseEE}}$  & 90.1 & 97.83 & 10.4 & 6.26 & 6.38 & 89.66 & 92.91 & 8.2 & 5.33 & 7.9  \\ 
\textit{NN-Patching}$_{\textit{incl,patchEE}}$ & \textbf{93.01} & 98.26 & \textbf{6.4} & \textbf{3.71} & \textbf{3.42}  & \textbf{94.56} & \textbf{97.9} & \textbf{5.6} & \textbf{2.31} & \textbf{1.54} \\
\textit{NN-Patching}$_{\textit{semi,patchEE}}$ & 89.48 & 96.47 & 12.0 & 6.87 & 8.72 & 93.54 & 97.68 & 5.9 & 3.53 & 2.68  \\
\textit{NN-Patching}$_{\textit{excl,patchEE}}$ & 90.14 & 97.81 & 10.7 & 6.21 & 6.42 & 90.89 & 95.43 & 7.3 & 5.41 & 6.16 \\ 
\textit{Freezing}                     & 92.82 & 98.24  & 7.7 & 4.03 & 4.16 & 87.22 & 95.91 & 13.9 & 8.03 & 4.84 \\
\textit{Base}$_{\textit{update}}$             & 92.75 & 98.28 & 8.6 & 4.21 & 4.0  & 86.31 & 95.7 & 15.5 & 8.67 & 5.62 \\

    \hline  
    \hline
		Dataset: & \multicolumn{5}{ c|}{$\mathbf{\textit{MNIST}_{\textit{remap}}}$}& \multicolumn{5}{ c|}{$\mathbf{\textit{MNIST}_{\textit{rotate}}}$}\\ \hline
		Model & A.Acc & F.Acc & R.Spd & Ad.Rk & F.Rk & A.Acc & F.Acc & R.Spd & Ad.Rk & F.Rk \\ \hline
Baseline & 32.66 & 33.42 & --- & 10.0 & 10.0 & 49.78 & 50.2 & --- & 7.83 & 8.36 \\ 
\textit{NN-Patching}$_{\textit{incl,noEE}}$    & 94.86 & \textbf{98.74} & 4.3 & \textbf{2.76}& 1.98 & 72.75 & 77.25 & --- & 6.41 & 4.22  \\ 
\textit{NN-Patching}$_{\textit{incl,baseEE}}$  & 88.42 & 91.77 & 6.7 & 6.05 & 7.64 & 73.24 & 72.73 & --- & 4.59 & 5.6 \\  
\textit{NN-Patching}$_{\textit{semi,baseEE}}$  & 88.27 & 91.79 & 6.3 & 6.43 & 7.58 & 73.4 & 73.85 & --- & 4.63 & 4.64  \\
\textit{NN-Patching}$_{\textit{excl,baseEE}}$  & 88.19 & 90.87 & 9.1 & 6.13 & 8.18 & 71.75 & 69.95 & --- & 4.83 & 6.7 \\ 
\textit{NN-Patching}$_{\textit{incl,patchEE}}$ & \textbf{94.89} & 98.71 & 4.1 & 2.8 & \textbf{1.84} & 72.48 & 76.97 & --- & 6.61 & 4.4 \\
\textit{NN-Patching}$_{\textit{semi,patchEE}}$ & 94.52 & 98.69 & \textbf{4.0} & 3.35 & 2.24 & 70.4 & 72.84 & --- & 7.28 & 6.52 \\
\textit{NN-Patching}$_{\textit{excl,patchEE}}$ & 92.9 & 96.95 & 5.3 & 4.16 & 5.74 & 68.79 & 65.54 & --- & 5.97 & 8.0 \\ 
\textit{Freezing}                     & 91.51 & 97.55 & 7.4 & 6.91 & 4.92 & \textbf{74.61} & \textbf{79.55} & --- & \textbf{3.08} & \textbf{3.1}  \\
\textit{Base}$_{\textit{update}}$             & 91.68 & 97.6 & 7.0 & 6.41 & 4.88 & 73.49 & 78.54 & --- & 3.76 & 3.46 \\

    \hline 
		\end{tabular}
		}
		\end{centering}
		\resizebox{10.783cm}{!}{%
	  \begin{tabular}{ |l | l l l l l |}
    \hline
		Dataset: & \multicolumn{5}{ c|}{$\mathbf{\textit{MNIST}_{\textit{transfer}}}$} \\ \hline
		Model & A.Acc & F.Acc & R.Spd & Ad.Rk & F.Rk \\ \hline
Baseline & 0.0 & 0.0 & --- & 10.0 & 10.0 \\ 
\textit{NN-Patching}$_{\textit{incl,noEE}}$    & \textbf{94.73} & 98.7 & \textbf{4.0} & \textbf{3.88} & \textbf{3.1} \\ 
\textit{NN-Patching}$_{\textit{incl,baseEE}}$  & 94.49 & 98.67 & 4.7 & 4.52 & 4.88 \\  
\textit{NN-Patching}$_{\textit{semi,baseEE}}$  & 94.7 & \textbf{98.71} & \textbf{4.0} & 3.99 & 3.34 \\
\textit{NN-Patching}$_{\textit{excl,baseEE}}$  & 94.65 & 98.68 & 4.7 & 4.33 & 4.04 \\ 
\textit{NN-Patching}$_{\textit{incl,patchEE}}$ & 94.04 & 98.67 & 4.4 & 4.47 & 4.18 \\
\textit{NN-Patching}$_{\textit{semi,patchEE}}$ & 94.41 & 98.69 & 4.6 & 4.38 & 4.56 \\
\textit{NN-Patching}$_{\textit{excl,patchEE}}$ & 94.36 & 98.67 & 4.4 & 4.21 & 4.36 \\ 
\textit{Freezing}                     & 91.08 & 97.92 & 5.7 & 7.73 & 8.28 \\
\textit{Base}$_{\textit{update}}$             & 91.3 & 97.88 & 5.5 & 7.49 & 8.26 \\
    \hline  
  \end{tabular}
	}
\caption{Comparison of neural network patching and transfer learning techniques on \textit{MNIST} with CNN base classifiers.}
\label{Comparison of neural network patching and transfer learning techniques on $MNIST$ with CNN base classifiers}
\end{table}

\begin{table}[!htbp]
\begin{centering}
\resizebox{\columnwidth}{!}{%
  \begin{tabular}{ |l | l l l l l | l l l l l |}
    \hline
		\multicolumn{11}{|c|}{Convolutional Neural Network on \textit{NIST}}\\ \hline
		Dataset: & \multicolumn{5}{ c|}{$\mathbf{\textit{NIST}_{\textit{appear}}}$}& \multicolumn{5}{ c|}{$\mathbf{\textit{NIST}_{\textit{flip}}}$}\\ \hline
		Model & A.Acc & F.Acc & R.Spd & Ad.Rk & F.Rk & A.Acc & F.Acc & R.Spd & Ad.Rk & F.Rk \\ \hline
Baseline & 69.82 & 70.11  & --- & 9.55 & 10.0 & 17.72 & 18.1 & --- & 9.96 & 10.0 \\ 
\textit{NN-Patching}$_{\textit{incl,noEE}}$    & 91.44 & 94.51 & 6.7 & 4.67 & 3.84 & 90.48 & 93.77 & 7.9 & \textbf{2.59} & \textbf{2.1} \\ 
\textit{NN-Patching}$_{\textit{incl,baseEE}}$  & 91.55 & 93.85 & 4.7 & 3.36 & 5.58 & 89.18 & 92.29 & 8.4 & 3.62 & 4.78 \\
\textit{NN-Patching}$_{\textit{semi,baseEE}}$  & \textbf{91.59} & 93.96 & \textbf{3.9} & \textbf{3.19} & 5.52 & 89.15 & 92.18 & 8.5 & 3.69 & 4.96 \\
\textit{NN-Patching}$_{\textit{excl,baseEE}}$  & 90.35 & 93.17 & 6.4 & 4.65 & 7.5  & 87.48 & 91.2 & 12.4 & 5.99 & 6.44 \\ 
\textit{NN-Patching}$_{\textit{incl,patchEE}}$ & 91.38 & 94.38 & 6.5 & 5.0 & 4.48  & \textbf{90.52} & 93.75 & \textbf{7.6} & 2.65 & 2.12 \\
\textit{NN-Patching}$_{\textit{semi,patchEE}}$ & 90.81 & 94.36 & 7.4 & 6.31 & 4.56 & 90.32 & \textbf{93.79} & 8.0 & 2.97 & \textbf{2.1} \\
\textit{NN-Patching}$_{\textit{excl,patchEE}}$ & 87.96 & 89.79 & 4.5 & 5.27 & 8.98 & 87.3 & 91.6 & 13.1 & 6.48 & 5.54 \\ 
\textit{Freezing}                     & 91.4 & 94.95 & 7.5 & 5.06 & 2.42  & 71.3 & 88.4 & 36.8 & 8.08 & 8.6 \\
\textit{Base}$_{\textit{update}}$             & 90.33 & \textbf{95.03} & 11.2 & 7.95 & \textbf{2.12} & 65.6 & 88.58 & 40.0 & 8.96 & 8.36 \\
    \hline  
    \hline
		Dataset: & \multicolumn{5}{ c|}{$\mathbf{\textit{NIST}_{\textit{remap}}}$}& \multicolumn{5}{ c|}{$\mathbf{\textit{NIST}_{\textit{rotate}}}$}\\ \hline
		Model & A.Acc & F.Acc & R.Spd & Ad.Rk & F.Rk & A.Acc & F.Acc & R.Spd & Ad.Rk & F.Rk \\ \hline
Baseline & 11.48 & 11.27 & --- & 10.0 & 10.0 & 39.13 & 43.09 & --- & 5.55 & 6.74 \\ 
\textit{NN-Patching}$_{\textit{incl,noEE}}$    & 93.67 & \textbf{96.96} & \textbf{5.1} & 3.47 & 2.5 & 59.02 & \textbf{61.99} & --- & 6.45 & \textbf{3.78}  \\ 
\textit{NN-Patching}$_{\textit{incl,baseEE}}$  & 92.63 & 95.85 & 5.5 & 4.62 & 5.7 & \textbf{59.83} & 61.22 & --- & 4.55 & 4.3  \\
\textit{NN-Patching}$_{\textit{semi,baseEE}}$  & 92.79 & 95.79 & 4.9 & 4.21 & 6.22 & 59.75 & 61.25 & --- & 5.01 & 4.28  \\
\textit{NN-Patching}$_{\textit{excl,baseEE}}$  & 92.52 & 95.6 & 6.0 & 4.75 & 6.9 & 53.91 & 49.78 & --- & 6.37 & 8.28  \\ 
\textit{NN-Patching}$_{\textit{incl,patchEE}}$ & 93.46 & 96.9 & 5.2 & 3.5 & \textbf{2.3} & 58.53 & 61.93 & --- & 6.39 & 3.98 \\
\textit{NN-Patching}$_{\textit{semi,patchEE}}$ & \textbf{93.83} & \textbf{96.96} & 5.2 & \textbf{2.83} & 2.26 & 58.13 & 61.11 & --- & 7.42 & 4.98 \\
\textit{NN-Patching}$_{\textit{excl,patchEE}}$ & 92.79 & 96.21 & 5.3 & 4.65 & 4.98 & 52.65 & 55.54 & --- & 6.73 & 7.06 \\ 
\textit{Freezing}                     & 85.92 & 95.21 & 13.7 & 8.19 & 7.66 & 52.42 & 58.48 & --- & \textbf{3.17} & 5.56 \\
\textit{Base}$_{\textit{update}}$             & 83.74 & 95.7 & 20.1 & 8.79 & 6.48 & 51.28 & 56.64 & --- & 3.36 & 6.04 \\
    \hline 
		\end{tabular}
		}
		\end{centering}
		\resizebox{10.783cm}{!}{%
	  \begin{tabular}{ |l | l l l l l |}
    \hline
		Dataset: & \multicolumn{5}{ c|}{$\mathbf{\textit{NIST}_{\textit{transfer}}}$} \\ \hline
		Model & A.Acc & F.Acc & R.Spd & Ad.Rk & F.Rk \\ \hline
Baseline & 0.0 & 0.0 & --- & 9.99 & 10.0 \\ 
\textit{NN-Patching}$_{\textit{incl,noEE}}$    & \textbf{88.41} & 93.78 & 17.7 & \textbf{3.19} & \textbf{3.68} \\ 
\textit{NN-Patching}$_{\textit{incl,baseEE}}$  & 87.88 & 93.67 & 18.5 & 4.23 & 4.08 \\
\textit{NN-Patching}$_{\textit{semi,baseEE}}$  & 88.07 & 93.76 & \textbf{17.3} & 4.18 & 3.8 \\
\textit{NN-Patching}$_{\textit{excl,baseEE}}$  & 88.05 & 93.68 & \textbf{17.3} & 4.48 & 4.26 \\ 
\textit{NN-Patching}$_{\textit{incl,patchEE}}$ & 87.9 & 93.71 & 17.8 & 4.53 & 4.22 \\
\textit{NN-Patching}$_{\textit{semi,patchEE}}$ & 87.72 & \textbf{93.8} & 17.8 & 3.69 & 4.16 \\
\textit{NN-Patching}$_{\textit{excl,patchEE}}$ & 87.72 & 93.76 & 18.4 & 3.72 & 4.04 \\ 
\textit{Freezing}                     & 80.28 & 91.46 & 32.4 & 8.09 & 8.54 \\
\textit{Base}$_{\textit{update}}$             & 71.59 & 91.79 & 40.6 & 8.89 & 8.22 \\
    \hline  
  \end{tabular}
	}
\caption{Comparison of neural network patching and transfer learning techniques on \textit{NIST} with CNN base classifiers.}
\label{Comparison of neural network patching and transfer learning techniques on $NIST$ with CNN base classifiers}
\end{table}

\clearpage
\subsection{Result Tables for the Residual Base Classifiers}

\begin{table}[!htbp]
\begin{centering}
\resizebox{\columnwidth}{!}{%
  \begin{tabular}{ |l | l l l l l | l l l l l |}
    \hline
		\multicolumn{11}{|c|}{Residual Neural Network on \textit{MNIST}}\\ \hline
		Dataset: & \multicolumn{5}{ c|}{$\mathbf{\textit{MNIST}_{\textit{appear}}}$}& \multicolumn{5}{ c|}{$\mathbf{\textit{MNIST}_{\textit{flip}}}$}\\ \hline
		Model & A.Acc & F.Acc & R.Spd & Ad.Rk & F.Rk & A.Acc & F.Acc & R.Spd & Ad.Rk & F.Rk \\ \hline
Baseline & 50.88 & 51.28 & --- & 9.11 & 10.0 & 39.3 & 38.84 & --- & 9.93 & 10.0 \\
\textit{NN-Patching}$_{\textit{incl,noEE}}$    & 91.35 & 97.02 & 7.1 & 3.33 & 4.1 & \textbf{95.15} & 97.35 & \textbf{3.7} & \textbf{2.38} & 2.36 \\
\textit{NN-Patching}$_{\textit{incl,baseEE}}$  & 90.45 & 97.01 & 9.4 & 5.03 & 4.54 & 91.53 & 94.14 & 4.1 & 4.55 & 6.14 \\
\textit{NN-Patching}$_{\textit{semi,baseEE}}$  & 90.37 & 97.04 & 8.6 & 4.67 & 4.18 & 91.51 & 93.67 & 5.2 & 4.6 & 5.94 \\
\textit{NN-Patching}$_{\textit{excl,baseEE}}$  & 88.69 & 96.34 & 10.6 & 6.09 & 6.46 & 90.59 & 93.78 & 5.7 & 6.05 & 6.64 \\
\textit{NN-Patching}$_{\textit{incl,patchEE}}$ & 91.46 & 97.18 & \textbf{6.8} & 3.68 & 3.58 & 94.54 & 97.42 & 4.4 & 2.41 & 2.26 \\
\textit{NN-Patching}$_{\textit{semi,patchEE}}$ & 87.14 & 95.18 & 11.0 & 7.21 & 7.98 & 91.22 & 95.59 & 6.7 & 5.81 & 5.02 \\
\textit{NN-Patching}$_{\textit{excl,patchEE}}$ & 86.3 & 95.87 & 12.1 & 6.81 & 6.68 & 90.67 & 95.04 & 6.4 & 5.78 & 5.72 \\
\textit{Freezing}                     & 89.2 & 96.42 & 12.1 & 6.15 & 6.1 & 63.01 & 75.9 & --- & 8.82 & 9.0 \\
\textit{Base}$_{\textit{update}}$             & \textbf{92.17} & \textbf{98.37} & 6.9 & \textbf{2.82} & \textbf{1.38} & 94.17 & \textbf{97.96} & 5.2 & 4.67 & \textbf{1.92} \\
    \hline  
    \hline
		Dataset: & \multicolumn{5}{ c|}{$\mathbf{\textit{MNIST}_{\textit{remap}}}$}& \multicolumn{5}{ c|}{$\mathbf{\textit{MNIST}_{\textit{rotate}}}$}\\ \hline
		Model & A.Acc & F.Acc & R.Spd & Ad.Rk & F.Rk & A.Acc & F.Acc & R.Spd & Ad.Rk & F.Rk \\ \hline
Baseline & 32.19 & 33.37 & --- & 9.95 & 10.0 & 51.54 & 52.05 & --- & 7.35 & 7.62 \\
\textit{NN-Patching}$_{\textit{incl,noEE}}$    & 92.41 & 97.4 & 5.2 & 3.09 & 2.6 & \textbf{68.61} & 72.55 & --- & 4.33 & 3.34 \\
\textit{NN-Patching}$_{\textit{incl,baseEE}}$  & 88.0 & 92.29 & 6.1 & 5.19 & 6.28 & 65.92 & 66.12 & --- & 4.37 & 4.76 \\
\textit{NN-Patching}$_{\textit{semi,baseEE}}$  & 88.19 & 92.4 & 6.6 & 5.16 & 5.78 & 65.13 & 65.77 & --- & 4.87 & 4.94 \\
\textit{NN-Patching}$_{\textit{excl,baseEE}}$  & 87.31 & 92.09 & 9.2 & 5.83 & 6.4 & 62.47 & 63.31 & --- & 6.19 & 6.34 \\
\textit{NN-Patching}$_{\textit{incl,patchEE}}$ & 92.13 & 97.35 & \textbf{5.1} & \textbf{3.08} & 2.42 & 68.08 & \textbf{73.82} & --- & 5.28 & \textbf{3.32} \\
\textit{NN-Patching}$_{\textit{semi,patchEE}}$ & 88.72 & 93.35 & 6.8 & 5.1 & 5.94 & 60.04 & 59.14 & --- & 7.37 & 7.46 \\
\textit{NN-Patching}$_{\textit{excl,patchEE}}$ & 88.61 & 93.43 & 9.1 & 5.43 & 5.7 & 58.77 & 59.14 & --- & 7.53 & 7.64 \\
\textit{Freezing}                     & 65.26 & 83.07 & 44.2 & 8.88 & 8.68 & 62.84 & 67.61 & --- & 4.67 & 5.0 \\
\textit{Base}$_{\textit{update}}$             & \textbf{92.82} & \textbf{98.88} & 5.4 & 3.29 & \textbf{1.2} & 67.49 & 69.43 & --- & \textbf{3.05} & 4.58 \\

    \hline 
		\end{tabular}
		}
		\end{centering}
		\resizebox{10.783cm}{!}{%
	  \begin{tabular}{ |l | l l l l l |}
    \hline
		Dataset: & \multicolumn{5}{ c|}{$\mathbf{\textit{MNIST}_{\textit{transfer}}}$} \\ \hline
		Model & A.Acc & F.Acc & R.Spd & Ad.Rk & F.Rk \\ \hline
Baseline & 0.0 & 0.0 & --- & 9.93 & 10.0 \\
\textit{NN-Patching}$_{\textit{incl,noEE}}$    & \textbf{92.01} & 97.35 & \textbf{5.4} & \textbf{4.23} & 4.28 \\
\textit{NN-Patching}$_{\textit{incl,baseEE}}$  & 91.95 & 97.29 & 6.7 & 4.5 & 4.44 \\
\textit{NN-Patching}$_{\textit{semi,baseEE}}$  & 91.98 & 97.25 & 6.4 & 4.25 & 4.54 \\
\textit{NN-Patching}$_{\textit{excl,baseEE}}$  & 91.88 & 97.23 & 6.0 & 4.39 & 5.02 \\
\textit{NN-Patching}$_{\textit{incl,patchEE}}$ & 90.51 & 97.1 & 5.9 & 4.65 & 5.46 \\
\textit{NN-Patching}$_{\textit{semi,patchEE}}$ & 91.22 & 97.37 & 5.5 & 4.65 & 4.88 \\
\textit{NN-Patching}$_{\textit{excl,patchEE}}$ & 91.24 & 97.26 & 5.8 & 4.59 & 5.48 \\
\textit{Freezing}                     & 73.71 & 91.54 & 30.6 & 8.89 & 9.0 \\
\textit{Base}$_{\textit{update}}$             & 91.78 & \textbf{98.03} & 6.4 & 4.92 & \textbf{1.9} \\
    \hline  
  \end{tabular}
	}
\caption{Comparison of neural network patching and transfer learning techniques on \textit{MNIST} with ResNet base classifiers.}
\label{Comparison of neural network patching and transfer learning techniques on $MNIST$ with ResNet base classifiers}
\end{table}

\begin{table}[!htbp]
\begin{centering}
\resizebox{\columnwidth}{!}{%
  \begin{tabular}{ |l | l l l l l | l l l l l |}
    \hline
		\multicolumn{11}{|c|}{Residual Neural Network on \textit{NIST}}\\ \hline
		Dataset: & \multicolumn{5}{ c|}{$\mathbf{\textit{NIST}_{\textit{appear}}}$}& \multicolumn{5}{ c|}{$\mathbf{\textit{NIST}_{\textit{flip}}}$}\\ \hline
		Model & A.Acc & F.Acc & R.Spd & Ad.Rk & F.Rk & A.Acc & F.Acc & R.Spd & Ad.Rk & F.Rk \\ \hline
Baseline                     & 68.54 & 68.72 & --- & 8.92 & 10.0 & 15.66 & 15.93 & --- & 9.95 & 10.0 \\
\textit{NN-Patching}$_{\textit{incl,noEE}}$    & 88.34 & 92.14 & 8.9 & 4.87 & 3.5 & \textbf{87.68} & 93.21 & 9.8 & 3.21 & \textbf{2.54} \\
\textit{NN-Patching}$_{\textit{incl,baseEE}}$  & 86.42 & 90.26 & 10.9 & 4.77 & 5.76 & 86.55 & 91.36 & 10.3 & 3.43 & 5.15 \\
\textit{NN-Patching}$_{\textit{semi,baseEE}}$  & 86.17 & 90.17 & 10.0 & 4.93 & 6.3 & 86.52 & 91.24 & 10.5 & 3.69 & 5.24 \\
\textit{NN-Patching}$_{\textit{excl,baseEE}}$  & 84.1 & 89.41 & 15.7 & 6.09 & 7.34 & 84.36 & 90.55 & 13.9 & 5.85 & 6.32 \\
\textit{NN-Patching}$_{\textit{incl,patchEE}}$ & 88.51 & 92.21 & 8.8 & 4.75 & 3.42 & 87.2 & 93.08 & \textbf{9.3} & 3.39 & 2.68 \\
\textit{NN-Patching}$_{\textit{semi,patchEE}}$ & 88.43 & 92.28 & 9.6 & 5.14 & 3.14 & 87.23 & \textbf{93.25} & 9.7 & \textbf{3.0} & 2.59 \\
\textit{NN-Patching}$_{\textit{excl,patchEE}}$ & 79.1 & 83.81 & 23.4 & 6.86 & 8.88 & 84.0 & 90.66 & 13.4 & 6.16 & 5.98 \\
\textit{Freezing}                     & 87.44 & 91.26 & 11.3 & 4.99 & 4.9 & 49.43 & 72.61 & --- & 9.07 & 9.15 \\
\textit{Base}$_{\textit{update}}$             & \textbf{89.67} & \textbf{93.42} & \textbf{7.4} & \textbf{3.68} & \textbf{1.76} & 80.64 & 91.3 & 31.6 & 7.25 & 5.35 \\

    \hline  
    \hline
		Dataset: & \multicolumn{5}{ c|}{$\mathbf{\textit{NIST}_{\textit{remap}}}$}& \multicolumn{5}{ c|}{$\mathbf{\textit{NIST}_{\textit{rotate}}}$}\\ \hline
		Model & A.Acc & F.Acc & R.Spd & Ad.Rk & F.Rk & A.Acc & F.Acc & R.Spd & Ad.Rk & F.Rk \\ \hline
Baseline                  & 12.79 & 10.98 & --- & 9.81 & 10.0 & 38.47 & 41.42 & --- & 5.72 & 7.82 \\
\textit{NN-Patching}$_{\textit{incl,noEE}}$    & \textbf{86.04} & 92.68 & \textbf{10.7} & \textbf{3.56} & 3.53 & 53.19 & 57.67 & ---- & 5.8 & 3.86 \\
\textit{NN-Patching}$_{\textit{incl,baseEE}}$  & 85.14 & 91.7 & 13.8 & 4.15 & 4.91 & \textbf{54.56} & 55.81 & --- & 4.27 & 4.52 \\
\textit{NN-Patching}$_{\textit{semi,baseEE}}$  & 85.08 & 91.1 & 11.4 & 3.97 & 5.38 & 54.37 & 56.2 & --- & 4.32 & 4.5 \\
\textit{NN-Patching}$_{\textit{excl,baseEE}}$  & 84.37 & 90.75 & 15.0 & 4.57 & 6.66 & 48.21 & 44.73 & --- & 6.13 & 8.2 \\
\textit{NN-Patching}$_{\textit{incl,patchEE}}$ & 85.0  & 92.24 & 11.6 & 4.13 & 3.71 & 53.62 & \textbf{58.0} & --- & 6.09 & \textbf{3.62} \\
\textit{NN-Patching}$_{\textit{semi,patchEE}}$ & 85.29 & 92.4 & 11.5 & 4.39 & 3.69 & 53.15 & 57.52 & --- & 6.1 & 4.1 \\
\textit{NN-Patching}$_{\textit{excl,patchEE}}$ & 83.64 & 91.63 & 15.5 & 5.07 & 5.28 & 46.35 & 52.06 & --- & 6.54 & 7.16 \\
\textit{Freezing}                     & 53.94 & 70.3 & 43.1 & 8.64 & 8.96 & 45.64 & 51.72 & --- & 5.91 & 5.84 \\
\textit{Base}$_{\textit{update}}$             & 83.75 & \textbf{92.93} & 16.2 & 6.71 & \textbf{2.88} & 49.14 & 54.25 & --- & \textbf{4.12} & 5.38 \\

    \hline 
		\end{tabular}
		}
		\end{centering}
		 \hfill{}
		\resizebox{10.783cm}{!}{%
	  \begin{tabular}{ |l | l l l l l |}
    \hline
		Dataset: & \multicolumn{5}{ c|}{$\mathbf{\textit{NIST}_{\textit{transfer}}}$} \\ \hline
		Model & A.Acc & F.Acc & R.Spd & Ad.Rk & F.Rk \\ \hline
Baseline & 0.0 & 0.0 & --- & 9.53 & 10.0 \\
\textit{NN-Patching}$_{\textit{incl,noEE}}$    & \textbf{77.83} & \textbf{87.22} & 55.8 & 3.98 & 3.94 \\
\textit{NN-Patching}$_{\textit{incl,baseEE}}$  & 77.65 & 87.15 & 55.9 & 4.05 & \textbf{3.93} \\
\textit{NN-Patching}$_{\textit{semi,baseEE}}$  & 77.71 & 86.97 & 56.1 & 4.01 & 4.28 \\
\textit{NN-Patching}$_{\textit{excl,baseEE}}$  & 77.6  & 87.17 & 55.8 & \textbf{3.97} & 4.02 \\
\textit{NN-Patching}$_{\textit{incl,patchEE}}$ & 75.88 & 86.93 & \textbf{55.6} & 4.98 & 4.34 \\
\textit{NN-Patching}$_{\textit{semi,patchEE}}$ & 76.05 & 86.94 & 55.7 & 4.85 & 4.5 \\
\textit{NN-Patching}$_{\textit{excl,patchEE}}$ & 74.46 & 87.17 & 56.1 & 5.46 & 4.26 \\
\textit{Freezing}                     & 66.51 & 80.27 & ---  & 7.62 & 8.42 \\
\textit{Base}$_{\textit{update}}$              & 73.36 & 86.38 & 57.9 & 6.55 & 7.22 \\
    \hline  
  \end{tabular}
	}
\caption{Comparison of neural network patching and transfer learning techniques on \textit{NIST} with ResNet base classifiers.}
\label{Comparison of neural network patching and transfer learning techniques on $NIST$ with ResNet base classifiers}
\end{table}

\end{document}